\documentclass[runningheads]{llncs}

\usepackage{eccv}

\usepackage{eccvabbrv}

\usepackage{graphicx}
\usepackage{booktabs}
\usepackage{wrapfig,lipsum,booktabs}

\usepackage[accsupp]{axessibility}  %

\newcommand{\ourmethod}{DGD}

\makeatletter
\DeclareRobustCommand\onedot{\futurelet\@let@token\@onedot}
\def\@onedot{\ifx\@let@token.\else.\null\fi\xspace}

\makeatother

\let\titleold\title
\renewcommand{\title}[1]{\titleold{#1}\newcommand{\thetitle}{#1}}

\AtEndPreamble{
    \usepackage[capitalize]{cleveref}
    \crefname{section}{Sec.}{Secs.}
    \Crefname{section}{Section}{Sections}
    \Crefname{table}{Table}{Tables}
    \crefname{table}{Tab.}{Tabs.}
}

\RequirePackage{enumitem}
\setlist[itemize]{noitemsep,leftmargin=*,topsep=0em}
\setlist[enumerate]{noitemsep,leftmargin=*,topsep=0em}

\usepackage[pagebackref,breaklinks,colorlinks,citecolor=eccvblue]{hyperref}

\usepackage{orcidlink}

\begin{document}

\title{\ourmethod{}: Dynamic 3D Gaussians Distillation}

\titlerunning{\ourmethod{}: Dynamic 3D Gaussians Distillation}

\author{Isaac Labe\inst{1}~
Noam Issachar\inst{1}~
Itai Lang\inst{2}~
Sagie Benaim\inst{1}}

\authorrunning{I.~Labe et al.}

\institute{The Hebrew University of Jerusalem \and
University of Chicago}

\maketitle

\begin{abstract}

We tackle the task of learning dynamic 3D semantic radiance fields given a single monocular video as input. Our learned semantic radiance field captures per-point semantics as well as color and geometric properties for a dynamic 3D scene, enabling the generation of novel views and their corresponding semantics. This enables the 
segmentation and tracking of a diverse set of 3D semantic entities, specified using a simple and intuitive interface that includes a user click or a text prompt. 
To this end, we present \ourmethod{}, a unified 3D representation for both the appearance and semantics of a dynamic 3D scene, building upon the recently proposed dynamic 3D Gaussians representation.
Our representation is optimized over time with both color and semantic information. Key to our method is the joint optimization of the appearance and semantic attributes, which jointly affect the geometric properties of the scene.
We evaluate our approach in its ability to enable dense semantic 3D object tracking
and   
demonstrate 
high-quality results that are fast to render, for a diverse set of scenes. Our project webpage is available on \url{https://isaaclabe.github.io/DGD-Website/}

\end{abstract}

\section{Introduction} \label{sec:introduction}

Representing the various aspects of our 3D world, including appearance, dynamics, and semantics, is a long-standing problem in computer vision. Recent work has shown a significant improvement in the ability to render high-fidelity dynamic 3D quickly and in tracking 3D points in space and time~\cite{kerbl20233dgd, yang2023deformable}. In this work, we take a step further and enable the tracking of a diverse set of 3D \textit{semantic entities}. These semantic entities can be specified using a simple and intuitive interface that includes \textit{ a click or a text prompt}, thus allowing for an intuitive interface for non-experts. 

A significant body of work in this space (e.g., \cite{yu2021pixelnerf, barron2021mipnerf, barron2022mipnerf360, verbin2022refnerf, fridovich2022plenoxels, chen2022tensorf, yang2022polynomial, benaim2024volumetric}) is based on NeRF~\cite{mildenhall2021nerf}, a 3D volumetric scene representation that allows high-quality rendering of novel views. While impressive, these works suffer from the inability to render high-quality views in real-time. To this end, 3D Gaussian Splatting (3DGS)~\cite{kerbl20233dgd} presented a novel 3D representation based on the rasterization of 3D Gaussians, enabling high-quality renderings in real-time. 
Based on this representation, recent concurrent work~\cite{zhou2023feature3dgs, qin2023langsplat, zuo2024fmgs} incorporated the distillation of semantic information into a static 3D Gaussian representation, enabling the segmentation of 3D entities and their subsequent high-quality renderings in real-time.

Beyond static scenes, recent work~\cite{luiten2023dynamic3dg, yang2023deformable, wu20234d, yang2023real} extended 3DGS to the dynamic setting, enabling high-quality real-time dynamic 3D reconstruction, as well as for tracking 3D points in space and time quickly and efficiently. While impressive, unlike our work, these works cannot track 3D semantic entities in space and time, based on an intuitive user interface such as text or a click. 

\begin{figure*}[t!]
\centering

\begin{tabular}{lccccc}

& $t=0$ & $t=200$ & $t=400$ & $t=600$ & $t=800$  \\ 

\rotatebox{90}{\hspace{0.3cm} \tiny{ Rendered View}} & 
\includegraphics[width=0.17\linewidth, trim={0 5.025cm 0 0},clip]{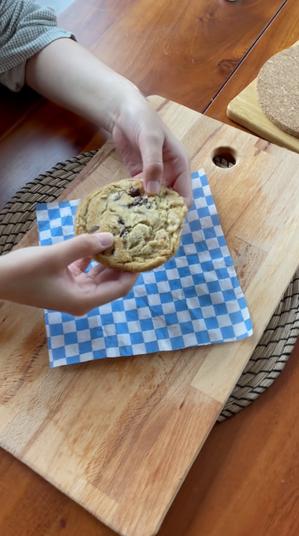} & 
\includegraphics[width=0.17\linewidth, trim={0 5.025cm 0 0},clip]{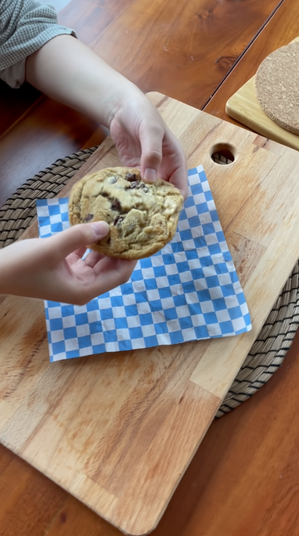} & 
\includegraphics[width=0.17\linewidth, trim={0 5.025cm 0 0},clip]{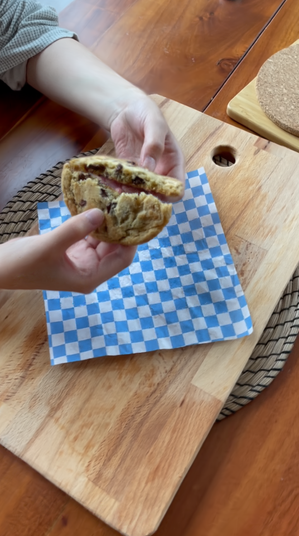} & 
\includegraphics[width=0.17\linewidth, trim={0 5.025cm 0 0},clip]{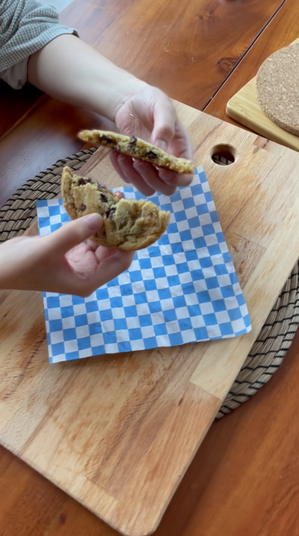} & 
\includegraphics[width=0.17\linewidth, trim={0 5.025cm 0 0},clip]{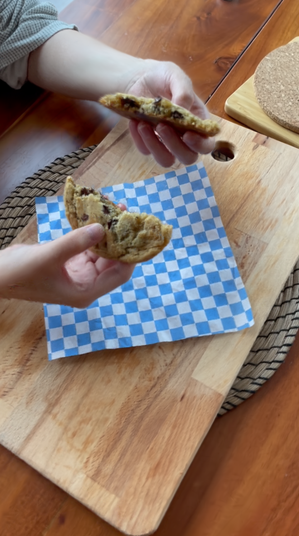} \\
\rotatebox{90}{\hspace{0.5cm} \tiny{Segmentation}} & 
\includegraphics[width=0.17\linewidth, trim={0 5.025cm 0 0},clip]{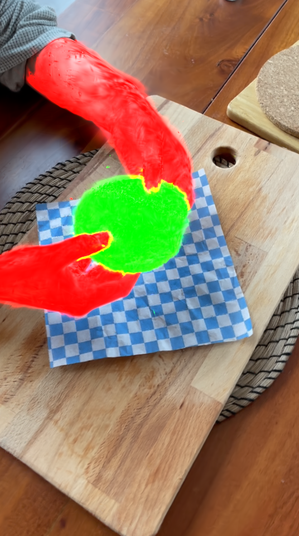} & 
\includegraphics[width=0.17\linewidth, trim={0 5.025cm 0 0},clip]{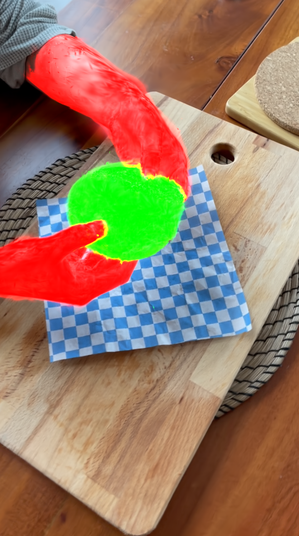} & 
\includegraphics[width=0.17\linewidth, trim={0 5.025cm 0 0},clip]{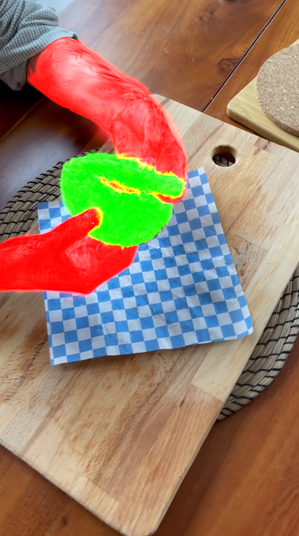} & 
\includegraphics[width=0.17\linewidth, trim={0 5.025cm 0 0},clip]{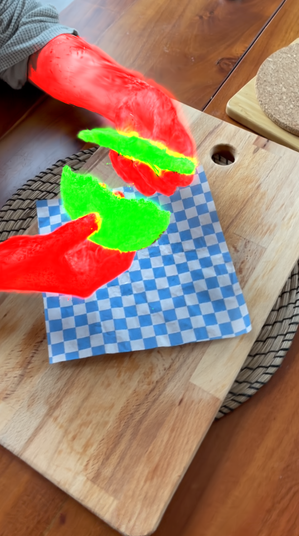} & 
\includegraphics[width=0.17\linewidth, trim={0 5.025cm 0 0},clip]{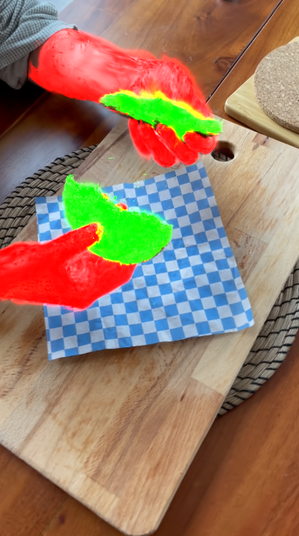} \\ 

\rotatebox{90}{\hspace{0.3cm} \tiny{ Rendered View}} & 
\includegraphics[width=0.17\linewidth, trim={0 5.025cm 0 0},clip]{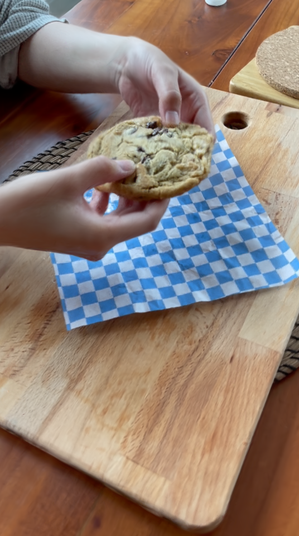} & 
\includegraphics[width=0.17\linewidth, trim={0 5.025cm 0 0},clip]{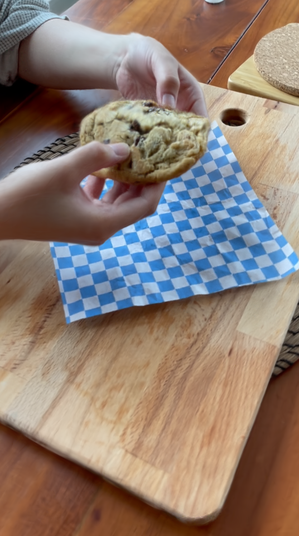} & 
\includegraphics[width=0.17\linewidth, trim={0 5.025cm 0 0},clip]{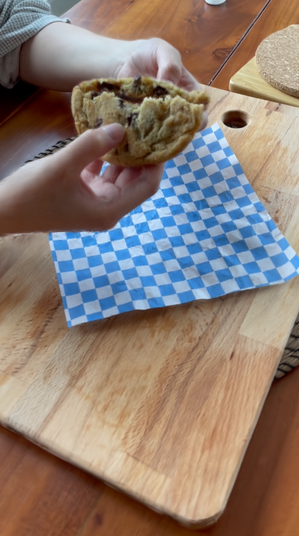} & 
\includegraphics[width=0.17\linewidth, trim={0 5.025cm 0 0},clip]{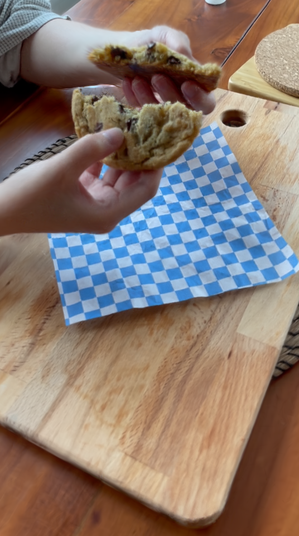} & 
\includegraphics[width=0.17\linewidth, trim={0 5.025cm 0 0},clip]{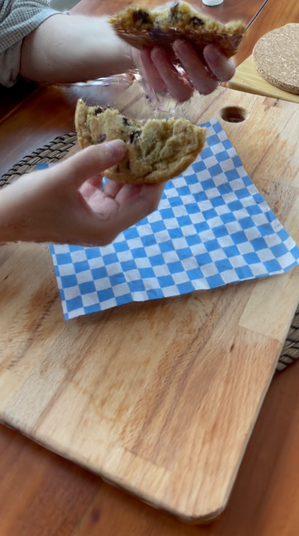} \\
\rotatebox{90}{\hspace{0.5cm} \tiny{Segmentation}} & 
\includegraphics[width=0.17\linewidth, trim={0 5.025cm 0 0},clip]{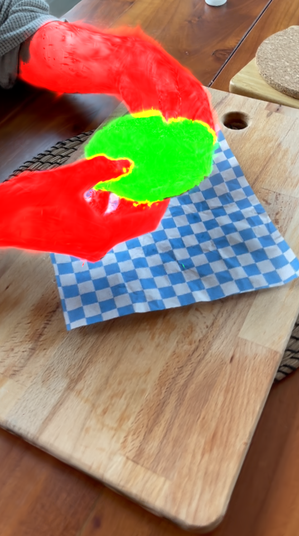} & 
\includegraphics[width=0.17\linewidth, trim={0 5.025cm 0 0},clip]{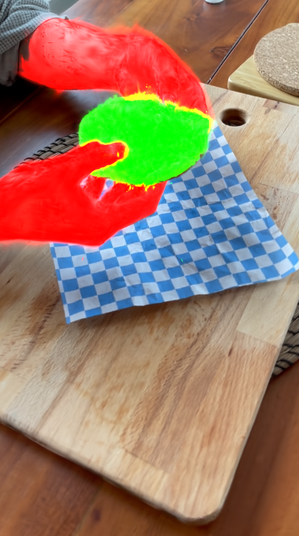} & 
\includegraphics[width=0.17\linewidth, trim={0 5.025cm 0 0},clip]{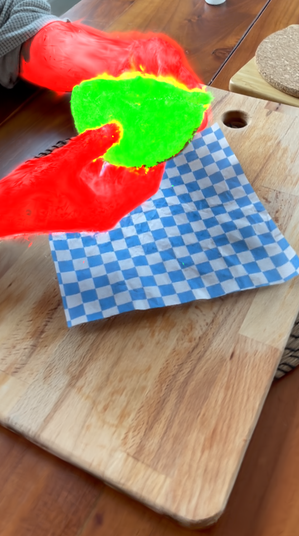} & 
\includegraphics[width=0.17\linewidth, trim={0 5.025cm 0 0},clip]{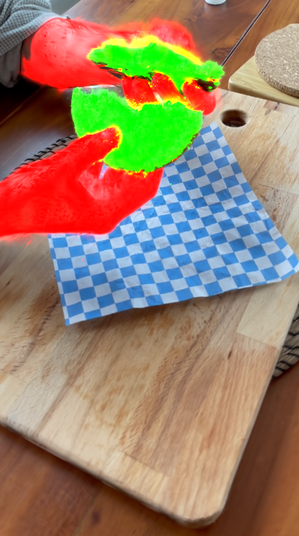} & 
\includegraphics[width=0.17\linewidth, trim={0 5.025cm 0 0},clip]{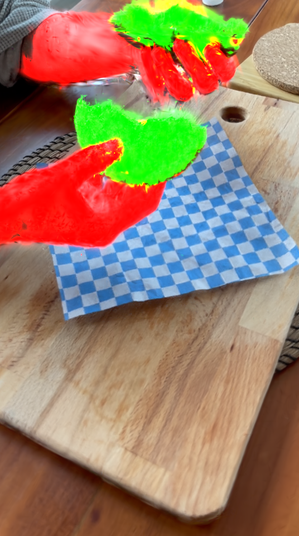} \\ 

\end{tabular}
\caption{Five
rendered novel views of our method in timesteps $0$, $200$, $400$, $600$, and $800$, for two fixed camera poses chosen at random (first/third rows). 
Below the novel views, we provide the corresponding segmentation and tracking of every dense point. We consider objects corresponding to a 3D click on  ``hands" (red) and ``cookie" (green). 
}
\label{fig:teaser}
\vspace{-0.4cm}
\end{figure*}

In this work, we present \ourmethod{}, a unified 3D representation for both the appearance and semantics of a dynamic 3D scene. We build upon the dynamic 3DGS optimization technique, and optimize Gaussian's changes through time along with its color and semantic information. The key to our method is the joint optimization of the appearance and semantic attributes, which jointly affect the geometric properties of the scene. %

More specifically, for each Gaussian, we define the learnable variables: spatial parameters (location, orientation, and scale), appearance parameters (color and density), and a high-dimensional semantic feature vector. Given a collection of images of the scene, we extract semantic feature maps for each view by utilizing 2D foundation models 
(e.g. \cite{li2022languagedriven, kirillov2023segment}). Then, we rasterize the Gaussians to different views and optimize their parameters to match the scene observations. Notably, the spatial parameters of the Gaussians are influenced by both the color and semantic feature optimization, thus better adhering to the supervision signals and improving the representation quality of the scene.

We evaluate our approach by considering its 
ability to segment and track 3D semantic entities in time and space, which are specified using either text or a 3D click. 
Our method 
enables high quality and rendering speed 
for a variety of real and synthetic scenes.

\section{Related Work} \label{sec:related_work}

\subsection{3D Radiance Fields}

In recent years, novel 3D representations have emerged, revolutionizing the field of novel view synthesis (e.g., \cite{barron2021mipnerf, 
park2019deepsdf, peng2020convolutional, mescheder2019occupancy, 
mildenhall2021nerf}). 
In the static case, NeRF~\cite{mildenhall2021nerf} introduced an implicit 3D representation that uses a coordinate-based neural network to represent a 3D scene volumetrically.
Although progress in implicit representation enabled compact~\cite{chen2022tensorf, chan2022efficient, fridovich2023k}, full-360~\cite{barron2022mipnerf360} and fast training time~\cite{muller2022instant}, due to slow rendering speeds and substantial memory usage during training, the community has moved to make use of hybrid or fully explicit structures instead~\cite{fridovich2022plenoxels, muller2022instant}. 
Most recently, 3D Gaussian Splatting~\cite{kerbl20233dgd}, has been proposed to model a scene as a set of 3D Gaussians, and achieved state-of-the-art novel view synthesis rendering quality and fast training and rendering. Our approach moves beyond the static setting and considers a dynamic and semantic representation. 

In the dynamic setting, one needs to effectively encode temporal information, 
especially when only a monocular video is given in training. 
One set of approaches model scene deformation by adding time $t$ as input to a radiance field, but this requires heavy regularization to ensure temporal consistency. Another set of approaches~\cite{park2021nerfies, park2021hypernerf, pumarola2021d} decouples the deformation from the static representation of the scene, by mapping points to a canonical space through a deformation field. Recently, a number of approaches~\cite{yang2023deformable, luiten2023dynamic3dg, wu20234d, yang2023real} were introduced, making use of an underlying 3D Gaussian representation of the dynamic scene for fast and high-quality renderings of novel views. 
Gaussian-Flow~\cite{lin2023gaussian}, introduced a point-based approach for fast dynamic scene reconstruction and real-time rendering, using an underlying 3D Gaussian representation, where the input is either multi-view or  monocular videos. Other work ~\cite{katsumata2023efficient} presents an efficient 3D Gaussian representation for dynamic scenes in which positions and rotations are defined as a function of time, while other time-invariant properties are left unchanged. 
Perhaps most similar to ours, is the work of \cite{yang2023deformable} that decouples the scene into a static 3D Gaussian representation in canonical space and a separate deformation network that operates on these Gaussians. 
Our approach similarly decouples a static 3D Gaussian based representation and a separate deformation network.
However, our work captures not only appearance and geometry but also 3D semantics and their deformation over time.

\subsection{3D Feature Distillation}

The ability to obtain a 3D representation capturing not only appearance and geometry, but also semantics, is of great importance for 3D understanding and editing. 
However, collecting and annotating 3D semantic data is significantly challenging, and so recent methods focused on the ability
to lift 2D semantic data to 3D while simultaneously representing geometry and appearance. 
For example, Semantic NeRF~\cite{zhi2021place} and Panoptic Lifting~\cite{siddiqui2023panoptic} have successfully lifted 2D segmentation data to 3D. 
Distilled Feature Fields~\cite{kobayashi2022decomposing},  LERF~\cite{kerr2023lerf}, and Neural Feature Fusion Fields~\cite{tschernezki2022neural},
have shown the ability to lift features from 2D pretrained feature maps. These maps could be generated using 2D vision foundation models such as CLIP~\cite{radford2021learning} or
DINO~\cite{caron2021emerging} into NeRF frameworks, resulting in point-based 3D semantic features in the same space of the underlying 2D features maps. 
When lifting CLIP features, as in LeRF, for instance, one can then query the underlying 3D representation (e.g. NeRF) with text, thus locating or segmenting the object with open-vocabulary text. Unlike our work, these works consider static scenes and cannot handle dynamic scenes. 

Recently, and concurrently with our work, authors considered a different underlying explicit representation of 3D Gaussians. This avoids a costly rendering process and obtains efficient 3D semantic fields. 
\cite{zhou2023feature3dgs} distilled the knowledge of 2D foundation models and improved each Gaussian with a semantic feature vector to enable various applications, such as interactive segmentation and language-guided scene editing. Similalry, \cite{qin2023langsplat} considered the distillation of language based features. However, they considered only static scenes and cannot handle dynamic scenes.

\subsection{3D Tracking}  %

Video tracking typically considers the case of tracking objects (bounding box or segmentation masks) over time ~\cite{kristan2021ninth, perazzi2016benchmark, voigtlaender2019mots}. 
As an alternative, it may be desirable to track dense points in a number of timesteps~\cite{beauchemin1995computation, vedula1999three, doersch2023tapir, doersch2022tap}. 
This task becomes more challenging when one wishes to track points not only in 2D but also in 3D. 
This is a more challenging task that was tackled with a number of deep learning approaches ~\cite{doersch2023tapir, doersch2022tap}, that unlike our method, were trained
on large supervised datasets with ground truth points. 
Recent work considered the dynamic tracking of 3D Gaussians through time, where only a monocular video is given as input~\cite{yang2023deformable}.
Although impressive, this work does not allow the tracking of semantic entities and their associated 3D points throughout time. 

Several works consider the setting of 3D semantic object tracking given monocular video as input. \cite{zheng2023editablenerf} considers topologically varying dynamics and can track and edit based on picked-out surface key points. Unlike our work, it cannot densly segment 3D points. $D^2NeRF$~\cite{wu2022d} is a self-supervised approach that learns a 3D scene representation that decomposes moving objects from the static background. Similarly, \cite{wong2023factored} factorizes raw RGBD monocular video to produce object-level neural representation. Our work, on the other hand, can be used to select any moving or static object based on a click or text prompt. 

SAFF~\cite{liang2023semantic} builds a NeRF of scene-flowed 3D density, radiance, semantics, and
attention. 
Concurrently to our work,  Semantic Flow~\cite{tian2023semantic} was proposed. It learns semantics from continuous flows that contain rich 3D motion information. 
In contrast to these works, our work (1). enables the distillation of arbitrary 2D semantic features, going beyond segmentation, (2). requires 2D moncoular video supervision only without ground truth flow input, and (3). is based on a 3D Gaussian splitting representation, as opposed to a NeRF based representation, thus enabling fast rendering times.

\begin{figure*}[t!]
\centering
\includegraphics[width=\linewidth]{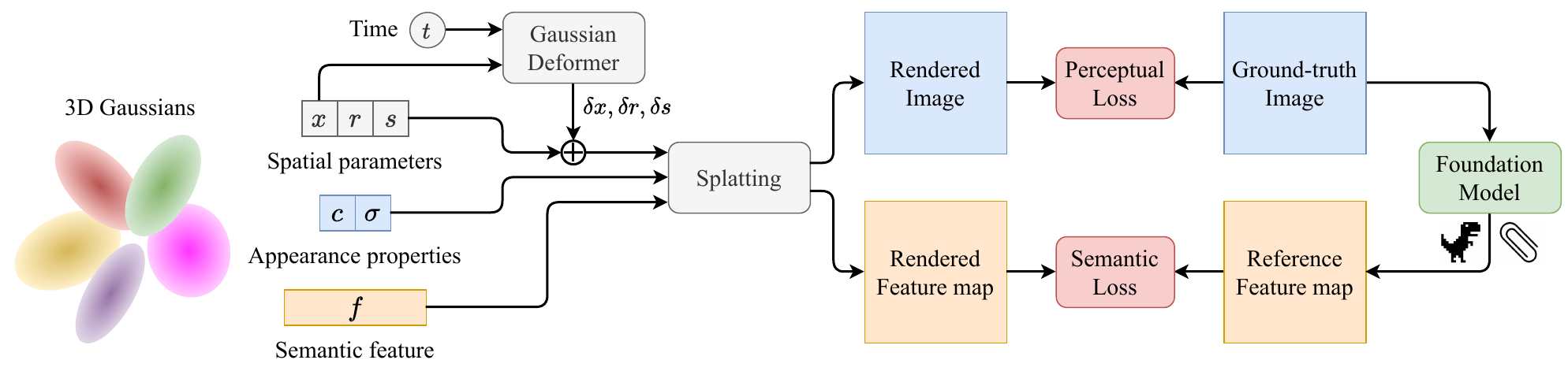}
\caption{\textbf{The proposed method.} \ourmethod{} utilizes 3D Gaussian representation and optimizes spatial parameters of the Gaussians and their deformation, concurrently with appearance properties with a semantic feature per Gaussian. Our learned representation enables efficient semantic understanding and manipulation of dynamic 3D scenes.
} 
\label{fig:system}
\vspace{-0.2cm}
\end{figure*}

\section{Method} \label{sec:method}

We begin by outlining the underlying representation used by our method for dynamic 3D reconstruction and novel view synthesis which is inspired by \cite{yang2023deformable}. Building upon this representation, we outline our novel distillation approach, which enables the depiction and tracking of 3D semantic entities using an intuitive user interface. 
We then provide implementation details. An illustration of our method is provided in \cref{fig:system}.

\subsection{Dynamic 3D Gaussians}

As input, we are given a monocular video, defined as a sequence of $N$ frames $\{I_1, \dots, I_N \}$, where $I_t \in \mathbb{R}^{H \times W \times 3}$ where $H$ and $W$ are the height and width. For each frame $I_t$, we assume the corresponding camera intrinsics and extrinsics parameters, $K_t$ and $E_t$ respectively, calibrated using SfM~\cite{schonberger2016structure}. SfM is also used to produce an initial sparse point cloud $P$. $P$ is then used to create an initial set of 3D Gaussians $\{ G(x_i, r_i, s_i, \sigma_i) \}$, where $x_i \in P$ is the center point of the Gaussian. 
$r_i$ is a quaternion defining the Gaussian's rotation, while $s_i$ represents the scaling parameter. Together they define the Gaussian's covariance matrix $\Sigma$.  

A separate deformation field $D$, parameterized as a multilayer perceptron (MLP), operates on the set of Gaussians. More specifically, given a Gaussian $G(x_i, r_i, s_i, \sigma_i)$, $D$ takes as input the current time $t$ and position $x_i$ of $G$ and outputs $\delta r_i$ $\delta r_i$, $\delta s_i$, resulting in a new Gaussian $G(x_i + \delta x_i, r_i + \delta r_i, s_i + \delta s_i, \sigma_i)$. 

\subsection{Dynamic Feature Distillation}
\label{sec:dynamic_feature_distillation}

Our work builds upon this representation by allowing Gaussians to represent not only colors, but also features. 
More specifically, each Gaussian $G(x_i, r_i, s_i, \sigma_i)$ has an associated learnable color value $c_i$ and a learnable feature value $f_i$. The Gaussian parameters, $\{x_i, r_i, s_i, c_i, f_i\}$, as well as the parameters of the deformation field $D$, are learned by using an associated rasterization strategy. As input, we assume a monocular video with frames $\{I_1, \dots, I_N \}$, and associated 2D features $\{F_1, \dots, F_N \}$, where $F_t \in \mathbb{R}^{H \times W \times C}$. We assume $F_t = F(I_t)$, where $F$ is some pretrained 2D feature extractor (e.g. CLIP~\cite{radford2021learning},
DINOv2~\cite{oquab2023dinov2}). We assume $I_t$ and $F_t$ have the same spatial dimension. This is achieved by upsampling $F_t$ to the resolution of $I_t$. 
It is also possible to assume a video feature extractor that translates a sequence of frames into a sequence of features. We begin by describing the optimization in the canonical space, where only Gaussian parameters are optimized. 

In the canonical space, rasterization and learning is performed using a methodology inspired by \cite{kerbl20233dgd}. More specifically, we can project 3D Gaussians to 2D and render each pixel using the following covariance matrix: 
\begin{align}
\Sigma' = J W \Sigma W^T J^T  
\end{align}
where $J$ is the Jacobian of the affine approximation of the
projective transformation, $W$ is a view transformation matrix, and $\Sigma$ denotes the covariance matrix. 
Following \cite{kerbl20233dgd}, for better optimization, the covariance matrix $\Sigma$ can be described as:
\begin{align}
\Sigma = R S S^T R^T
\end{align}
where $S$ is a scaling matrix, $R$ is a rotation matrix derived from a 3D scaling vector $s$ and quaternion $r$.

To render 2D colors and features, for a given pixel position $p$, one can use volumetric rendering:
\begin{align}
C(p) &= \sum_{i=1}^N T_i \alpha_i c_i \label{eq:rendering} \\
F(p) &= \sum_{i=1}^N T_i \alpha_i f_i \\
\alpha_i &= \sigma_i  e^{-0.5 (p-\mu_i)^T \Sigma' (p-\mu_i)}
\end{align}
where $T_i$ is the transmitance, defined by $\prod_{j=1}^{i-1}(1-\alpha_i)$, $c_i$ and $f_i$ are the color and feature values of a Gaussian, and $\mu_i$ is the $uv$ coordinate of the 3D Gaussian coordinates projected to 2D planes.

Given a center position $x$ of the Gaussian and a time $t$, a deformation MLP $D$, produces offsets to the Gaussians. That is:
\begin{align}
\delta x_i, \delta r_i, \delta s_i = D(\gamma(x_i), \gamma(t))
\end{align}
where $\gamma$ is frequency based encoding of \cite{tancik2020fourier}. %

Given colors and positions, a reconstruction loss can be used with the ground truth colors and features constructed as above. %
In particular, we use $D$ to deform Gaussians to timeframe $t$ and then use the rasterization process defined in \cref{eq:rendering} to render color and feature values for a given pixel position. These can be compared to the input ground truth color and feature values. 
We note that the Gaussian parameters, including position, density, and covariance matrix, as well as defomation network parameters, are jointly optimized against both the ground truth color and the feature values.

\subsection{3D Semantic Tracking}

\label{sec:tracking}
After optimization, a subset of the Gaussians can be selected using an intuitive user interface. This can be done, for example, in the following ways, which can also be combined together.  

First, assuming that CLIP features are used for distillation, each Gaussian is then associated with a feature value $f_i$ in the CLIP feature space. Hence, given a text prompt $txt$, one can compare the similarity of the CLIP encoding $E(txt)$ with each of the $f_i$'s and select those (as well as the associated Gaussians) most similar to $E(txt)$ - i.e those with cosine similarity above a certain threshold. 
Note that to obtain the Gassians at timestep $t$, one can apply the deformation network $D$ to the selected Gaussians in the canonical space. We can also consider CLIP encoding of images instead of $E(txt)$ using this approach. 

Second, a user can select a 3D Gaussian in 3D space, with an associated feature $f_i$ (e.g. a feature in the DINO representation space, following distillation from 2D DINO features). We can then consider the similarity of $f_i$ to all other features of Gaussians in 3D space and only select those with a similarity above a certain threshold. Note that one can also select a group of Gaussains and consider similarity to the average feature $f_i$. In our experiments, we select Gaussians based on a single feature $f_i$ selected by the user. 

Third, given an input video used for training, one can select a subset of pixels from all frames. We can then consider the set of Gaussians that affect those pixels or contribute to the color of these pixels using \cref{eq:rendering}. More specifically, one can select Gaussians whose weight, $T_i \alpha_i$ is close to 1.
In this work, we consider the first two selection types and leave the third to future work.

\begin{figure*}
\centering
\begin{tabular}{ccccc}

$t=0$ & $t=200$ & $t=400$ & $t=600$ & $t=800$ \\
\rotatebox{90}{\hspace{0.5cm}\tiny{Rendered View}}
\includegraphics[width=0.17\linewidth, trim={0 5.025cm 0 0},clip]{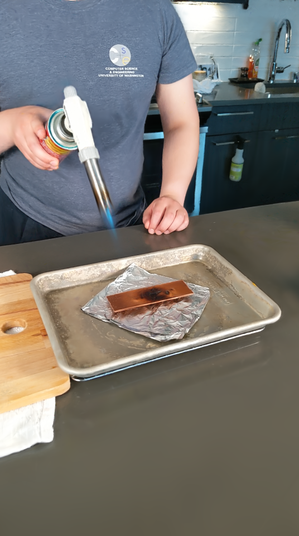} & 
\includegraphics[width=0.17\linewidth, trim={0 5.025cm 0 0},clip]{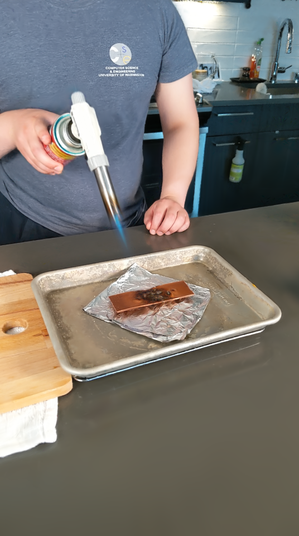} & 
\includegraphics[width=0.17\linewidth, trim={0 5.025cm 0 0},clip]{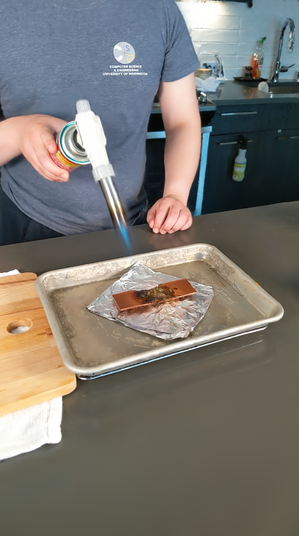} & 
\includegraphics[width=0.17\linewidth, trim={0 5.025cm 0 0},clip]{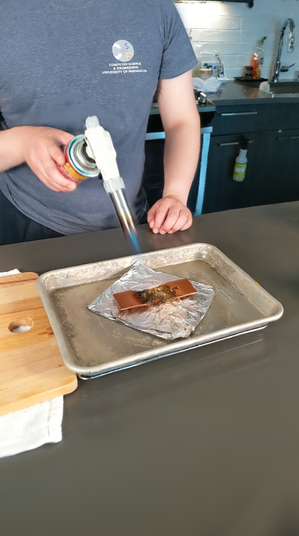} & 
\includegraphics[width=0.17\linewidth, trim={0 5.025cm 0 0},clip]{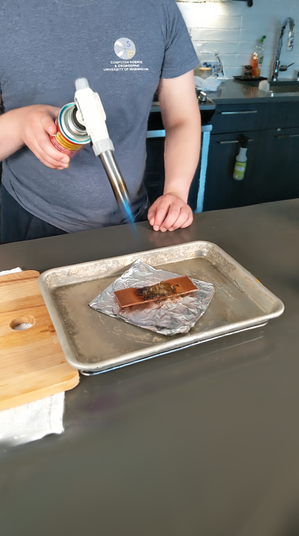} \\
\rotatebox{90}{\hspace{0.5cm}\tiny{Segmentation}}
\includegraphics[width=0.17\linewidth, trim={0 5.025cm 0 0},clip]{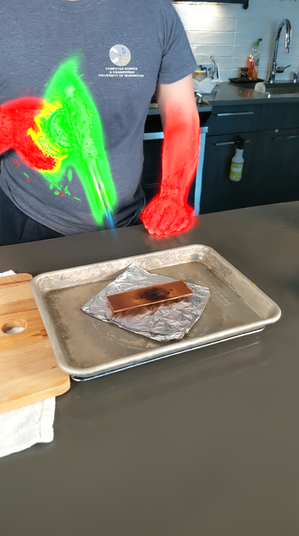} & 
\includegraphics[width=0.17\linewidth, trim={0 5.025cm 0 0},clip]{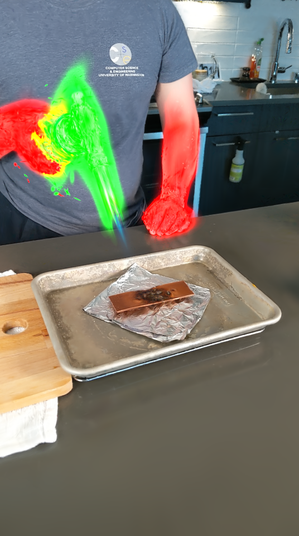} & 
\includegraphics[width=0.17\linewidth, trim={0 5.025cm 0 0},clip]{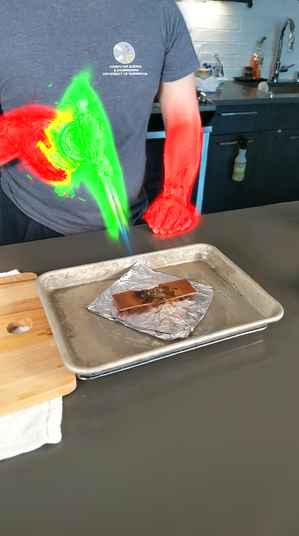} & 
\includegraphics[width=0.17\linewidth, trim={0 5.025cm 0 0},clip]{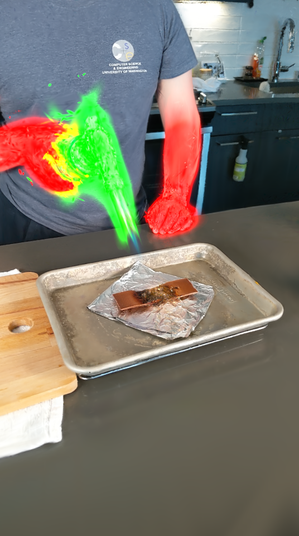} & 
\includegraphics[width=0.17\linewidth, trim={0 5.025cm 0 0},clip]{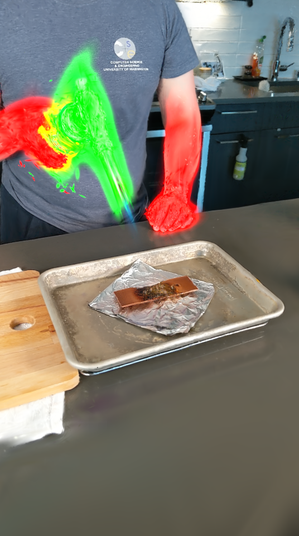} \\

\rotatebox{90}{\hspace{0.5cm}\tiny{Rendered View}}
\includegraphics[width=0.17\linewidth, trim={0 5.025cm 0 0},clip]{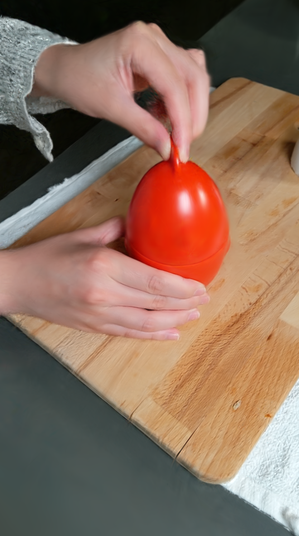} & 
\includegraphics[width=0.17\linewidth, trim={0 5.025cm 0 0},clip]{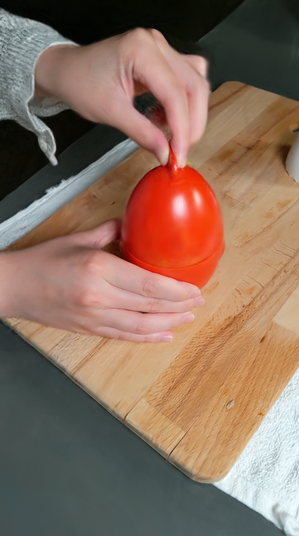} & 
\includegraphics[width=0.17\linewidth, trim={0 5.025cm 0 0},clip]{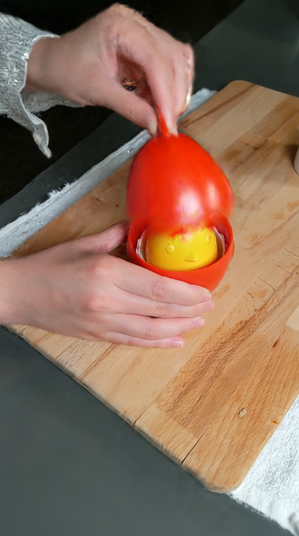} & 
\includegraphics[width=0.17\linewidth, trim={0 5.025cm 0 0},clip]{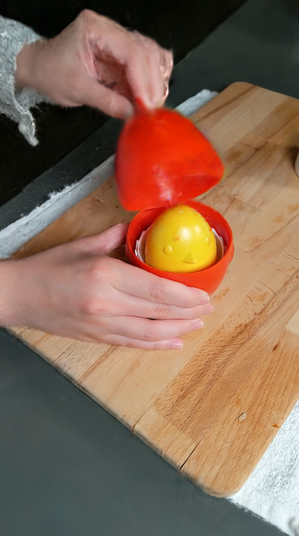} & 
\includegraphics[width=0.17\linewidth, trim={0 5.025cm 0 0},clip]{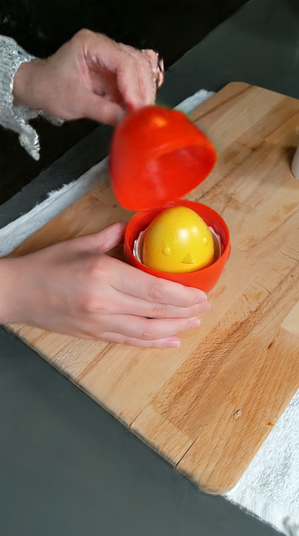} \\
\rotatebox{90}{\hspace{0.5cm}\tiny{Segmentation}}
\includegraphics[width=0.17\linewidth, trim={0 5.025cm 0 0},clip]{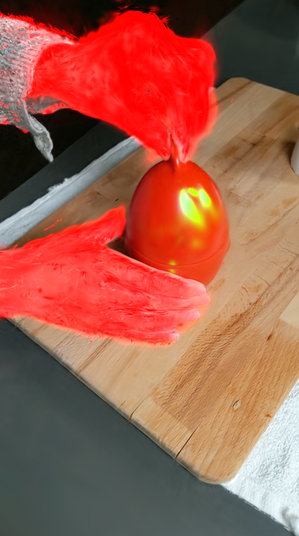} & 
\includegraphics[width=0.17\linewidth, trim={0 5.025cm 0 0},clip]{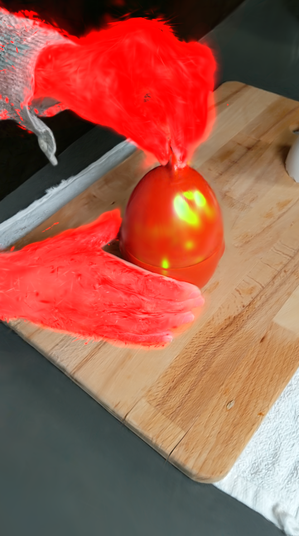} & 
\includegraphics[width=0.17\linewidth, trim={0 5.025cm 0 0},clip]{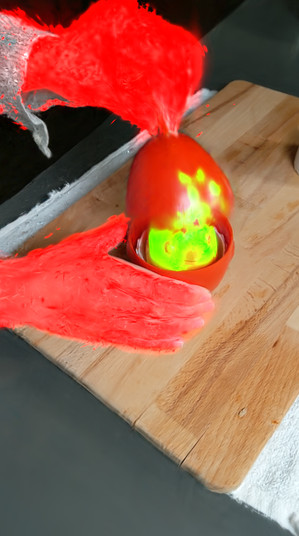} & 
\includegraphics[width=0.17\linewidth, trim={0 5.025cm 0 0},clip]{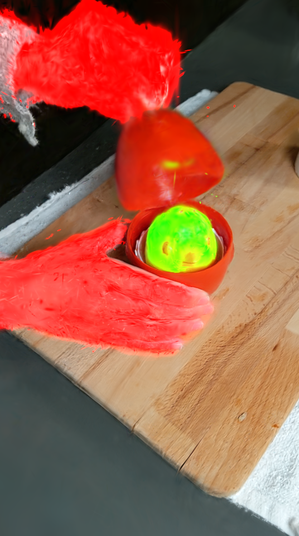} & 
\includegraphics[width=0.17\linewidth, trim={0 5.025cm 0 0},clip]{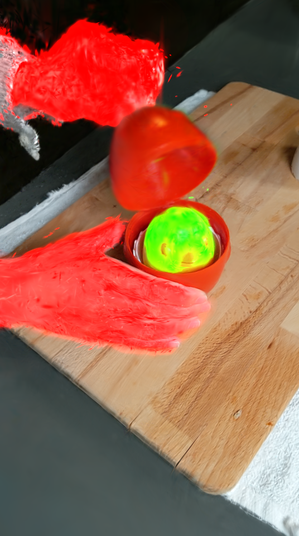} \\

\rotatebox{90}{\hspace{0.5cm}\tiny{Rendered View}}
\includegraphics[width=0.17\linewidth, trim={0 5.025cm 0 0},clip]{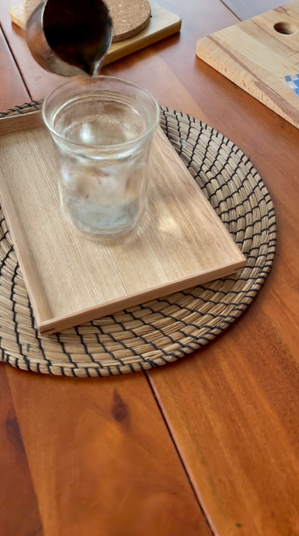} & 
\includegraphics[width=0.17\linewidth, trim={0 5.025cm 0 0},clip]{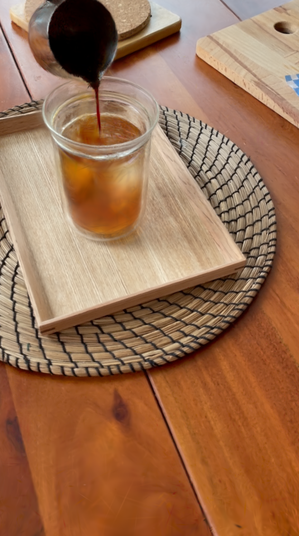} & 
\includegraphics[width=0.17\linewidth, trim={0 5.025cm 0 0},clip]{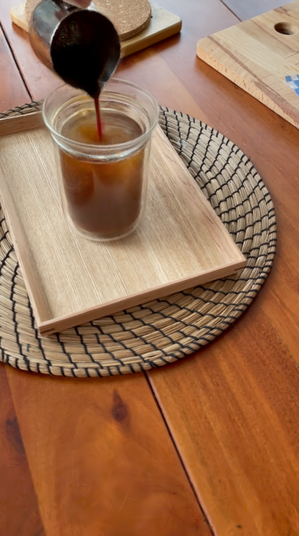} & 
\includegraphics[width=0.17\linewidth, trim={0 5.025cm 0 0},clip]{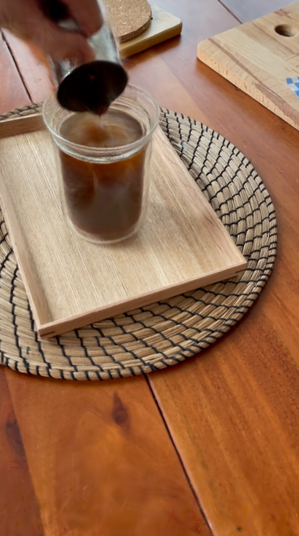} & 
\includegraphics[width=0.17\linewidth, trim={0 5.025cm 0 0},clip]{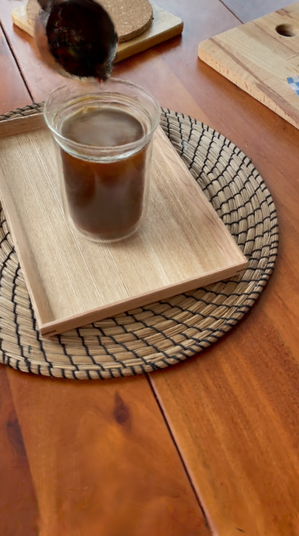} \\
\rotatebox{90}{\hspace{0.5cm}\tiny{Segmentation}}
\includegraphics[width=0.17\linewidth, trim={0 5.025cm 0 0},clip]{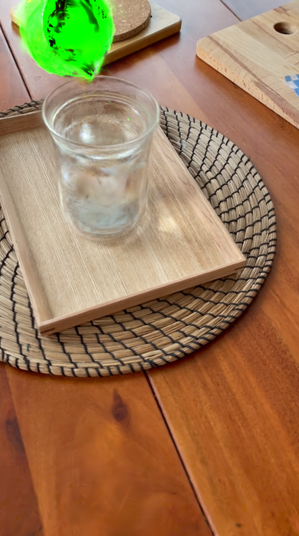} & 
\includegraphics[width=0.17\linewidth, trim={0 5.025cm 0 0},clip]{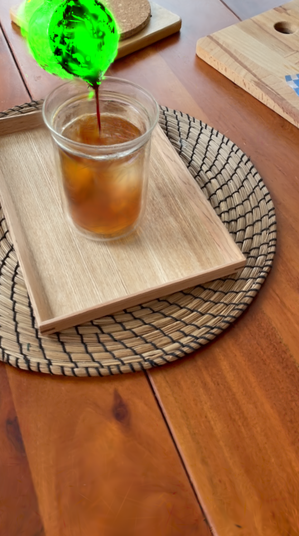} & 
\includegraphics[width=0.17\linewidth, trim={0 5.025cm 0 0},clip]{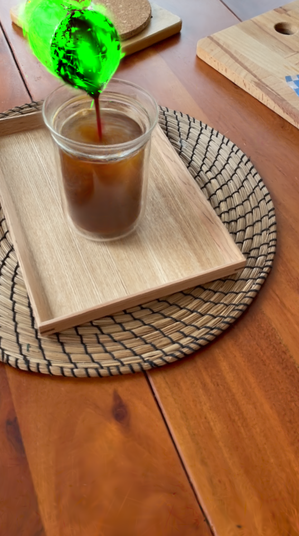} & 
\includegraphics[width=0.17\linewidth, trim={0 5.025cm 0 0},clip]{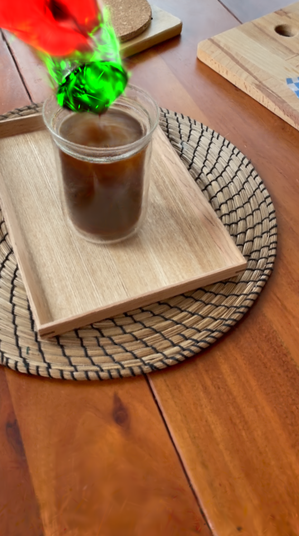} & 
\includegraphics[width=0.17\linewidth, trim={0 5.025cm 0 0},clip]{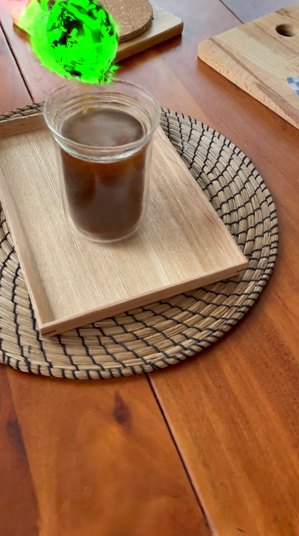} \\

\end{tabular}
\caption{Visual illustration of our method's ability to semantically segment and track objects over time, accompanying \cref{fig:teaser}, for real world scenes of the HyperNeRF dataset. We consider a random novel view for each scene over time. 
}
\label{fig:tracking}
\end{figure*}

\begin{figure*}
\centering
\begin{tabular}{ccccc}

$t=0$ & $t=200$ & $t=400$ & $t=600$ & $t=800$ \\
\rotatebox{90}{\hspace{0.3cm}\tiny{Rendered View}}
\includegraphics[width=0.165\linewidth]{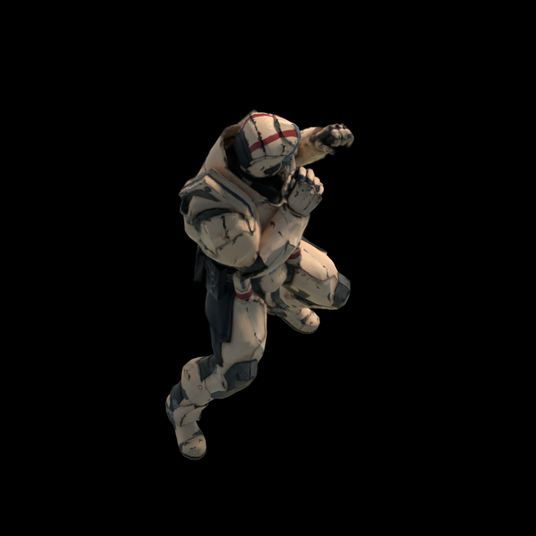} & 
\includegraphics[width=0.165\linewidth]{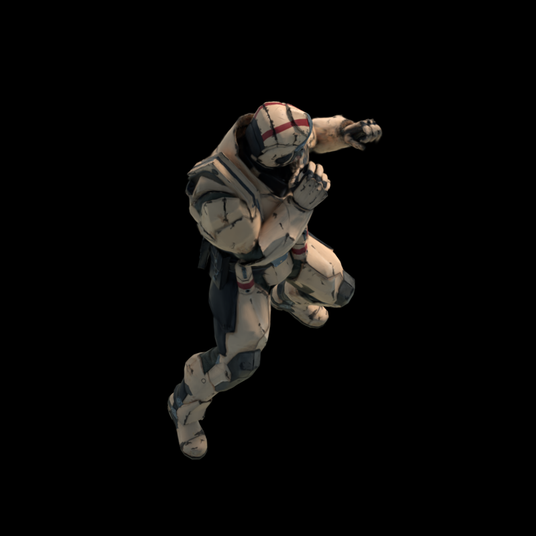} & 
\includegraphics[width=0.165\linewidth]{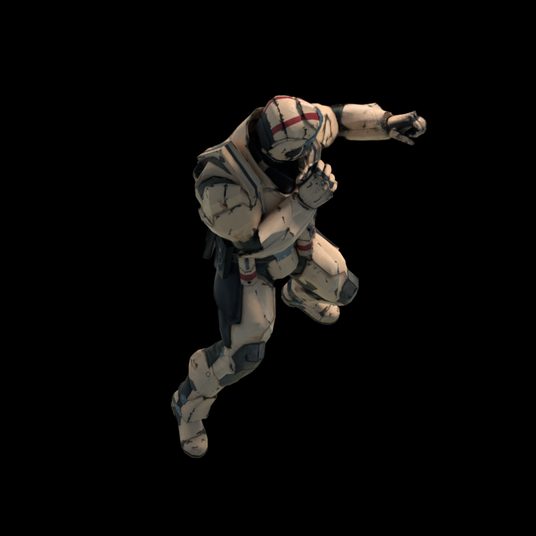} & 
\includegraphics[width=0.165\linewidth]{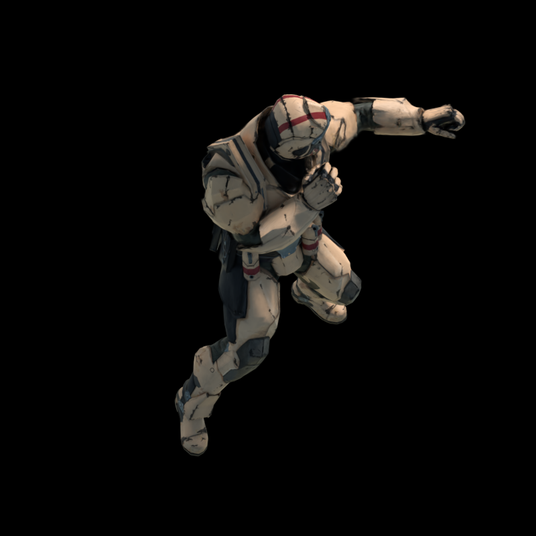} & 
\includegraphics[width=0.165\linewidth]{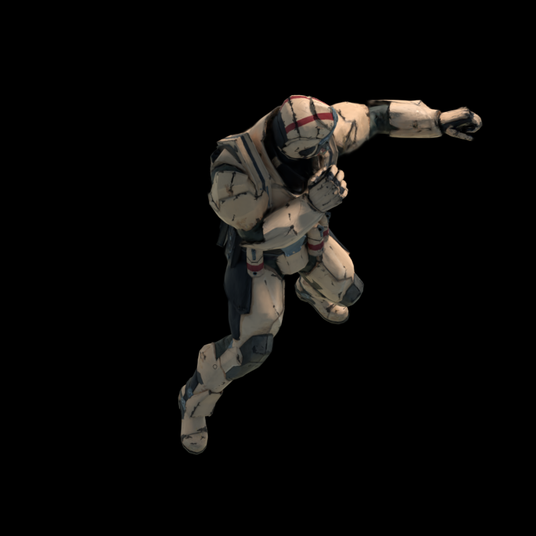} \\
\rotatebox{90}{\hspace{0.3cm}\tiny{Segmentation}}
\includegraphics[width=0.165\linewidth]{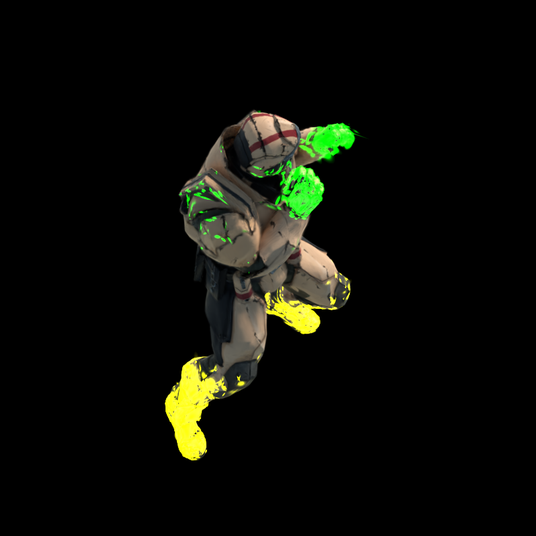} & 
\includegraphics[width=0.165\linewidth]{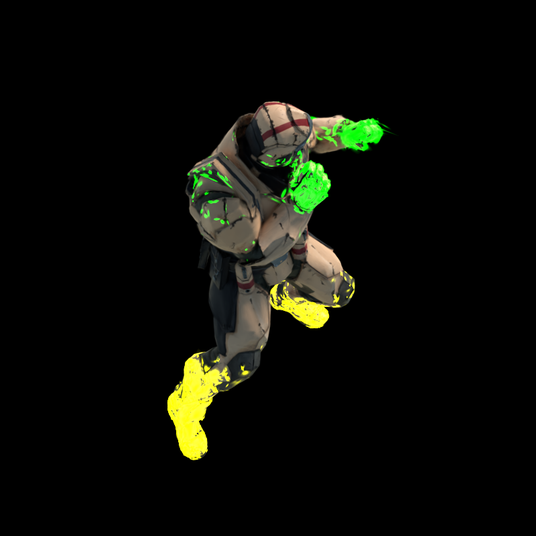} & 
\includegraphics[width=0.165\linewidth]{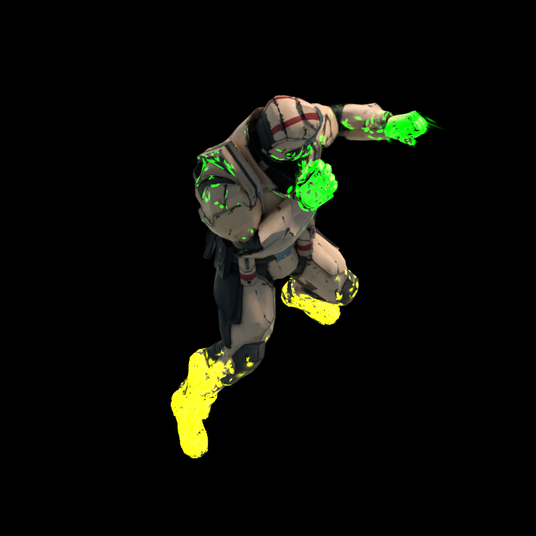} & 
\includegraphics[width=0.165\linewidth]{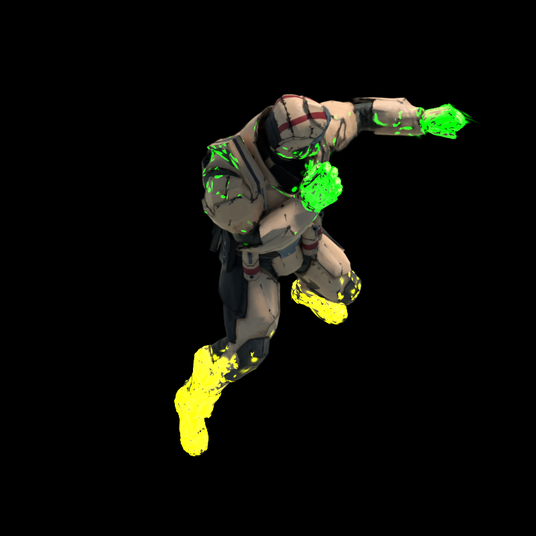} & 
\includegraphics[width=0.165\linewidth]{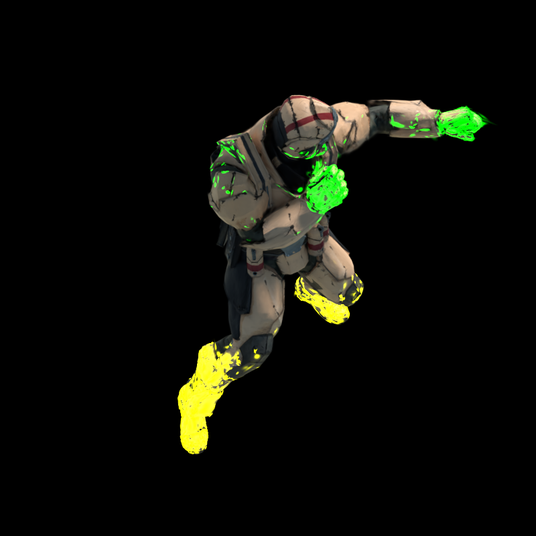} \\

\rotatebox{90}{\hspace{0.3cm}\tiny{Rendered View}}
\includegraphics[width=0.165\linewidth]{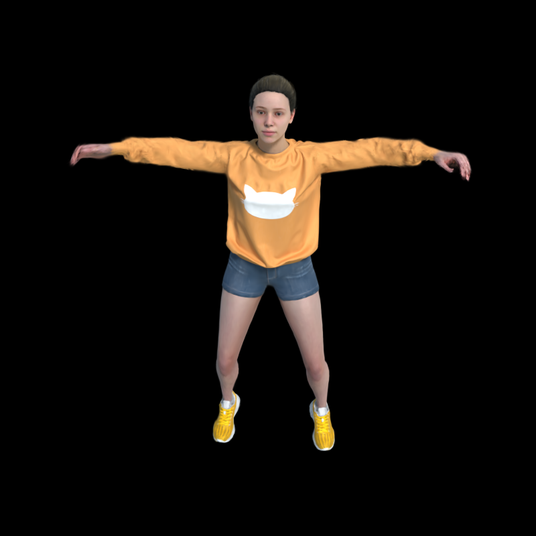} & 
\includegraphics[width=0.165\linewidth]{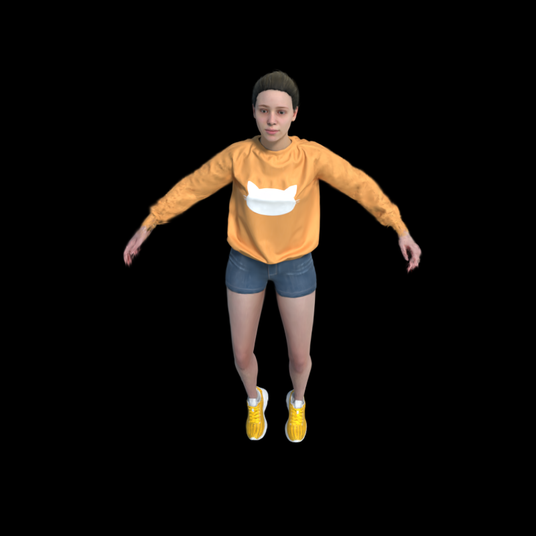} & 
\includegraphics[width=0.165\linewidth]{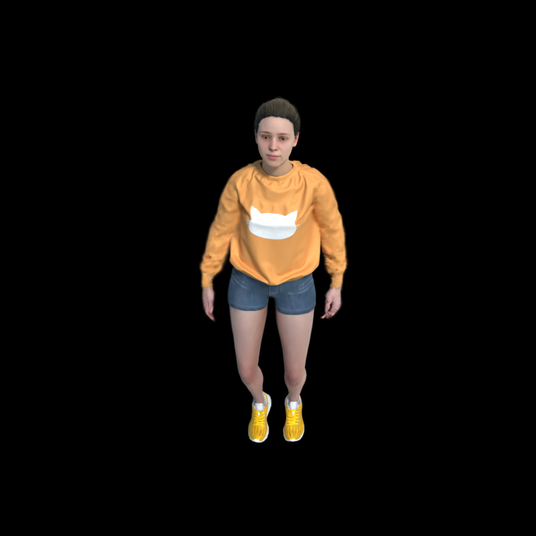} & 
\includegraphics[width=0.165\linewidth]{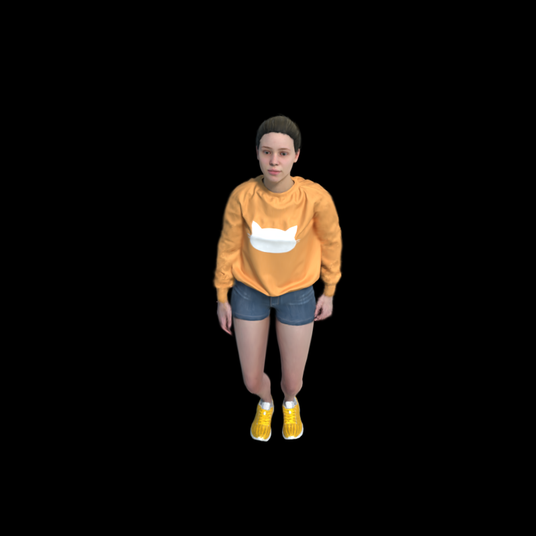} & 
\includegraphics[width=0.165\linewidth]{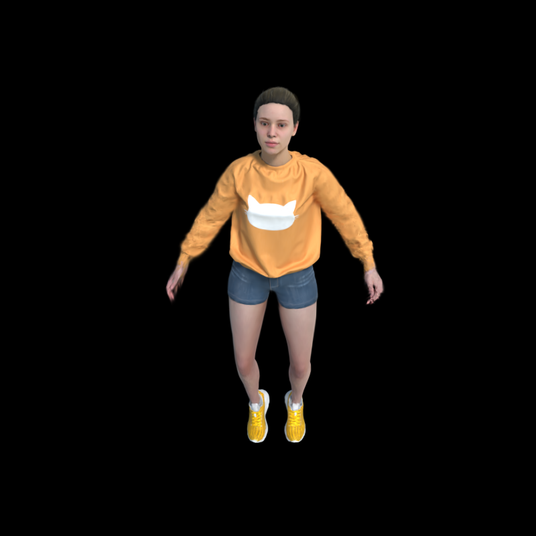} \\
\rotatebox{90}{\hspace{0.3cm}\tiny{Segmentation}}
\includegraphics[width=0.165\linewidth]{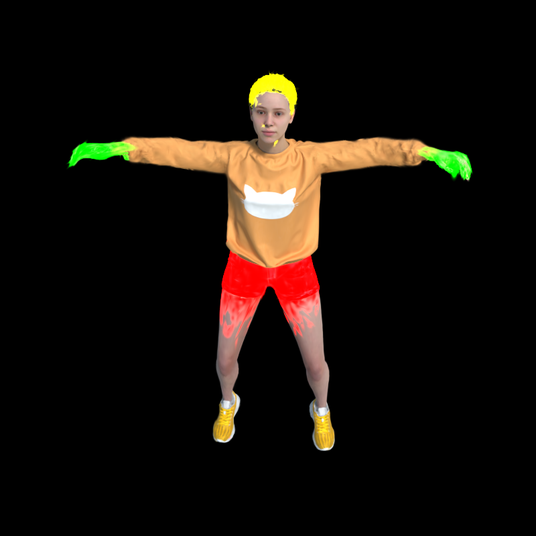} & 
\includegraphics[width=0.165\linewidth]{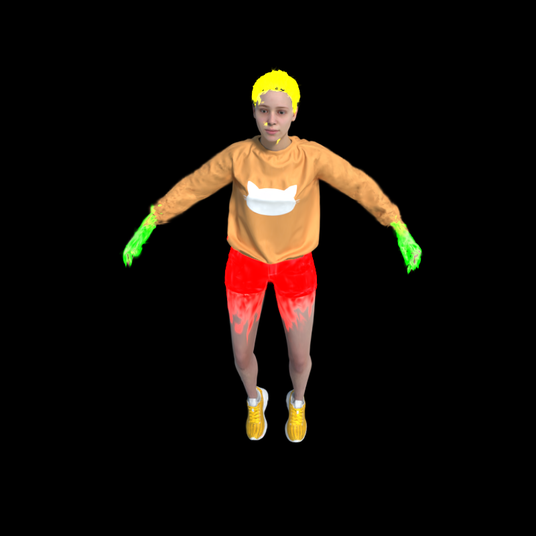} & 
\includegraphics[width=0.165\linewidth]{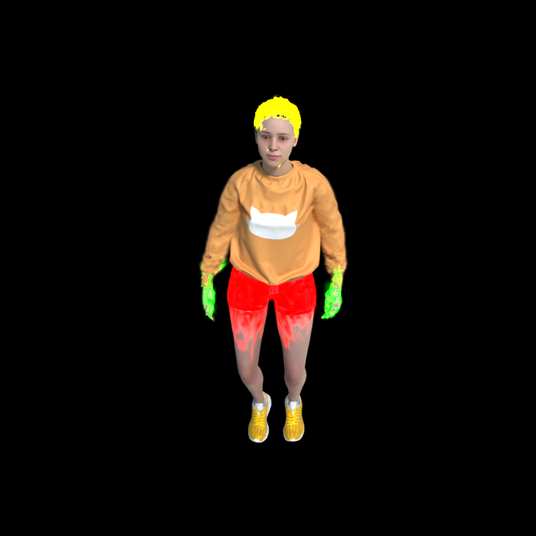} & 
\includegraphics[width=0.165\linewidth]{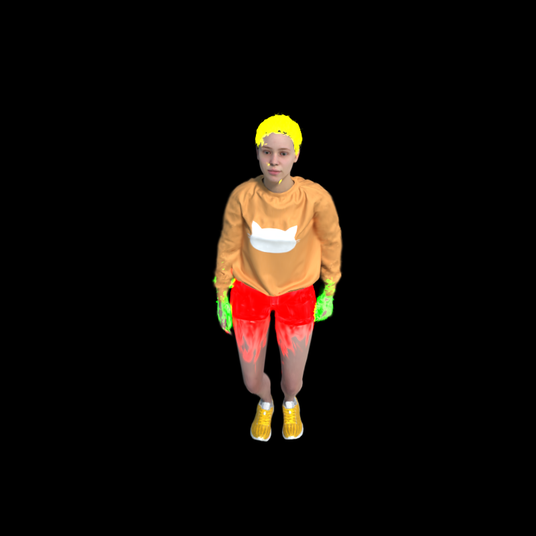} & 
\includegraphics[width=0.165\linewidth]{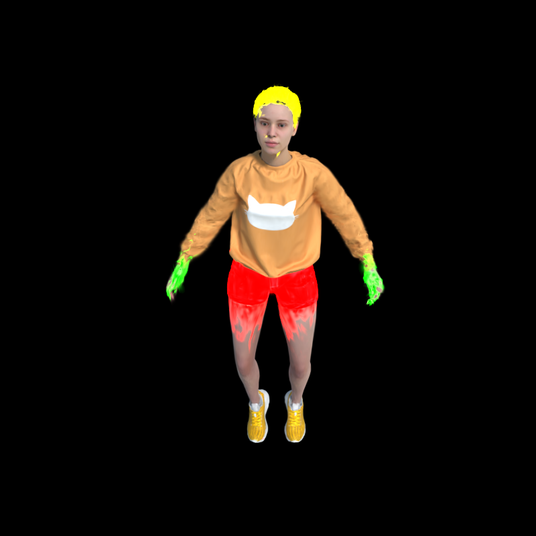} \\ 

\rotatebox{90}{\hspace{0.3cm}\tiny{Rendered View}}
\includegraphics[width=0.165\linewidth]{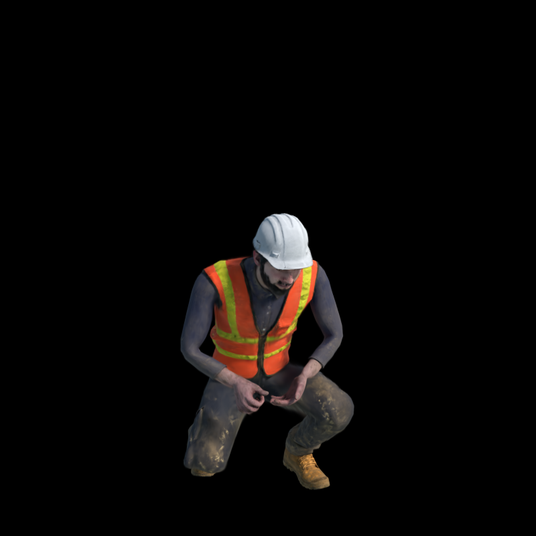} & 
\includegraphics[width=0.165\linewidth]{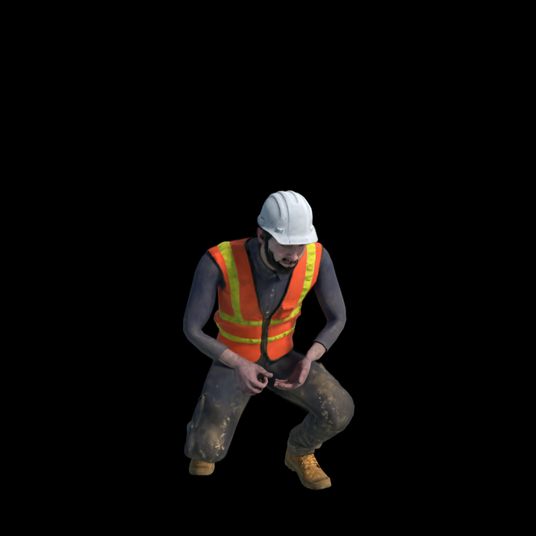} & 
\includegraphics[width=0.165\linewidth]{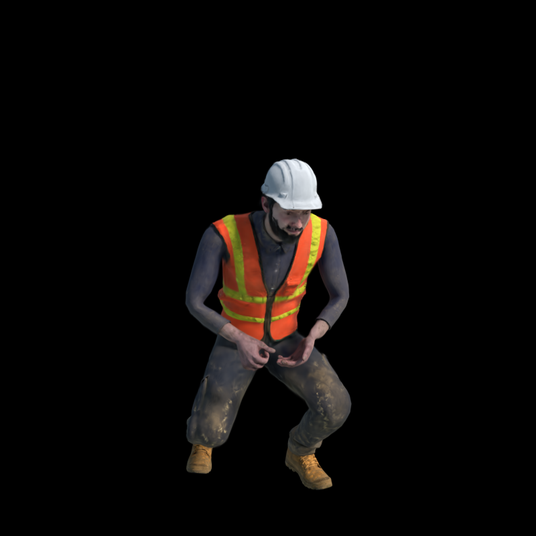} & 
\includegraphics[width=0.165\linewidth]{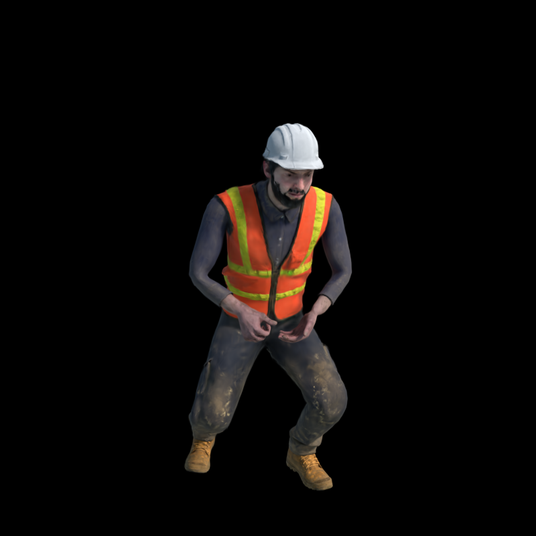} & 
\includegraphics[width=0.165\linewidth]{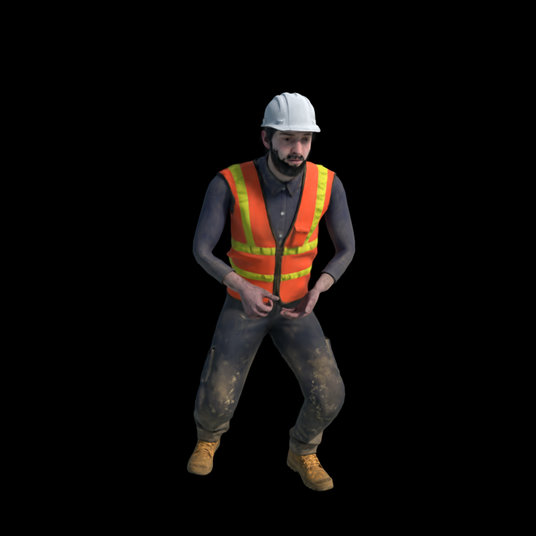} \\
\rotatebox{90}{\hspace{0.3cm}\tiny{Segmentation}}
\includegraphics[width=0.165\linewidth]{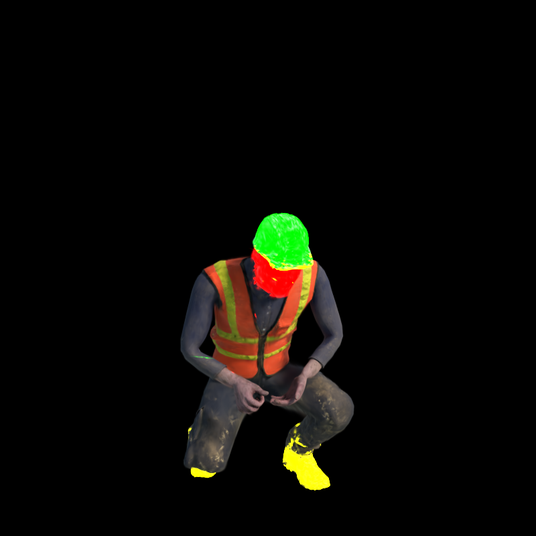} & 
\includegraphics[width=0.165\linewidth]{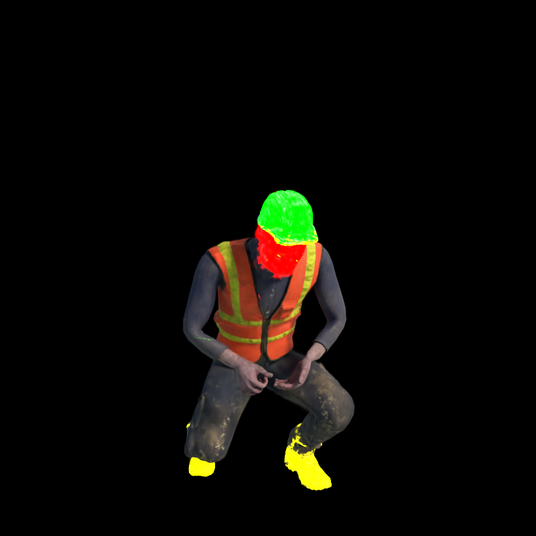} & 
\includegraphics[width=0.165\linewidth]{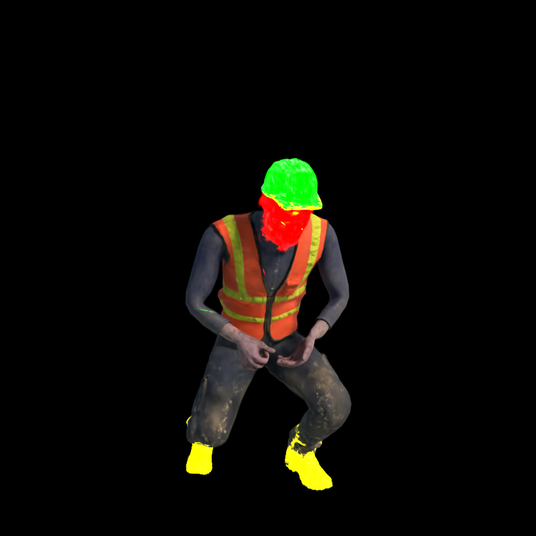} & 
\includegraphics[width=0.165\linewidth]{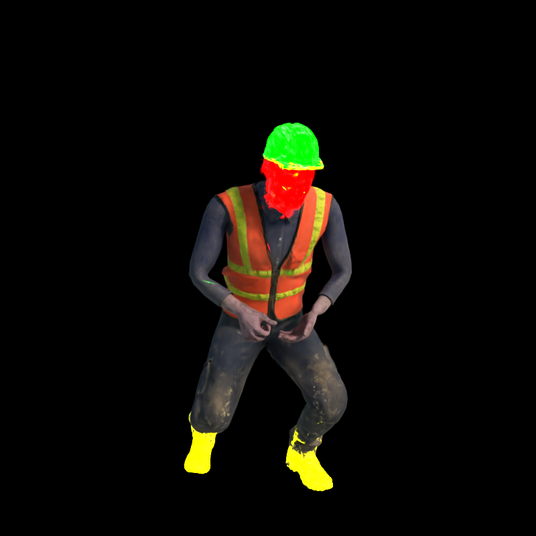} & 
\includegraphics[width=0.165\linewidth]{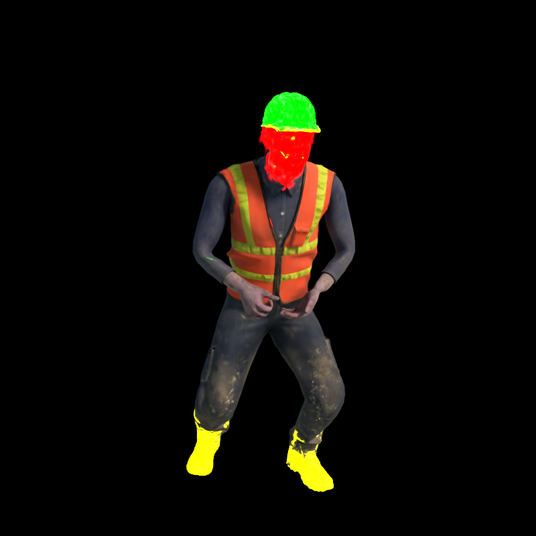} \\

\rotatebox{90}{\hspace{0.3cm}\tiny{Rendered View}}
\includegraphics[width=0.165\linewidth]{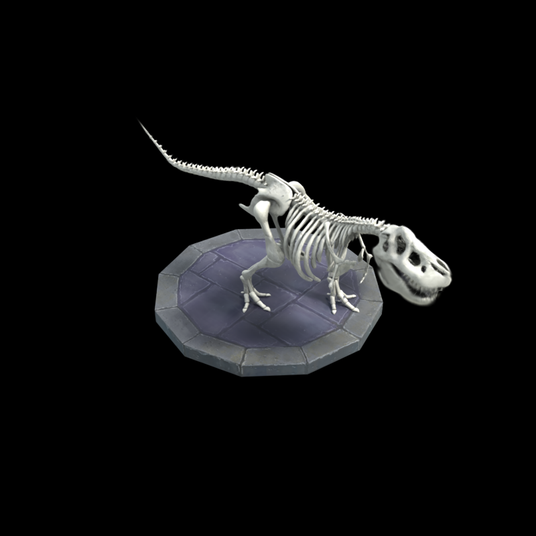} & 
\includegraphics[width=0.165\linewidth]{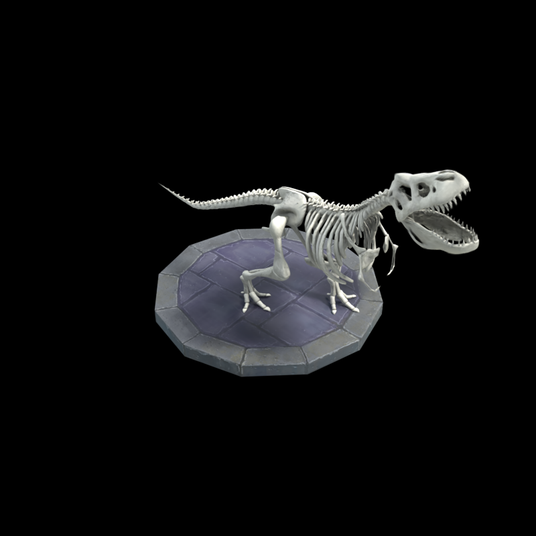} & 
\includegraphics[width=0.165\linewidth]{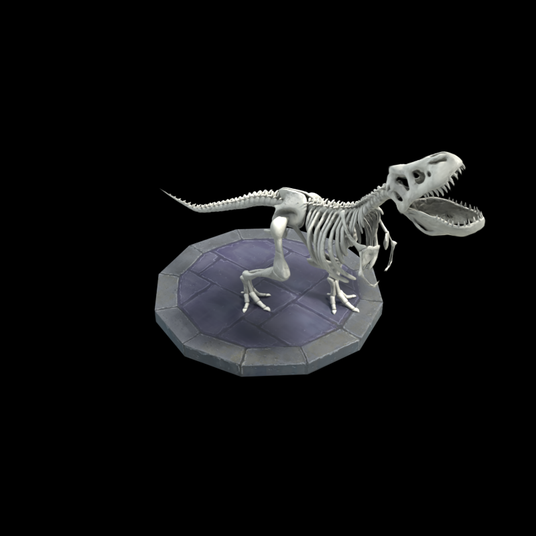} & 
\includegraphics[width=0.165\linewidth]{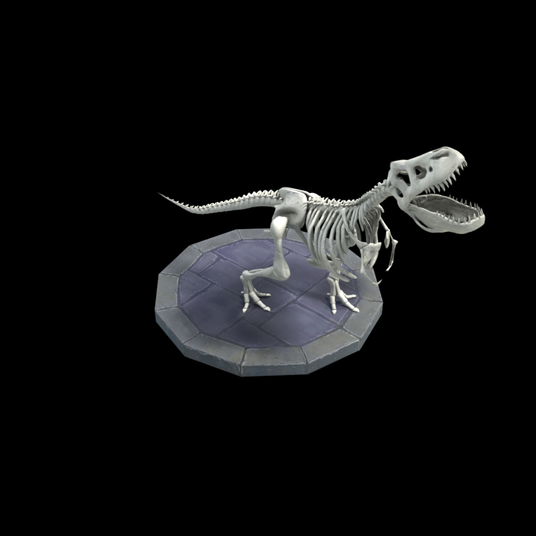} & 
\includegraphics[width=0.165\linewidth]{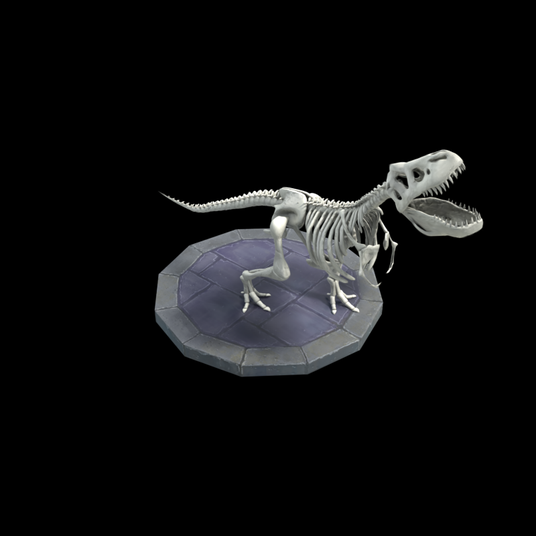} \\
\rotatebox{90}{\hspace{0.3cm}\tiny{Segmentation}}
\includegraphics[width=0.165\linewidth]{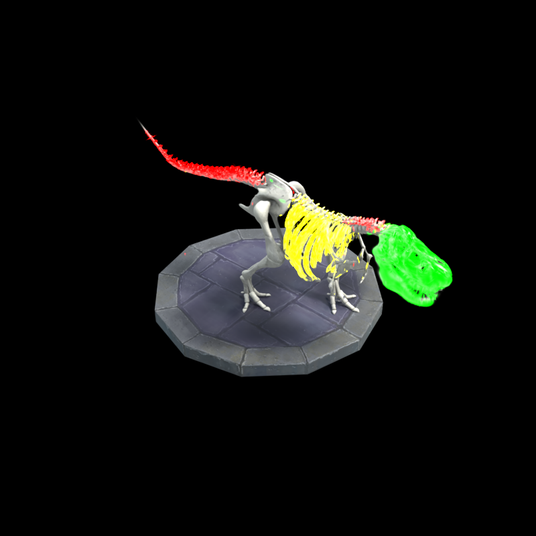} & 
\includegraphics[width=0.165\linewidth]{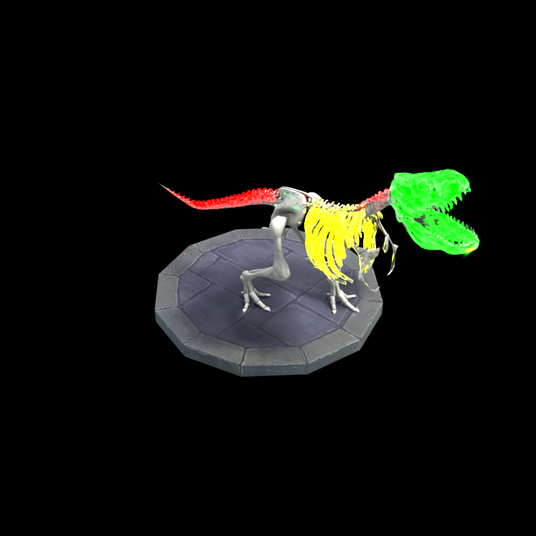} & 
\includegraphics[width=0.165\linewidth]{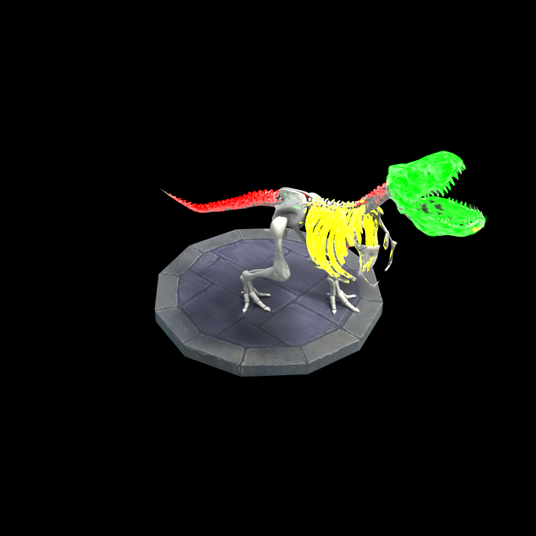} & 
\includegraphics[width=0.165\linewidth]{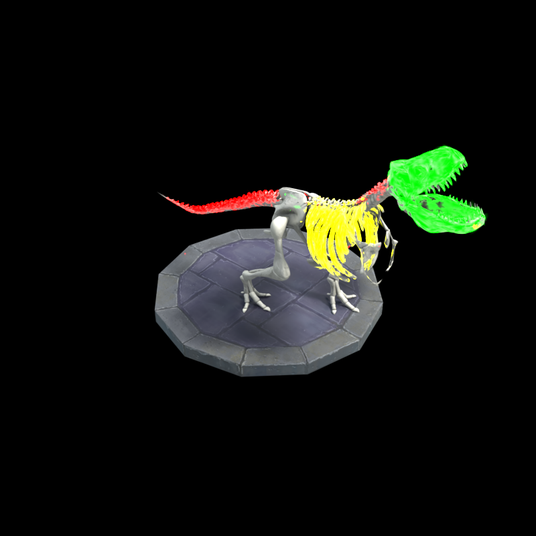} & 
\includegraphics[width=0.165\linewidth]{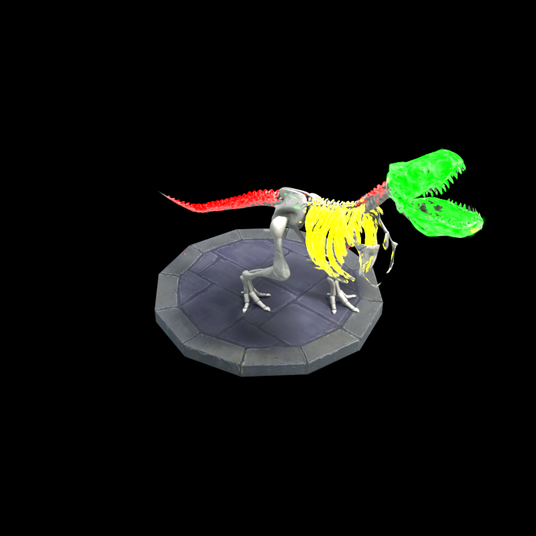} \\ 

\end{tabular}
\caption{Visual illustration of our method's ability to semantically segment and track objects over time as in \cref{fig:tracking_synthetic}, but for synthetic scenes on D-NeRF synthetic dataset. 
}
\label{fig:tracking_synthetic}
\end{figure*}

\subsection{Implementation Details}

We use PyTorch for our implementation. 
Training is performed for 40K iterations, with a warmup of 3K steps used only for training the static 3D Gaussians (i.e freezing $D$ to the identity deformation). 
After 3K iterations, we jointly train 3D Gaussians parameters, as well as associated colors and features, and the deformation field $D$.
We use Adam for optimization. 
For $D$, we use an LR with exponential decay, with a start of $8e^{-4}$ and end of $1.6e^{-6}$. The $\beta$ are set to $(0.9, 0.999)$. To train the defromartion field $D$, we make use of the annealing smooth training (AST) mechanism introduced in \cite{yang2023deformable}. 
For all our experiments, for any given scene optimization, we use a single Nvidia A100 GPU.

\section{Experiments} \label{sec:experiments}
\label{sec:semantic_tracking}

\begin{wraptable}{r}{5.72cm}
\centering
\vspace{-0.7cm}
\begin{tabular}{lcc}
\toprule
& LSeg & Ours  \\
\midrule
Split-cookie (Cookie) & 76.09 & \textbf{78.65}   \\
Americano (Hands) & 13.90 & \textbf{75.98}  \\
Torchocolate (Hands) & 13.90 & \textbf{75.98}  \\
Chickchicken (Hands) & 22.92 & \textbf{77.92}  \\
\midrule
Standup (Helmet) & 1.08 & \textbf{93.80}  \\
Jumpingjack (Hands) & 0.89 & \textbf{68.35}  \\
Trex (Skull) & 2.48 & \textbf{70.39}  \\
\bottomrule
\end{tabular}
\caption{Mean IoU for object segmentation of objects from the HyperNeRF~\cite{park2021hypernerf} dataset (top four scenes) and synthetic D-NeRF~\cite{pumarola2021d} dataset (bottom three scenes), in comparison to the 2D baseline of LSeg~\cite{li2022languagedriven}. For each scene, we show in brackets the object to be segmented. 
}

\label{tab:iou}
\vspace{-0.5cm}
\end{wraptable}

We evaluate our method on the task of 3D dense segmentation and tracking of semantic objects through time, where the input supervision is of a single monocular video, and where the object to be tracked is specified using a click or a text prompt. 
Full videos for our method and baselines, depicting novel views through time are provided in the project webpage.

We consider the following baselines:
(1). A 2D baseline of LSeg~\cite{li2022languagedriven}, which is applied on training views, (2).
A concurrent work of \cite{zhou2023feature3dgs}, performing 3D segmentation in static scenes. As we are only given a monocular video for training, and \cite{zhou2023feature3dgs} requires a collection of static views for training, we treat the views from the video as static views and train  \cite{zhou2023feature3dgs} 
on these views to serve as the baseline, 
so as to provide a fair comparison. 
\begin{wraptable}{r}{5.72cm}
\centering
\vspace{-0.7cm}
\begin{tabular}{lcc}
\toprule
& Baseline & Ours  \\
\midrule
3D Segmentation - Q1 & 2.71 & \textbf{4.63} \\
3D Segmentation - Q2 & 3.4  & \textbf{4.63} \\
\bottomrule
\end{tabular}
\caption{\textbf{Perceptual user study}
for 3D semantic Segmentation (Q1 and Q2). }
\label{tab:user_study}
\vspace{-0.5cm}
\end{wraptable}

We consider the following datasets for our evaluation: 
(i). The synthetic dataset of D-NeRF~\cite{pumarola2021d} which consists of eight scenes. (ii). The real world dataset of HyperNeRF~\cite{park2021hypernerf} which consists of seventeen real world scenes. 
For (i) and (ii), as no ground truth segmentation is provided, we consider training views and apply manual annotation. In particular, we begin by applying Segment Anything~\cite{kirillov2023segment} on individual frames and then refine the provided segmentation manually. See \cref{fig:annotations} for visual illustration.

\cref{fig:tracking} and \cref{fig:tracking_synthetic}  provide a visual illustration of our results for the HyperNerf real-world dataset (ii) and the D-NeRF synthetic dataset (i), obtained using 3D user clicks and DINOv2~\cite{oquab2023dinov2} features as detailed in \cref{sec:method}. 
For \cref{fig:tracking}, in all scenes, we segment the ``hands" (red). For the first scene, we also segment the ``torch'' (green). For the second scene, we also segment the ``egg" (green). For the third scene, we also segment the ``coffee jar'' (green).  For \cref{fig:tracking_synthetic}, for the first scene we segment the ``hands" (green) and ``legs'' (yellow). For the second scene we segment the ``hands" (green), ``pants'' (red) and ``hair" (yellow). The the third 
\begin{wrapfigure}{r}{0.55\textwidth} 
\vspace{-0.2cm}
\begin{tabular}{cccc}
\tiny{view 1} & \tiny{view 2} & \tiny{view 3} & \tiny{view 4} \\ 

\rotatebox{90}{\hspace{0.3cm}\tiny{ Ours}}
\includegraphics[width=0.20\linewidth, height=0.20\linewidth]{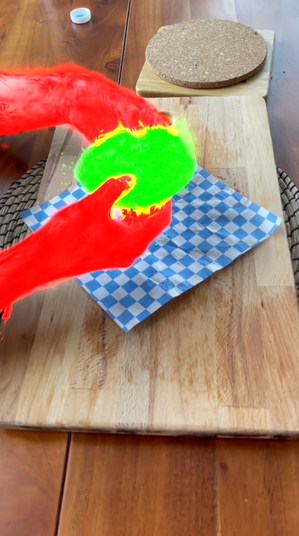} & 
\includegraphics[width=0.20\linewidth, height=0.20\linewidth]{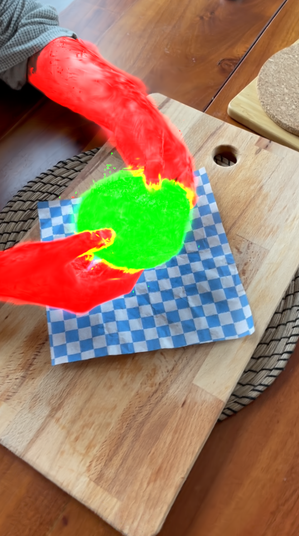} &
\includegraphics[width=0.20\linewidth, height=0.20\linewidth]{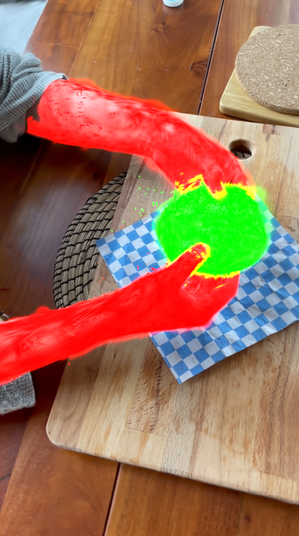} & 
\includegraphics[width=0.20\linewidth, height=0.20\linewidth]{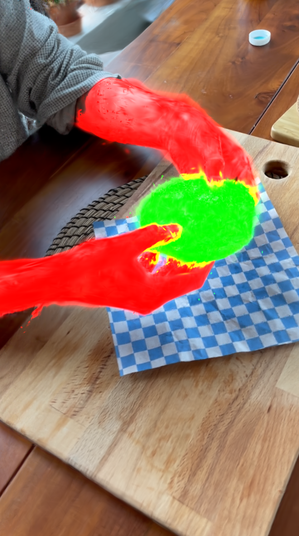} \\
\rotatebox{90}{\hspace{0.1cm} \tiny{Baseline}}
\includegraphics[width=0.20\linewidth, height=0.20\linewidth]{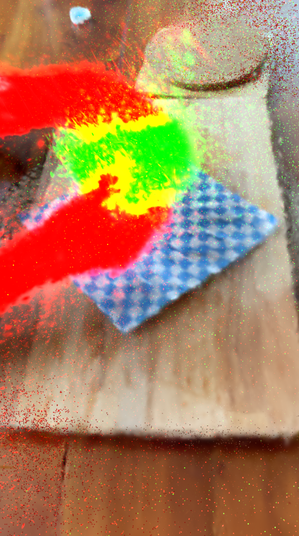} & 
\includegraphics[width=0.20\linewidth, height=0.20\linewidth]{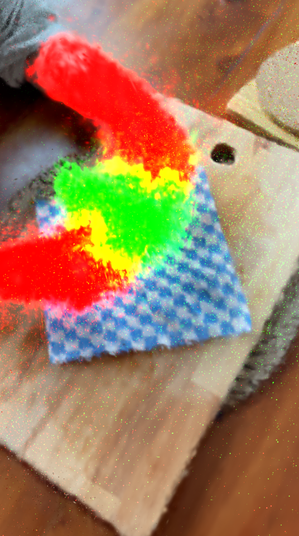}  & 
\includegraphics[width=0.20\linewidth, height=0.20\linewidth]{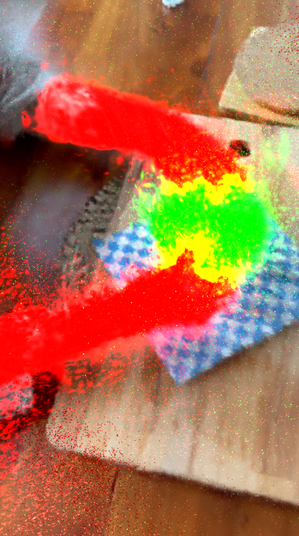} & 
\includegraphics[width=0.20\linewidth, height=0.20\linewidth]{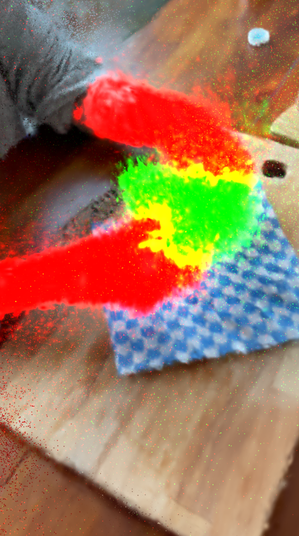} \\

\end{tabular}
\caption{Comparison of our method on 3D segmentation. 
We consider our segmentation for the canonical frame for 4 different random views for 
``cookie" (green) and ``hands" (red). For the baseline, we consider the 3D static semantic segmentation of \cite{zhou2023feature3dgs}. 
}
\label{fig:tracking_visual_comparison}
\vspace{-0.8cm}
\end{wrapfigure} 

scene we segment the ``helment'' (green), ``face" (red) and ``legs" (yellow). For the fourth scene, we segment the ``tail" (red), ``skull" (green) and ``rib cage" (yellow).

As can be seen, given only a single monocular video as input, our method correctly segments 3D regions over time (displayed for some random view). 
Additional novel views are provided in \cref{fig:tracking_extra_views} and \cref{fig:tracking_synthetic_extra_views}.

\begin{figure*}[t]
\centering
\begin{tabular}{cccc}
Input & Ground Truth & Ours & LSeg~\cite{li2022languagedriven}  \\
\rotatebox{90}{\hspace{0.2cm}\tiny{Split-cookie (Cookie)}}
\includegraphics[width=0.22\linewidth, height=0.22\linewidth]{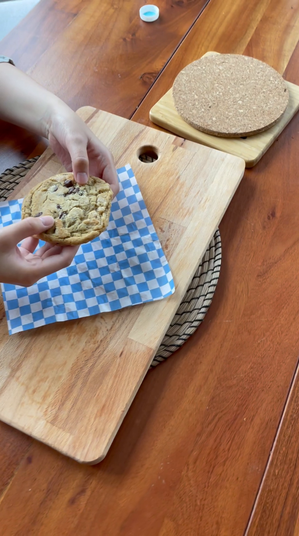} & 
\includegraphics[width=0.22\linewidth, height=0.22\linewidth]{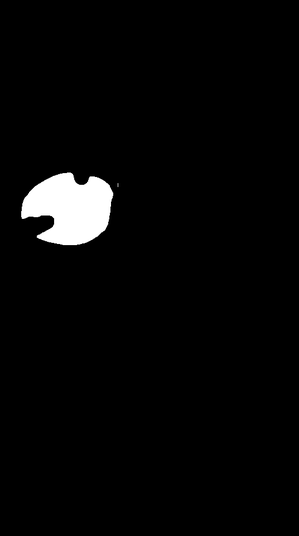} & 
\includegraphics[width=0.22\linewidth, height=0.22\linewidth]{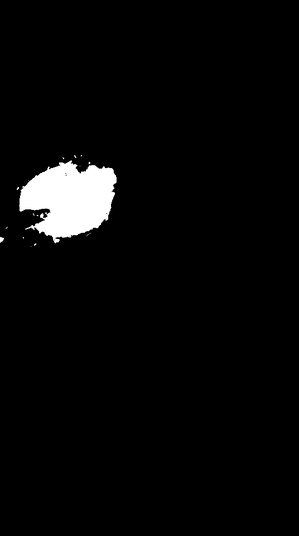} & 
\includegraphics[width=0.22\linewidth, height=0.22\linewidth]{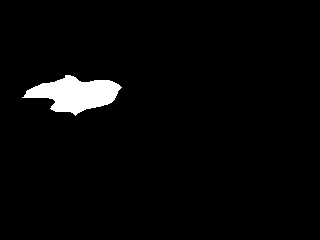} \\  
\rotatebox{90}{\hspace{0.4cm}\tiny{Americano (Hands)}}
\includegraphics[width=0.22\linewidth, height=0.22\linewidth]{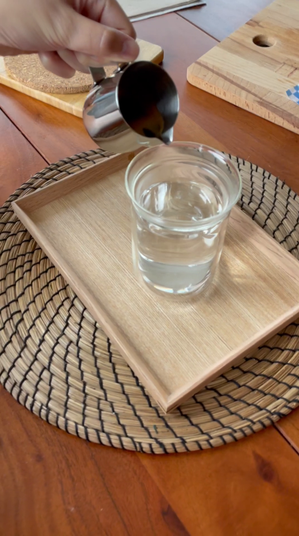} & 
\includegraphics[width=0.22\linewidth, height=0.22\linewidth]{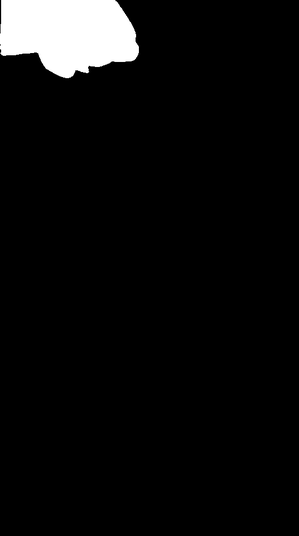} & 
\includegraphics[width=0.22\linewidth, height=0.22\linewidth]{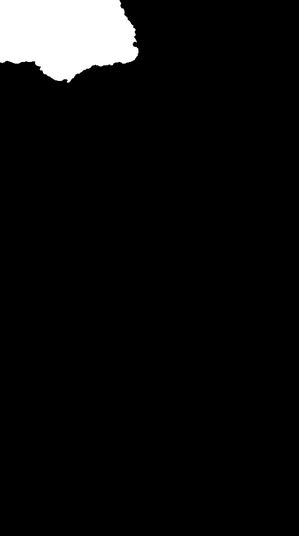} & 
\includegraphics[width=0.22\linewidth, height=0.22\linewidth]{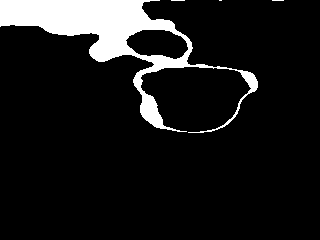} \\  
\rotatebox{90}{\hspace{0.4cm}\tiny{Torchocolate (Hands)}}
\includegraphics[width=0.22\linewidth, height=0.22\linewidth]{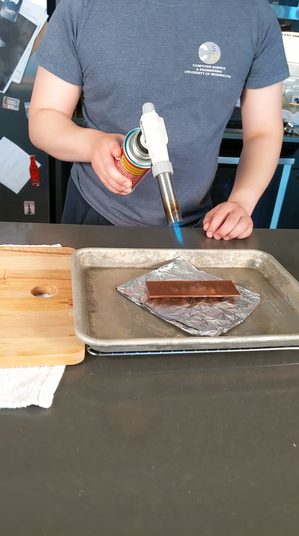} & 
\includegraphics[width=0.22\linewidth, height=0.22\linewidth]{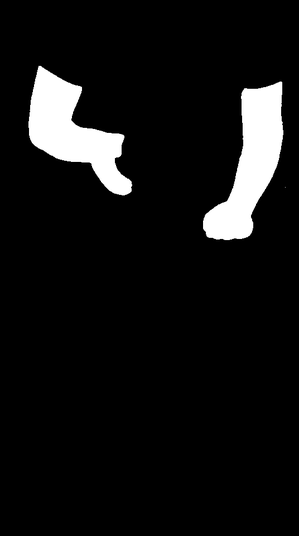} & 
\includegraphics[width=0.22\linewidth, height=0.22\linewidth]{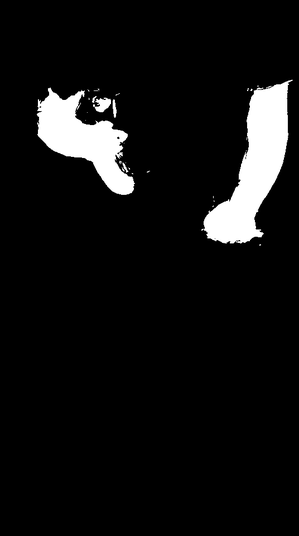} & 
\includegraphics[width=0.22\linewidth, height=0.22\linewidth]{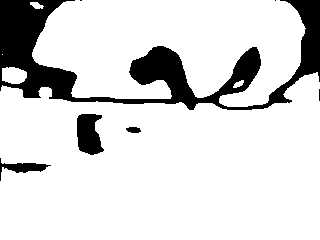} \\  
\rotatebox{90}{\hspace{0.1cm}\tiny{Chickchicken (Hands)}}
\includegraphics[width=0.22\linewidth, height=0.22\linewidth]{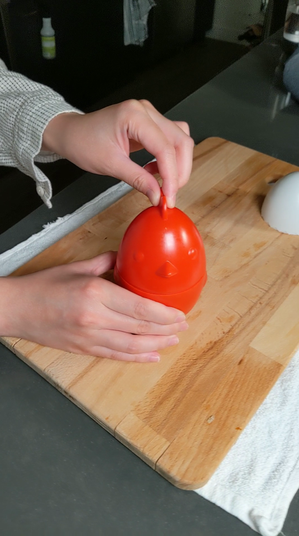} & 
\includegraphics[width=0.22\linewidth, height=0.22\linewidth]{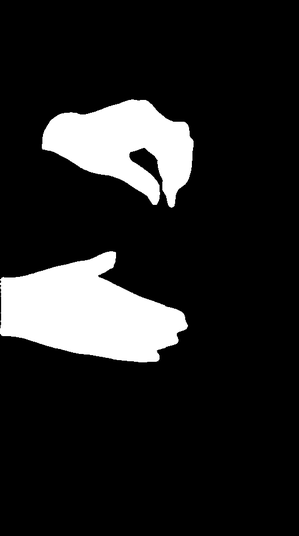} & 
\includegraphics[width=0.22\linewidth, height=0.22\linewidth]{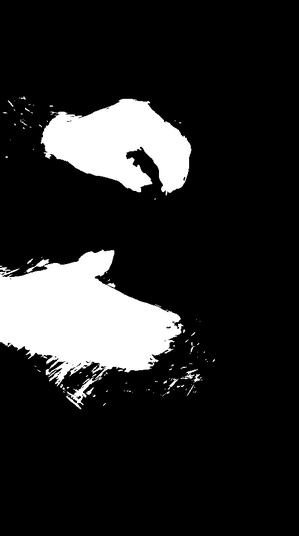} & 
\includegraphics[width=0.22\linewidth, height=0.22\linewidth]{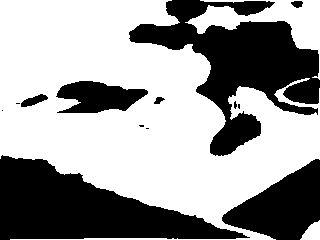} \\  
\end{tabular}
\caption{Illustration of segmentation masks obtained for a single view and timestep for  objects from the HyperNerf~\cite{park2021hypernerf} dataset, in comparison to the 2D baseline of LSeg~\cite{li2022languagedriven}. For each scene, we show in brackets the object to be segmented. 
}
\vspace{-0.7cm}
\label{fig:tracking_lseg}
\end{figure*}

\begin{figure*}[t]
\centering
\begin{tabular}{ccccc}

$t=0$ & $t=200$ & $t=400$ & $t=600$ & $t=800$ \\
\rotatebox{90}{\hspace{0.4cm}\tiny{Rendered View}}
\includegraphics[width=0.18\linewidth, height=0.18\linewidth]{figures/tracking/cookie/renders_cookie_view1/00200.png} & 
\includegraphics[width=0.18\linewidth, height=0.18\linewidth]{figures/tracking/cookie/renders_cookie_view1/00400.png} & 
\includegraphics[width=0.18\linewidth, height=0.18\linewidth]{figures/tracking/cookie/renders_cookie_view1/00600.png} & 
\includegraphics[width=0.18\linewidth, height=0.18\linewidth]{figures/tracking/cookie/renders_cookie_view1/00800.png} & 
\includegraphics[width=0.18\linewidth, height=0.18\linewidth]{figures/tracking/cookie/renders_cookie_view1/00999.png} \\
\rotatebox{90}{\hspace{0.8cm}\tiny{Editing}}
\includegraphics[width=0.18\linewidth, height=0.18\linewidth]{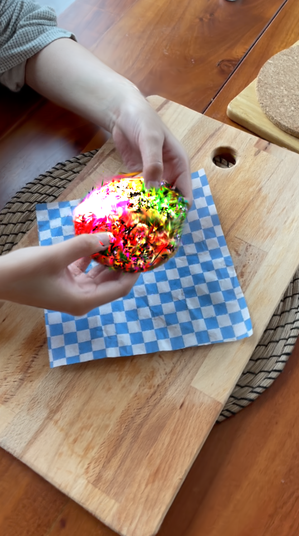} & 
\includegraphics[width=0.18\linewidth, height=0.18\linewidth]{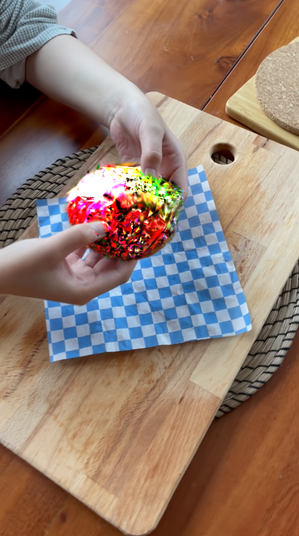} & 
\includegraphics[width=0.18\linewidth, height=0.18\linewidth]{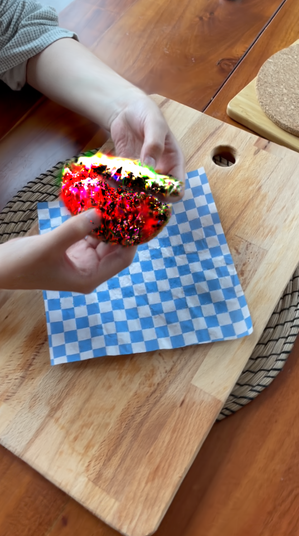} & 
\includegraphics[width=0.18\linewidth, height=0.18\linewidth]{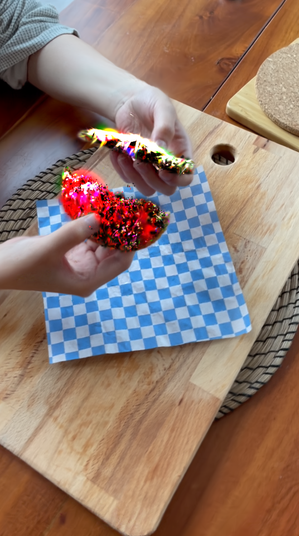} & 
\includegraphics[width=0.18\linewidth, height=0.18\linewidth]{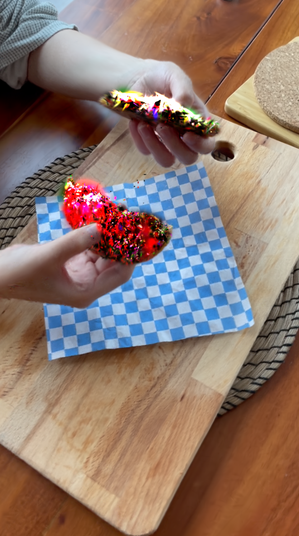} \\

\rotatebox{90}{\hspace{0.4cm}\tiny{Rendered View}}
\includegraphics[width=0.18\linewidth, height=0.18\linewidth]{figures/tracking/torch/renders_orign_view400/00200.png} & 
\includegraphics[width=0.18\linewidth, height=0.18\linewidth]{figures/tracking/torch/renders_orign_view400/00400.png} & 
\includegraphics[width=0.18\linewidth, height=0.18\linewidth]{figures/tracking/torch/renders_orign_view400/00600.png} & 
\includegraphics[width=0.18\linewidth, height=0.18\linewidth]{figures/tracking/torch/renders_orign_view400/00800.png} & 
\includegraphics[width=0.18\linewidth, height=0.18\linewidth]{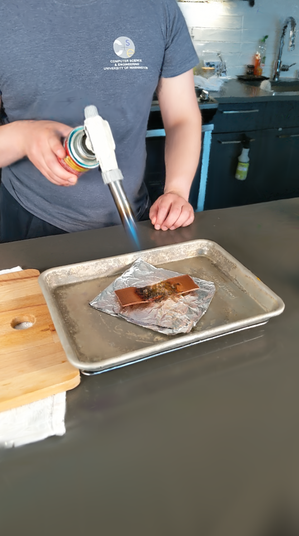} \\
\rotatebox{90}{\hspace{0.8cm}\tiny{Editing}}
\includegraphics[width=0.18\linewidth, height=0.18\linewidth]{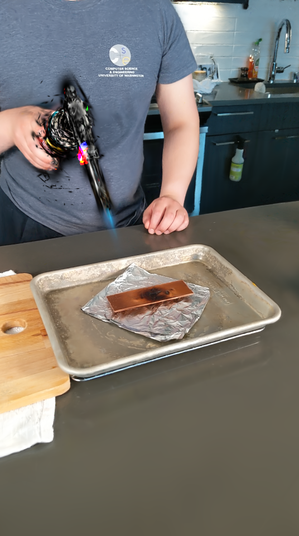} & 
\includegraphics[width=0.18\linewidth, height=0.18\linewidth]{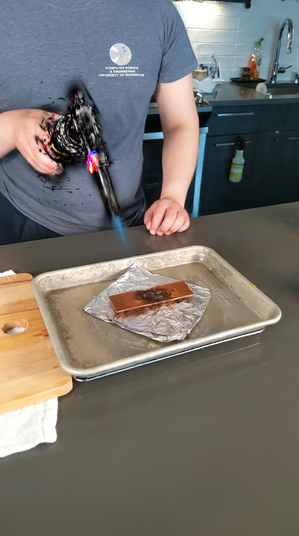} & 
\includegraphics[width=0.18\linewidth, height=0.18\linewidth]{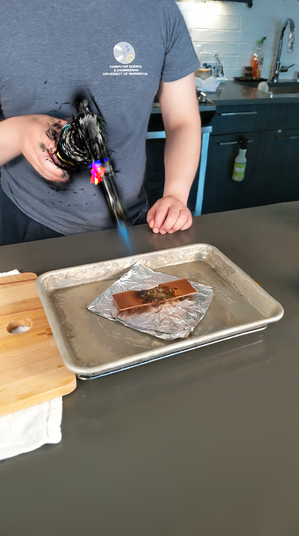} & 
\includegraphics[width=0.18\linewidth, height=0.18\linewidth]{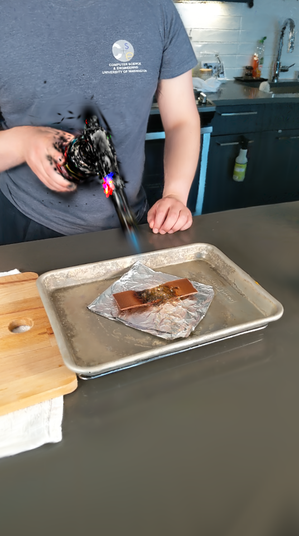} & 
\includegraphics[width=0.18\linewidth, height=0.18\linewidth]{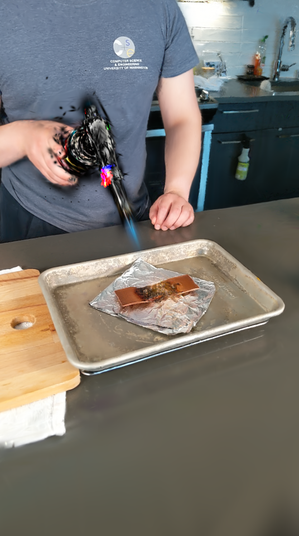} \\

\end{tabular}
 \vspace{-0.2cm}
\caption{Visual illustration of our method's ability to semantically edit segmented and tracked objects. 
 Given the segmented region of the ``cookie'' (resp. ``torch''), we consider the ability to edit its texture using the prompt ``strawberry'' (resp. ``handgun''). 
 Here, we display a randomly selected novel view over time. }
\label{fig:editing}
 \vspace{-0.2cm}
\end{figure*}

To evaluate our tracking and segmentation through time numerically, in \cref{tab:iou} we consider the
mean IoU for objects selected from dynamic scenes of HyperNerf dataset and synthetic D-NeRF dataset. 
For instance, we consider objects such as 
``hands" or ``cookie" presented in these scenes.  We consider the input 2D views and associated segmentations  (obtained manually, as noted above) as ground truth segmentation masks. 
For LSeg, we consider the segmentation of individual frames when using text prompts that correspond to the desired object. See \cref{sec:appendix_comparison_lseg} for more details.

As can be seen in \cref{tab:iou}, LSeg struggles to segment semantic elements in the scene, while our method is much better at localizing them. For the synthetic scenes, we found that LSeg tends to segment almost the entire object and not the local part specified by the text prompt, hence, its mIoU scores are very low. In addition, we enable the generation of novel views of semantic objects at different timesteps, which is not possible with LSeg. We provide a corresponding visual illustration of the masks produced by our method in comparison to LSeg in \cref{fig:tracking_lseg}. 
Additional masks for different views, including for the D-NeRF~\cite{pumarola2021d} synthetic dataset, are provided in \cref{sec:appendix_mask}.

As a further evaluation, \cref{fig:tracking_visual_comparison} provides a visual comparison of our method to the 3D static baseline (2), where we consider the segmentation at the canonical frame in comparison to that of the baseline. As can be seen, our method results in a much sharper output. 
We also conducted a perceptual user study on the HyperNeRF real world dataset (ii) comparing to the 3D static baseline (2). We consider a collection of videos where objects are tracked for two different random views. 
We ask users to rate on a scale of $1-5$: (Q1) ``How well was the object segmented?'' and (Q2) ``How consistent is the scene for the two different views?''. We consider $50$ users, and mean opinion scores are shown in \cref{tab:user_study}. Our method is superior in all cases. See additional details in \cref{sec:appendix_user_study} on how the user study was conducted.

\subsection{Semantic Editing}

We note that once a set of Gaussians, corresponding to a semantic entity, are selected, as in \cref{sec:tracking}, one can edit this entity semantically using an intuitive user input. Doing so involves two components. 
The first component entails the choice of the type of editing to be performed. Recall that each Gaussian consists of spatial parameters ($x_i, r_i, s_i$), appearance parameters ($c_i, \sigma_i$), semantic parameters ($f_i$), as well as an associated deformation network $D$ that operates on these Gaussians. As such, these properties can be modified in isolation. For example, to alter texture, one can manipulate only the colors $c_i$ of selected Gaussians. To alter geometry, one can modify the Gaussian's position $x_i$, density $\sigma_i$ as well as rotation and scale parameters that affect the covariance matrix $\Sigma_i$. Lastly, to affect deformation, one can modify $D$ in isolation. In this work, we focus on texture changes and leave other properties for future work.

The second component involves the signal used to affect these manipulations. 
To this end, we adopt a similar methodology of \cite{poole2022dreamfusion} using SDS-like loss. First, random views $V_1, \dots, V_N$ from $N$ from randomly selected angles are rendered. In our experiments, $N$ is chosen tp be 500. 
Given a pre-trained diffusion model $G$, one can consider whether $V_1, \dots, V_N$ maximize the underlying probability density function of $G$ when conditioned on given input (such as text). This can be realized by noising $V_1, \dots, V_N$ and considering the models' reconstruction loss when trying to reconstruct $V_1, \dots, V_N$. This signal can then be backpropagated to the input 3D Gaussians or to a subset of such Gaussians selected according to \cref{sec:tracking}, optimized according to this objective. 
We note that $G$ can be a video or image-based diffusion model, but in our setting, we use an image-based diffusion model only of Stable Diffusion v2.1~\cite{rombach2022high}. Note also that in our setting, novel views can be generated for different random timesteps, and so our random views are selected for random timesteps.

A visual illustration of our method is shown in \cref{fig:editing}. 
Additional novel views are provided in \cref{sec:additional_views}. 
As can be seen, our method successfully edits given regions according to the target text. We note that only texture is manipulated. More expressive and realistic editing could be enabled by also manipulating geometric properties, which we leave for future work. 

\subsection{Ablations}
\subsubsection{Different backbones.}
\begin{figure}[t]
\centering
\vspace{-0.2cm}
\begin{tabular}{c@{~}c@{~}c@{~}c@{~}c@{~}c@{~}c@{~}c@{~}c@{~}c}
  \multicolumn{1}{c}{\tiny{DINOv2}} & 
  \multicolumn{1}{c}{\tiny{CLIP (text)}} & 
  \multicolumn{1}{c}{\tiny{CLIP}} & 
  \multicolumn{1}{c}{\tiny{CLIP+DINOv2 }} & 
  \multicolumn{1}{c}{\tiny{SAM}} \\
\includegraphics[width=0.18\linewidth, height=0.18\linewidth]{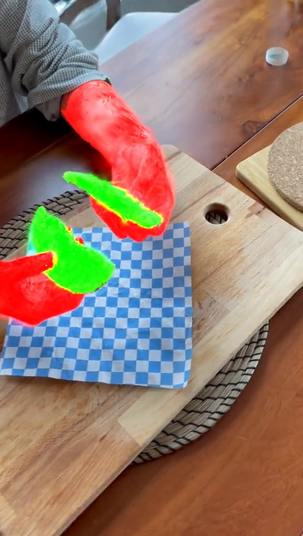} & 

\includegraphics[width=0.18\linewidth, height=0.18\linewidth]{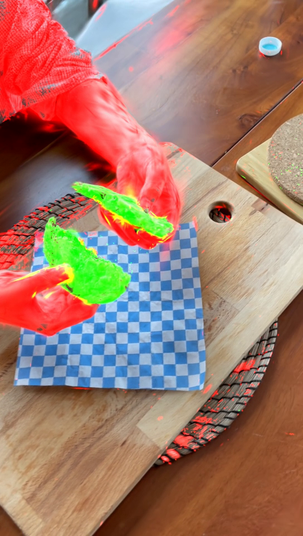} & 
\includegraphics[width=0.18\linewidth, height=0.18\linewidth]{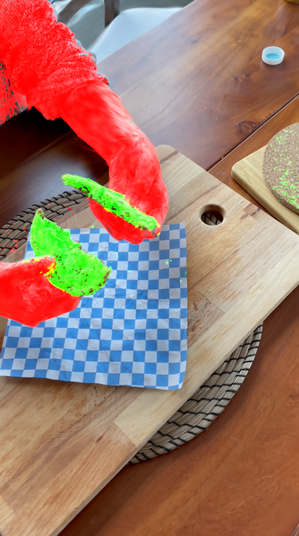} & 
\includegraphics[width=0.18\linewidth, height=0.18\linewidth]{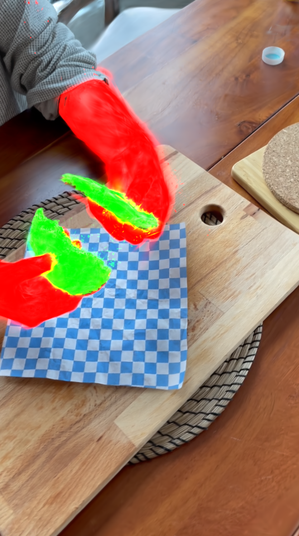} & 
\includegraphics[width=0.18\linewidth, height=0.18\linewidth]{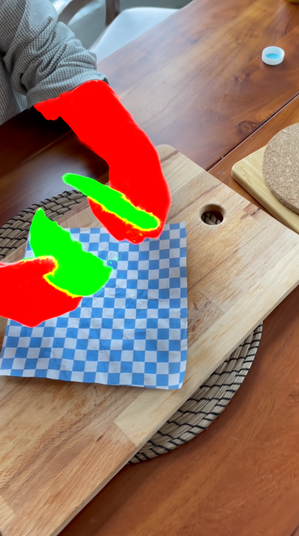}  \\
\end{tabular}
 \vspace{-0.2cm}
\caption{Visual comparison of our method with different semantic backbones}
\label{fig:backbones}
\end{figure}

\begin{wraptable}[13]{R}{3.5cm}
\centering
\small{
\vspace{-0.7cm}
\begin{tabular}{lcc}
\toprule
& mIoU  \\
\midrule
DINOv2 & 78.65   \\
CLIP(text) & 65.28  \\
CLIP & 75.01  \\
CLIP+DINOv2 & 75.58  \\
SAM & 81.90  \\
\bottomrule
\end{tabular}
\caption{Mean IoU for object segmentation of the Split-cookie scene from the HyperNeRF~\cite{park2021hypernerf} dataset using different semantic backbones.}}
\label{tab:backbones}
\end{wraptable}

Our framework is general to the choice of the backbone semantic feature extractor. We present both qualitative \cref{fig:backbones} and quantitative \cref{tab:backbones} comparison of different backbones. We note that although SAM \cite{kirillov2023segment} yields the best mIoU, we do not use it as our final backbone due to the fact that the features it provides are not semantic features but segmentation features.

\subsubsection{Granularity.}
\begin{figure}[t]
\centering
\vspace{-0.2cm}
\begin{tabular}{c@{}c@{}c@{~}c@{}c@{}c@{~}c@{}c@{}c@{~}c}
  \multicolumn{3}{c}{\tiny{$\theta=0.7$} (Hand) } & 
  \multicolumn{3}{c}{\tiny{$\theta=0.9$} (Finger)} \\
\includegraphics[width=0.15\linewidth, height=0.18\linewidth]{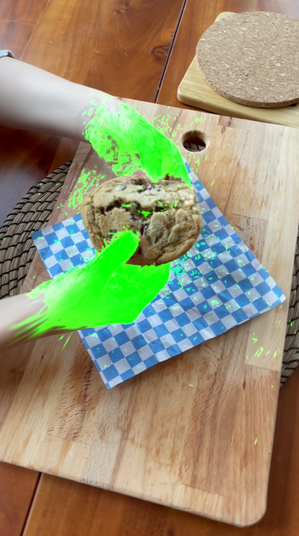} &

\includegraphics[width=0.15\linewidth, height=0.18\linewidth]{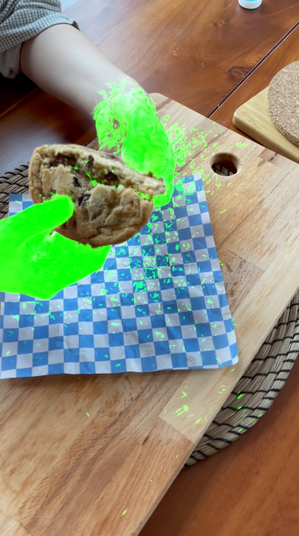} &  

\includegraphics[width=0.15\linewidth, height=0.18\linewidth]{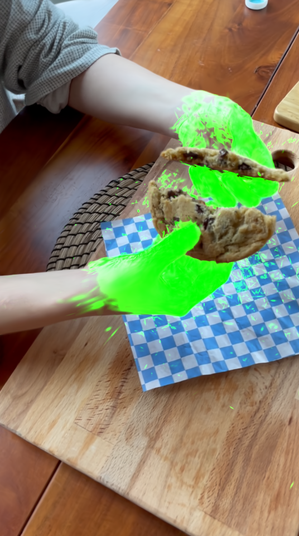} &

\includegraphics[width=0.15\linewidth, height=0.18\linewidth]{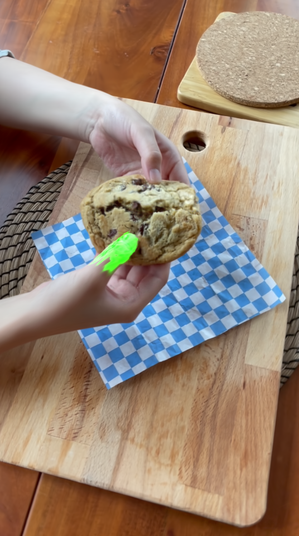} &  

\includegraphics[width=0.15\linewidth, height=0.18\linewidth]{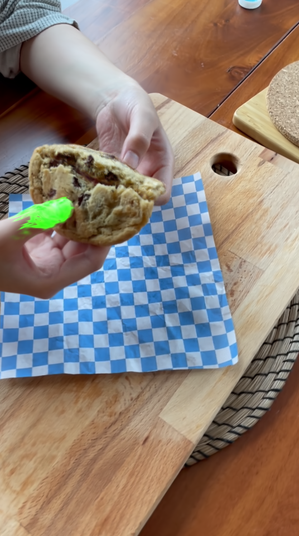} &

\includegraphics[width=0.15\linewidth, height=0.18\linewidth]{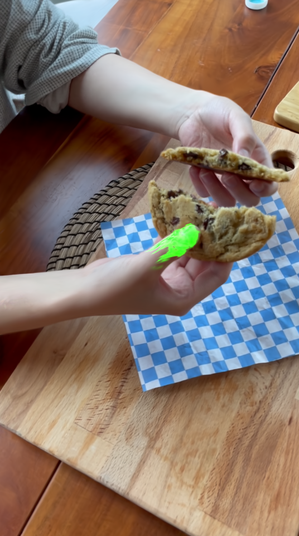} &

\end{tabular}
 \vspace{-0.2cm}
\caption{Demonstration of our method segmentation with different granularity thresholds $\theta$ using a 3D click}
\label{fig:granularity}
\end{figure}

Given a semantic feature $f_i$, as explained in \cref{sec:semantic_tracking}, we segment the scene by selecting the gaussians whose semantic feature's cosine similarity with $f_i$ is above some threshold $\theta$.
By selecting different choices of $\theta$ we can segment the scene with different levels of granularity. \cref{fig:granularity} demonstrate an example of such choices using DINOv2 as the semantic backbone. 

\subsection{Speed and Memory Requirements}
\label{sec:requirements}

As our method relies on the use of an underlying 3D Gaussian splatting representation, it also inherits its fast rendering speed. In particular, our method shows a rendering speed comparable to \cite{yang2023deformable}, of at least 30 FPS where the number of 3D Gaussians used is ~200K Gaussians. 
In terms of training time, due to the additional supervision of training against 2D feature maps, in addition to 2D RGB images, the training time takes about 5 hours for a standard scene with ~200K Gaussians. This is higher than \cite{yang2023deformable} which only requires 2D RGB images for supervision. See \cref{sec:appendix_speed_memory} for more details. 

In terms of memory requirement, our method requires the storage of per Gaussian feature in addition to the other parameters. Such feature is either 384-dimensional for DINO or 512-dimensional for CLIP. This requires a higher memory requirement than that of non-semantic dynamic 3D Gaussians~\cite{yang2023deformable}. 

\subsection{Limitations}
\label{sec:limitations}
\begin{figure}[t]
\centering
\vspace{-0.2cm}
\begin{tabular}{c@{~}c@{~}c@{~}c@{~}c@{~}c@{~}c@{~}c@{~}c@{~}c}

\rotatebox{90}{\hspace{0.25cm}\tiny{Low Frame Rate}} 
\includegraphics[width=0.18\linewidth, height=0.18\linewidth]{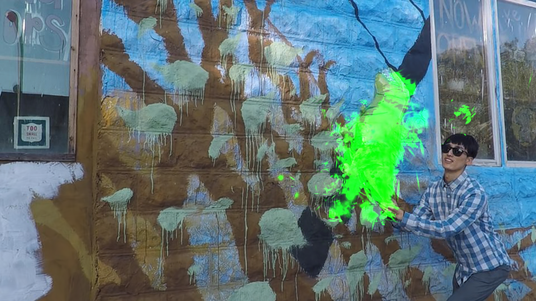} & 
\includegraphics[width=0.18\linewidth, height=0.18\linewidth]{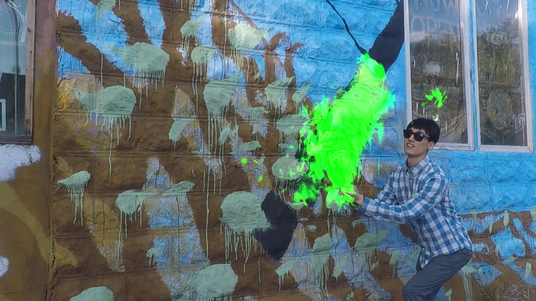} & 
\includegraphics[width=0.18\linewidth, height=0.18\linewidth]{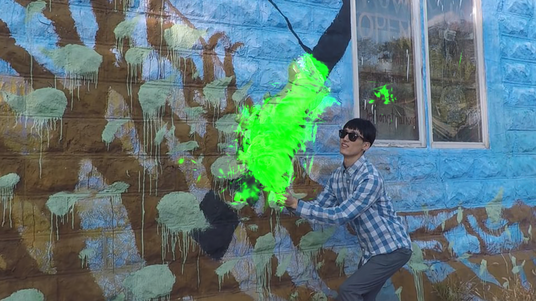} & 
\includegraphics[width=0.18\linewidth, height=0.18\linewidth]{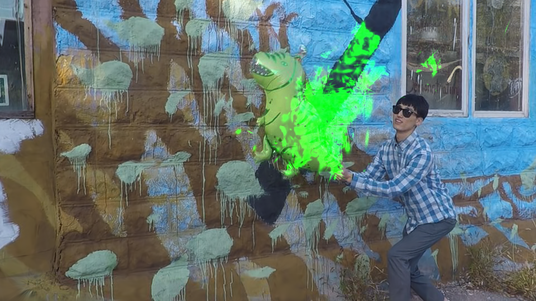} & 
\includegraphics[width=0.18\linewidth, height=0.18\linewidth] 
{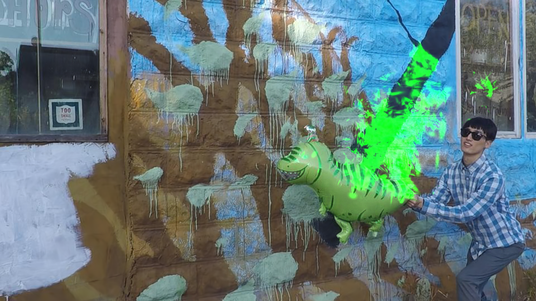} \\

\rotatebox{90}{\hspace{0.35cm}\tiny{Transparency}} 
\includegraphics[width=0.18\linewidth, height=0.18\linewidth]{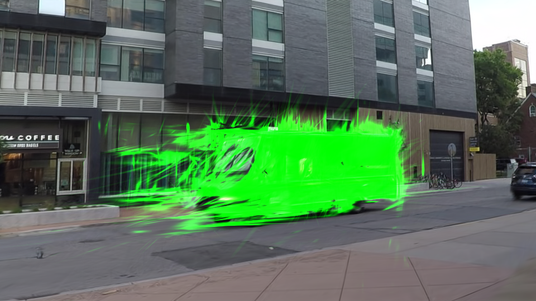} & 
\includegraphics[width=0.18\linewidth, height=0.18\linewidth]{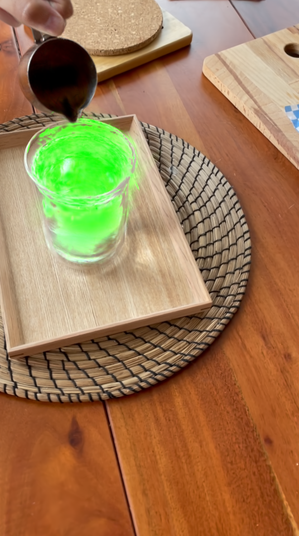} & 
\includegraphics[width=0.18\linewidth, height=0.18\linewidth]{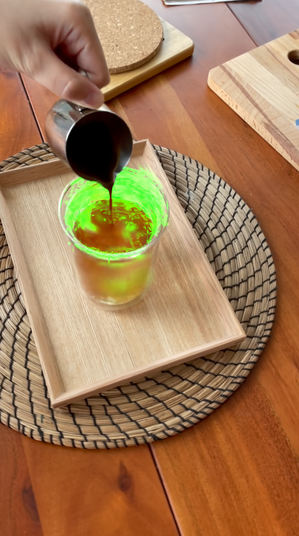} & 
\includegraphics[width=0.18\linewidth, height=0.18\linewidth]{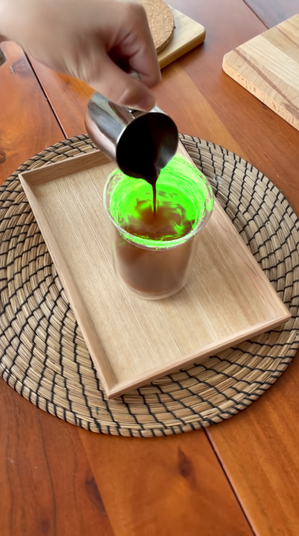} & 
\includegraphics[width=0.18\linewidth, height=0.18\linewidth] 
{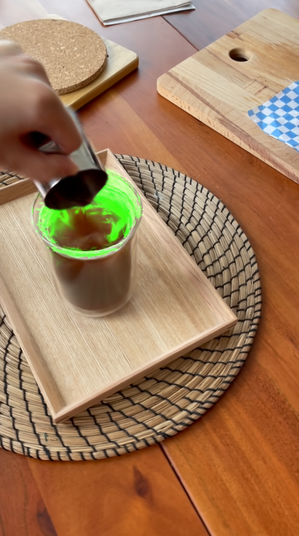} \\
\end{tabular}
 \vspace{-0.2cm}
\caption{Illustration of failure cases of our method, (i) when given a low frame rate video as input, (ii) when trying to segment a transparent object.}

\label{fig:limitations}
\end{figure}

Beyond training time and memory usage, as noted above, our method has some limitations.
We noticed two main failure cases of our method (i) when the given video frame rate is low (ii) when the object we wish to segment is highly transparent. \cref{fig:limitations} illustrate those failure cases where in the case of (i) we provide our method as input a video from NVIDIA Dynamic Scene dataset \cite{yoon2020novel}, where  each two consecutive frames  exhibit large changes. In the case of (ii) our method fails when trying to segment a highly transparent object, specifically, the \emph{glass} in the Americano scene of HyperNerf \cite{park2021hypernerf}.

As noted in \cref{sec:method}, as the 2D feature maps are assumed to be of the same resolution as RGB images, one has to upsample the 2D feature maps which are typically of lower resolution. This results in lower resolution 3D features. As such, our 3D segmentation struggles in capturing fine grained details. For future work, we hope to incorporate higher resolution features. 
We also hope to incorporate video-based features, which inherently capture 3D consistency between frames. 
In addition, with the use of only a single monocular video as input, our method still exhibits artifacts in the 3D reconstruction of novel views. Further, we note that our editing results are limited, and while they capture the desired texture, high-resolution editing is still elusive. Advances in 3D editing could be used in conjunction with our segmentation approach to obtain improved localized editing of objects in dynamic scenes.

\section{Conclusion} \label{sec:conclusion}

We introduced DGD, a novel 3D representation that goes beyond geometry and appearance, so as to capture the semantics of dynamic objects in real-world scenes. Our representation is based on 3D Gaussians with a set disentangled controls of geometry, texture, deformation, and semantics, which can be quickly and efficiently rasterized to 2D views. DGD enables applications like real-time photorealistic rendering, dense semantic 3D tracking, and semantic editing of dynamic objects through an intuitive text or click interface. Our experiments demonstrate DGD's success in these applications, given only a single monocular video as input, for challenging real world settings. Our work takes another step towards intuitive interaction with dynamic 3D scenes. In the future, we hope to extend the envelope of manipulations to include also manipulations involving geometric changes and deformations.

\bibliographystyle{splncs04}
\bibliography{references.bib}
\clearpage

\appendix
\section{Additional Views}
\label{sec:additional_views}

\begin{figure*}
\centering
\begin{tabular}{ccccc}

$t=0$ & $t=200$ & $t=400$ & $t=600$ & $t=800$ \\
\rotatebox{90}{\hspace{0.5cm}\tiny{Rendered View}}
\includegraphics[width=0.17\linewidth, trim={0 5.025cm 0 0},clip]{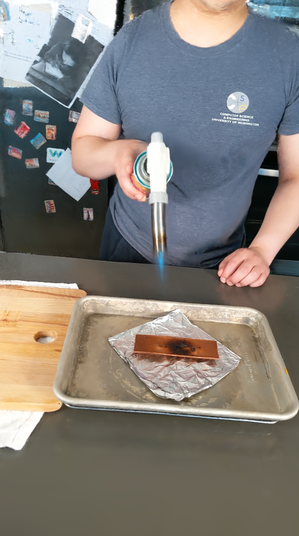} & 
\includegraphics[width=0.17\linewidth, trim={0 5.025cm 0 0},clip]{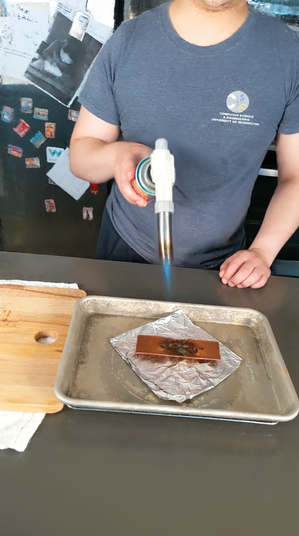} & 
\includegraphics[width=0.17\linewidth, trim={0 5.025cm 0 0},clip]{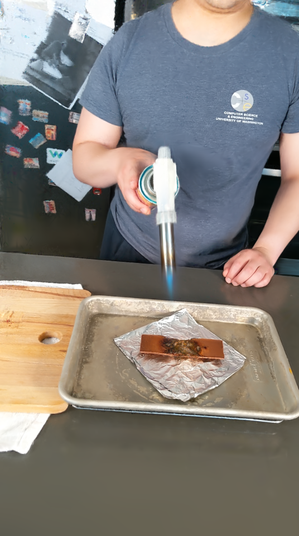} & 
\includegraphics[width=0.17\linewidth, trim={0 5.025cm 0 0},clip]{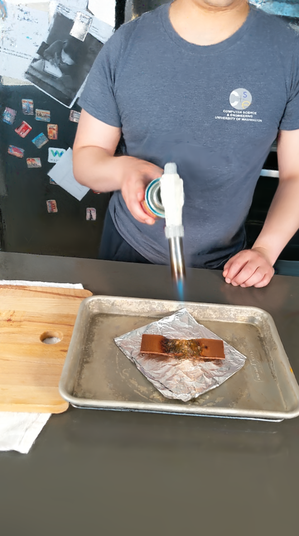} & 
\includegraphics[width=0.17\linewidth, trim={0 5.025cm 0 0},clip]{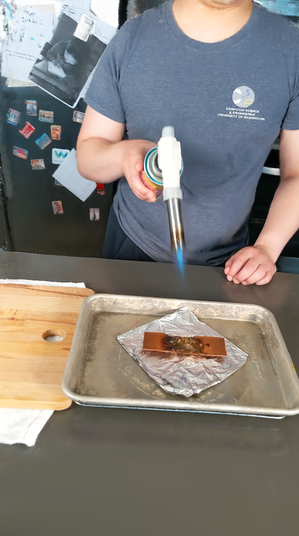} \\
\rotatebox{90}{\hspace{0.5cm}\tiny{Segmentation}}
\includegraphics[width=0.17\linewidth, trim={0 5.025cm 0 0},clip]{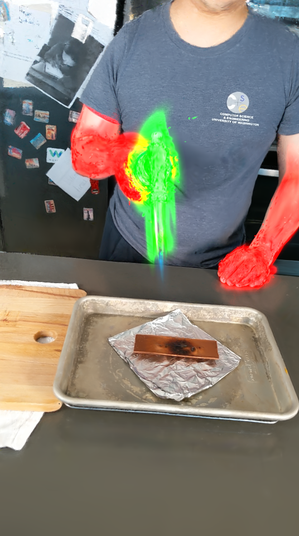} & 
\includegraphics[width=0.17\linewidth, trim={0 5.025cm 0 0},clip]{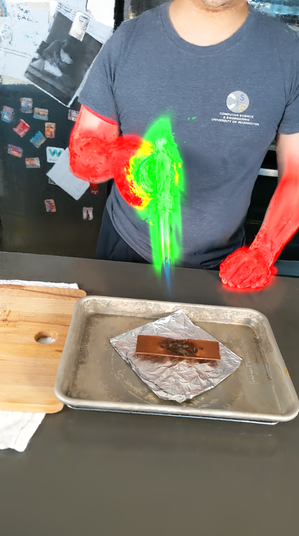} & 
\includegraphics[width=0.17\linewidth, trim={0 5.025cm 0 0},clip]{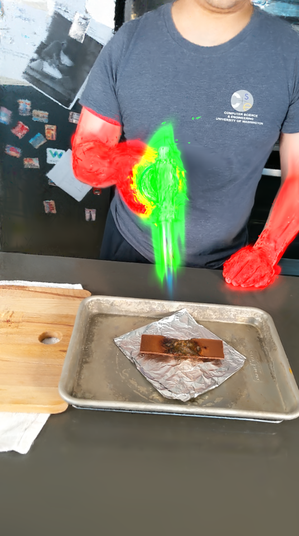} & 
\includegraphics[width=0.17\linewidth, trim={0 5.025cm 0 0},clip]{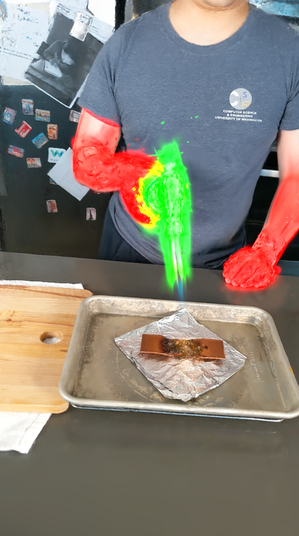} & 
\includegraphics[width=0.17\linewidth, trim={0 5.025cm 0 0},clip]{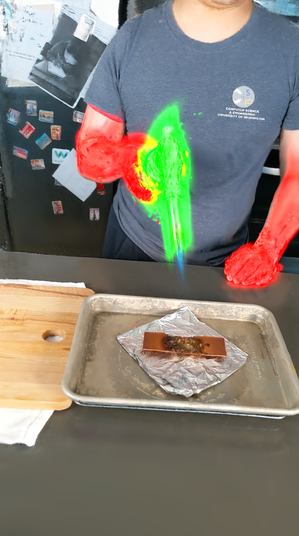} \\

\rotatebox{90}{\hspace{0.5cm}\tiny{Rendered View}}
\includegraphics[width=0.17\linewidth, trim={0 5.025cm 0 0},clip]{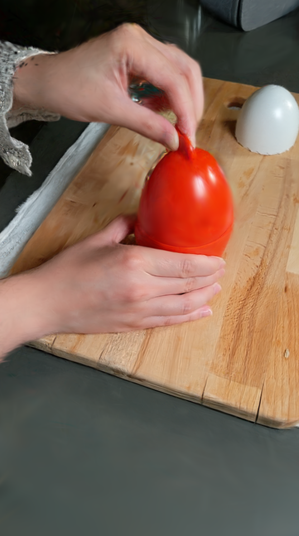} & 
\includegraphics[width=0.17\linewidth, trim={0 5.025cm 0 0},clip]{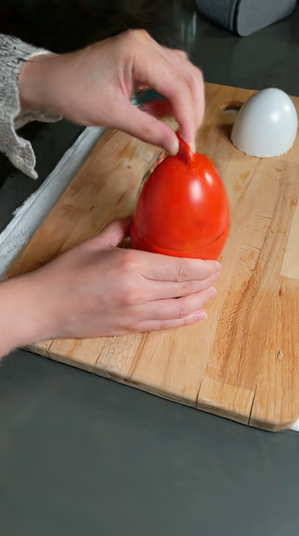} & 
\includegraphics[width=0.17\linewidth, trim={0 5.025cm 0 0},clip]{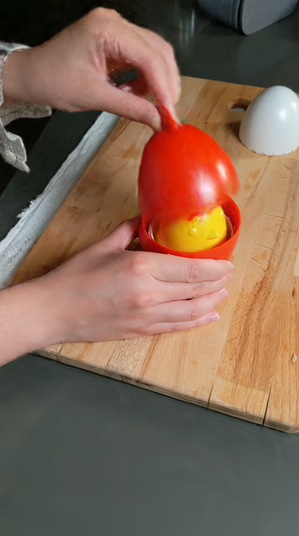} & 
\includegraphics[width=0.17\linewidth, trim={0 5.025cm 0 0},clip]{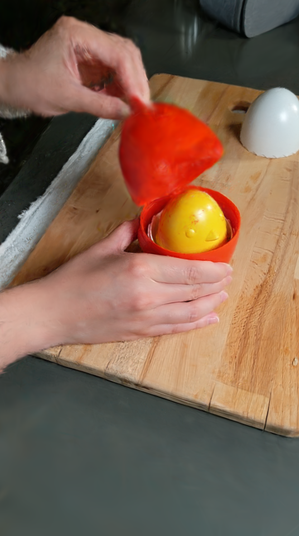} & 
\includegraphics[width=0.17\linewidth, trim={0 5.025cm 0 0},clip]{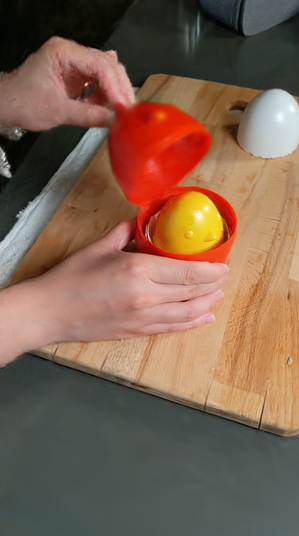} \\
\rotatebox{90}{\hspace{0.5cm}\tiny{Segmentation}}
\includegraphics[width=0.17\linewidth, trim={0 5.025cm 0 0},clip]{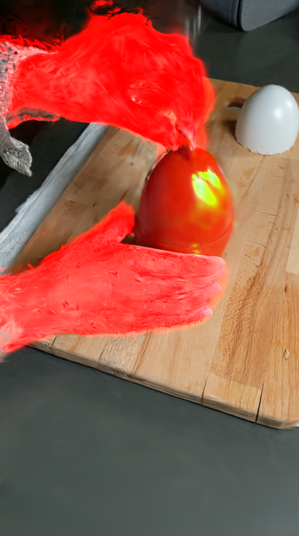} & 
\includegraphics[width=0.17\linewidth, trim={0 5.025cm 0 0},clip]{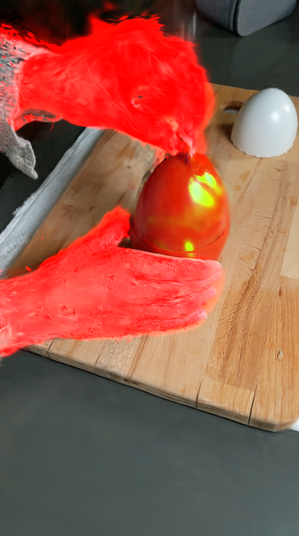} & 
\includegraphics[width=0.17\linewidth, trim={0 5.025cm 0 0},clip]{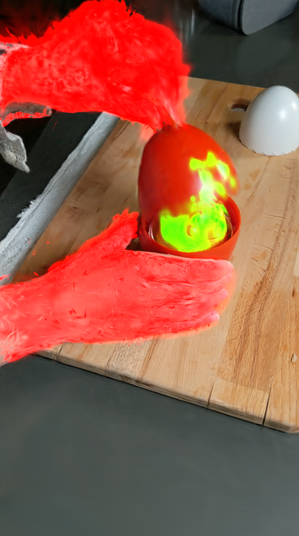} & 
\includegraphics[width=0.17\linewidth, trim={0 5.025cm 0 0},clip]{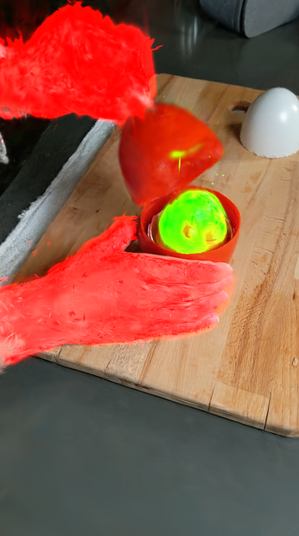} & 
\includegraphics[width=0.17\linewidth, trim={0 5.025cm 0 0},clip]{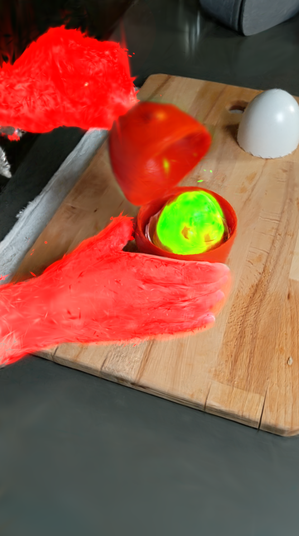} \\

\rotatebox{90}{\hspace{0.5cm}\tiny{Rendered View}}
\includegraphics[width=0.17\linewidth, trim={0 5.025cm 0 0},clip]{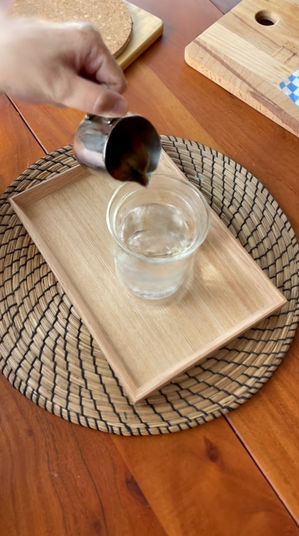} & 
\includegraphics[width=0.17\linewidth, trim={0 5.025cm 0 0},clip]{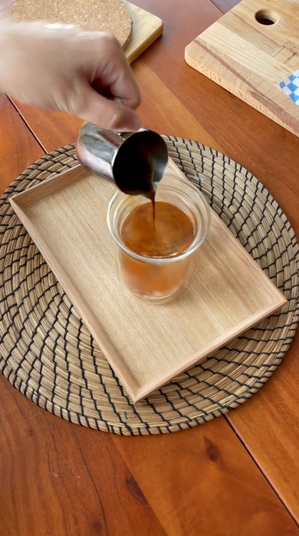} & 
\includegraphics[width=0.17\linewidth, trim={0 5.025cm 0 0},clip]{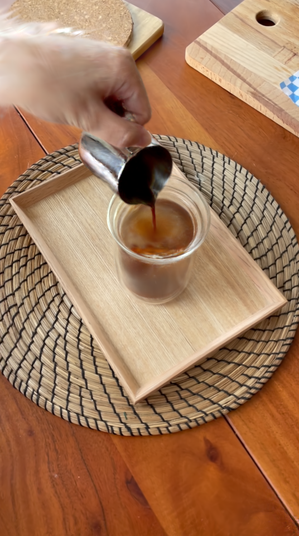} & 
\includegraphics[width=0.17\linewidth, trim={0 5.025cm 0 0},clip]{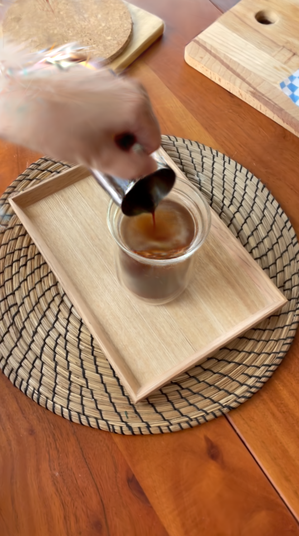} & 
\includegraphics[width=0.17\linewidth, trim={0 5.025cm 0 0},clip]{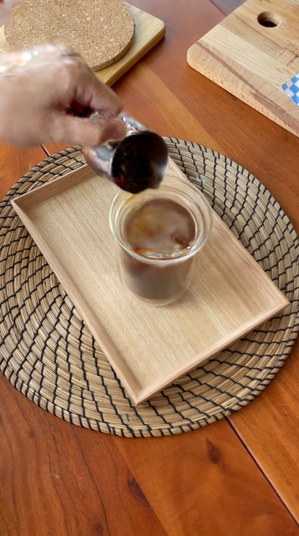} \\
\rotatebox{90}{\hspace{0.5cm}\tiny{Segmentation}}
\includegraphics[width=0.17\linewidth, trim={0 5.025cm 0 0},clip]{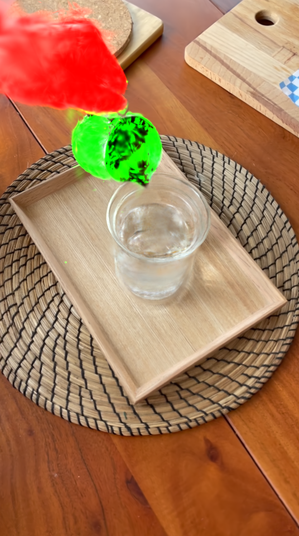} & 
\includegraphics[width=0.17\linewidth, trim={0 5.025cm 0 0},clip]{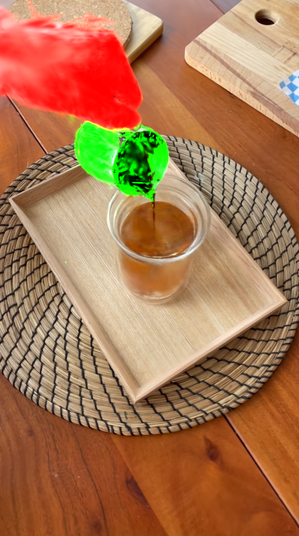} & 
\includegraphics[width=0.17\linewidth, trim={0 5.025cm 0 0},clip]{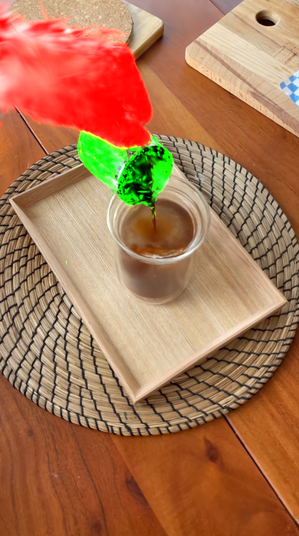} & 
\includegraphics[width=0.17\linewidth, trim={0 5.025cm 0 0},clip]{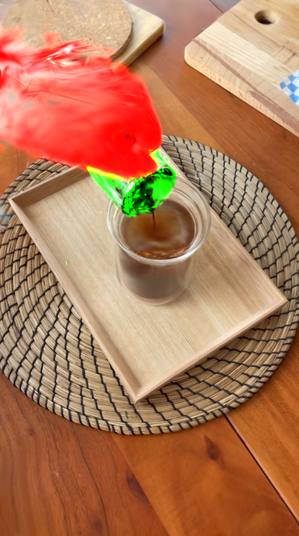} & 
\includegraphics[width=0.17\linewidth, trim={0 5.025cm 0 0},clip]{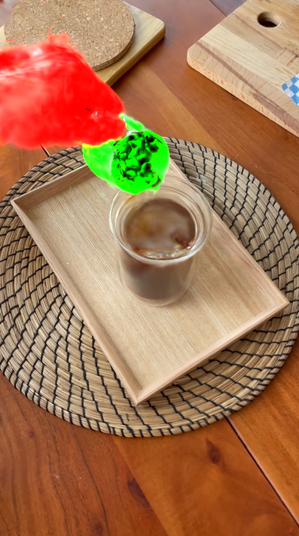} \\

\end{tabular}
\caption{Additional novel views corresponding to scenes displayed in Fig.~3 of the main text for segmenting and tracking objects semantically over time for the real-world HyperNerf dataset \cite{park2021hypernerf}. The considered objects are marked by green and red colors. }

\label{fig:tracking_extra_views}
\end{figure*}

\begin{figure*}
\centering
\begin{tabular}{ccccc}

$t=0$ & $t=200$ & $t=400$ & $t=600$ & $t=800$ \\
\rotatebox{90}{\hspace{0.3cm}\tiny{Rendered View}}
\includegraphics[width=0.165\linewidth]{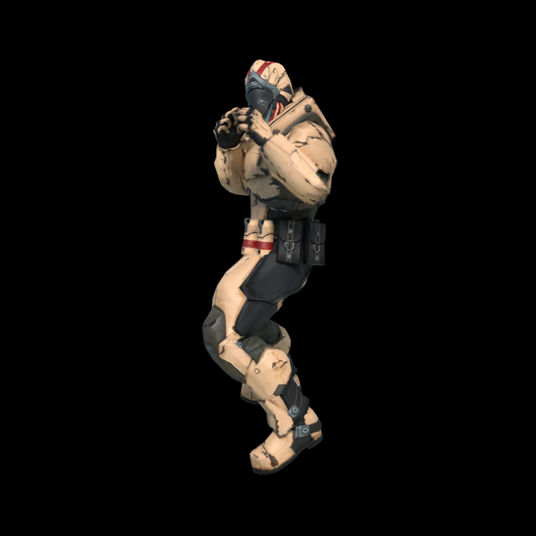} & 
\includegraphics[width=0.165\linewidth]{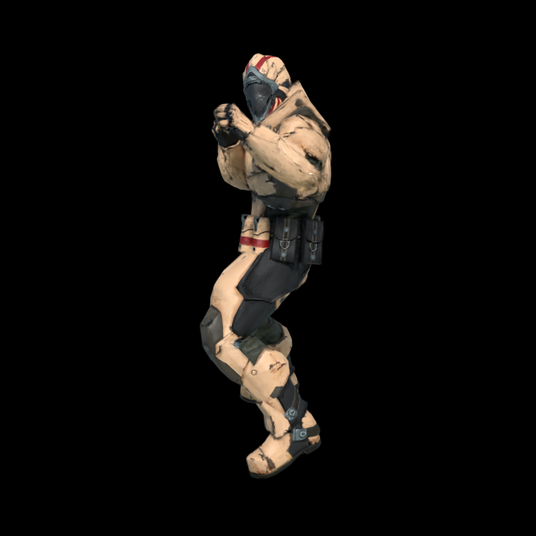} & 
\includegraphics[width=0.165\linewidth]{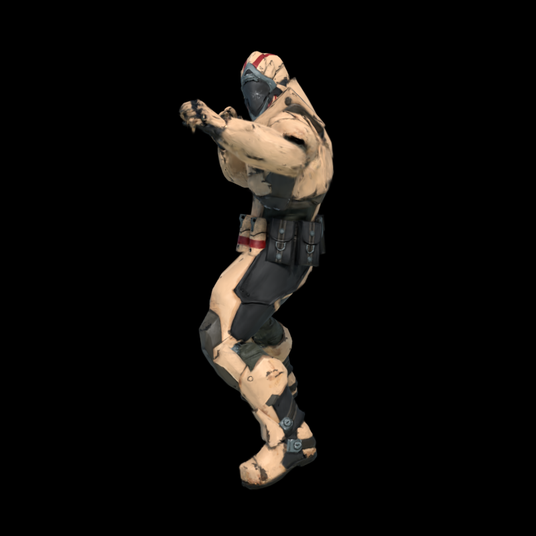} & 
\includegraphics[width=0.165\linewidth]{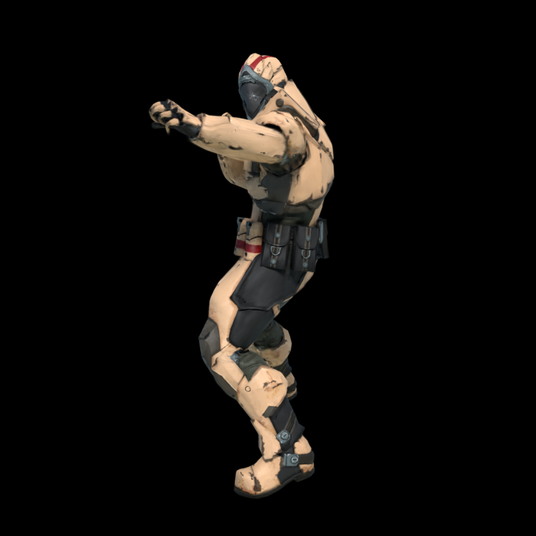} & 
\includegraphics[width=0.165\linewidth]{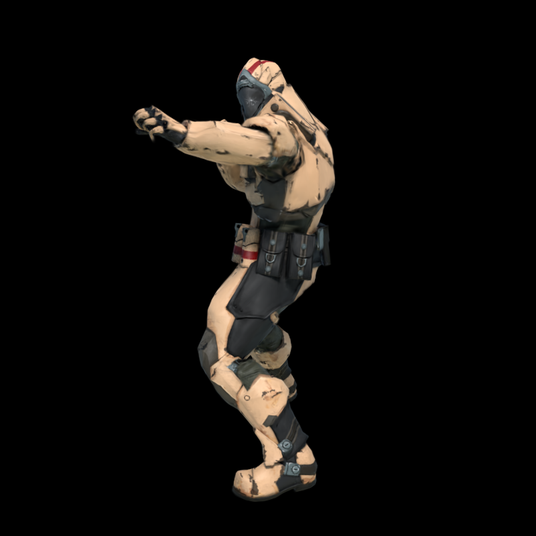} \\
\rotatebox{90}{\hspace{0.3cm}\tiny{Segmentation}}
\includegraphics[width=0.165\linewidth]{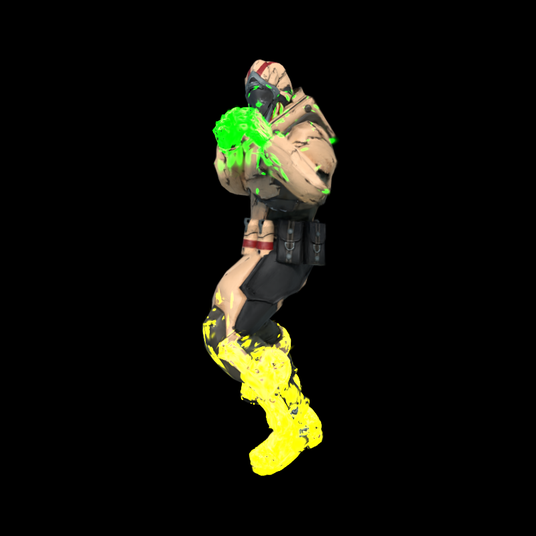} & 
\includegraphics[width=0.165\linewidth]{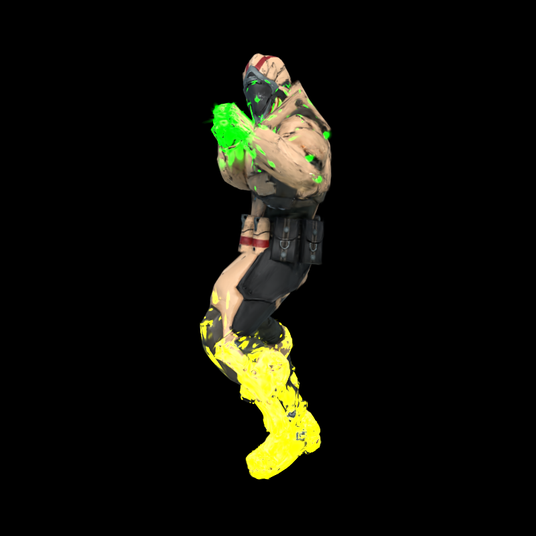} & 
\includegraphics[width=0.165\linewidth]{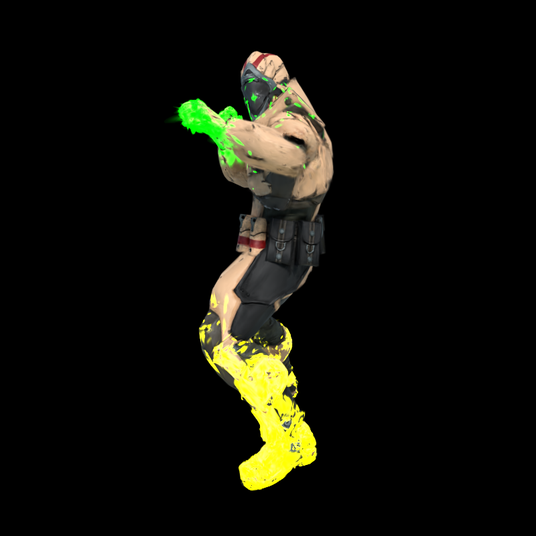} & 
\includegraphics[width=0.165\linewidth]{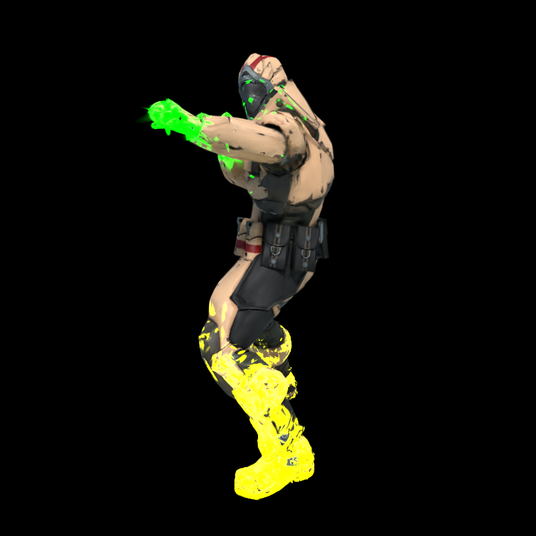} & 
\includegraphics[width=0.165\linewidth]{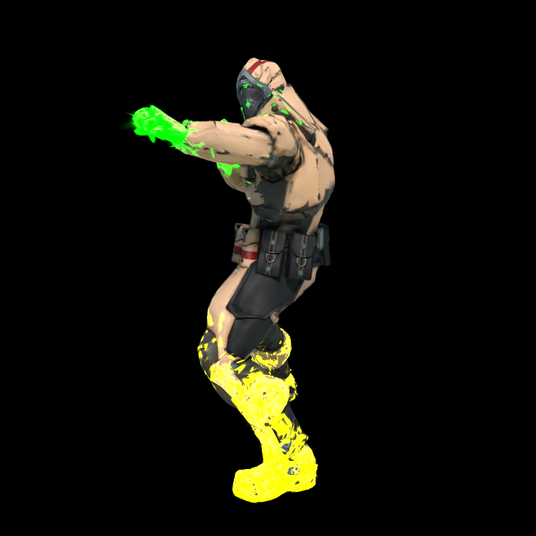} \\

\rotatebox{90}{\hspace{0.3cm}\tiny{Rendered View}}
\includegraphics[width=0.165\linewidth]{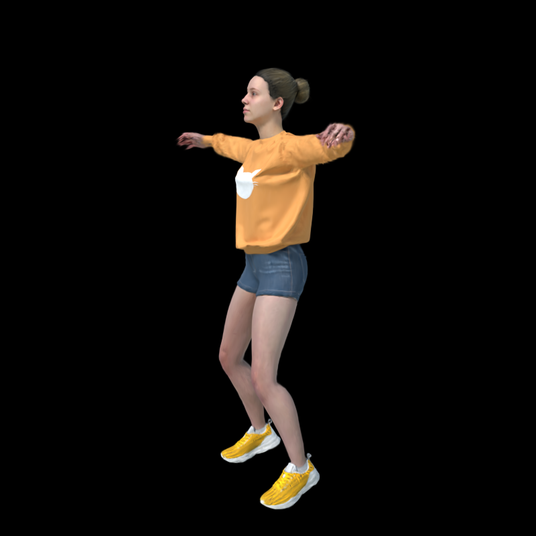} & 
\includegraphics[width=0.165\linewidth]{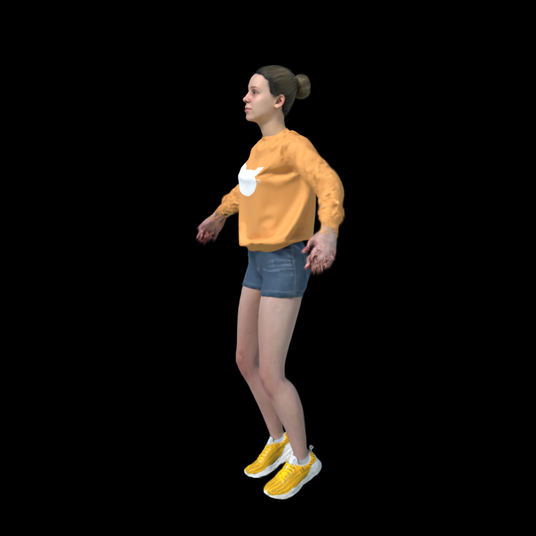} & 
\includegraphics[width=0.165\linewidth]{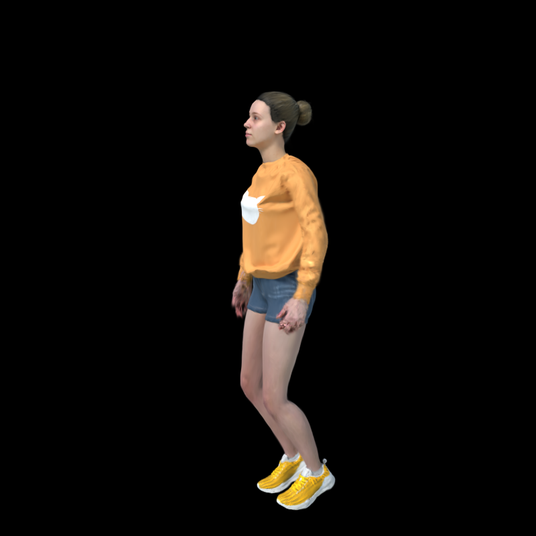} & 
\includegraphics[width=0.165\linewidth]{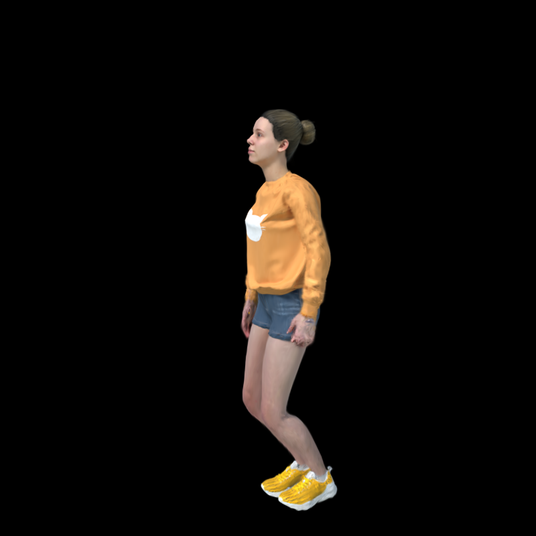} & 
\includegraphics[width=0.165\linewidth]{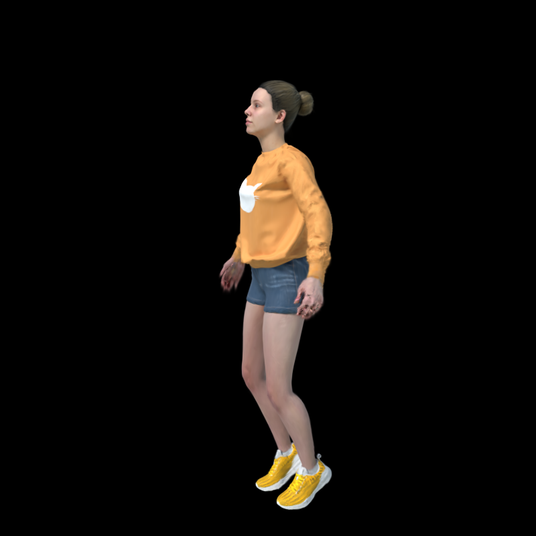} \\
\rotatebox{90}{\hspace{0.3cm}\tiny{Segmentation}}
\includegraphics[width=0.165\linewidth]{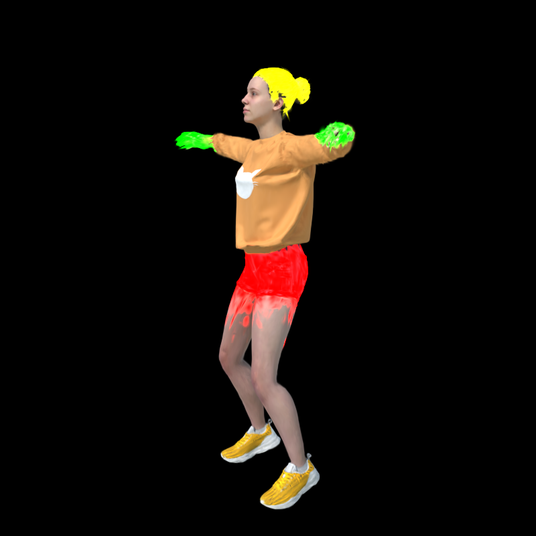} & 
\includegraphics[width=0.165\linewidth]{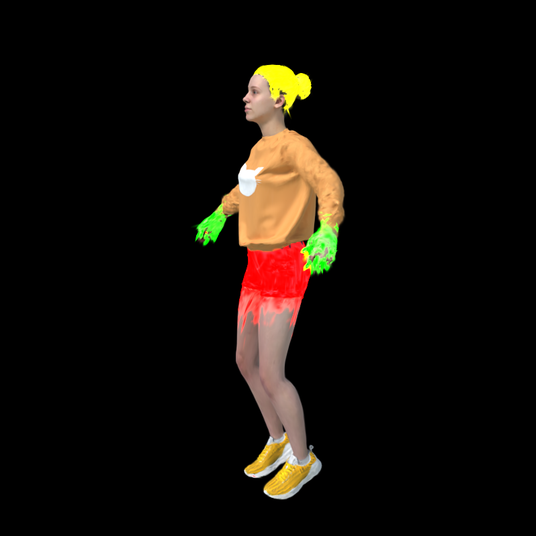} & 
\includegraphics[width=0.165\linewidth]{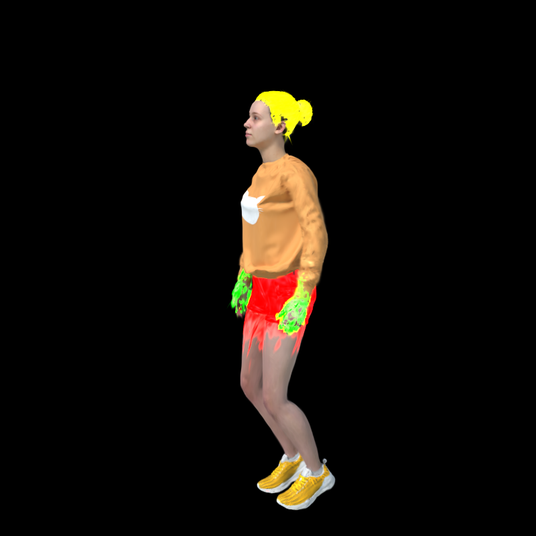} & 
\includegraphics[width=0.165\linewidth]{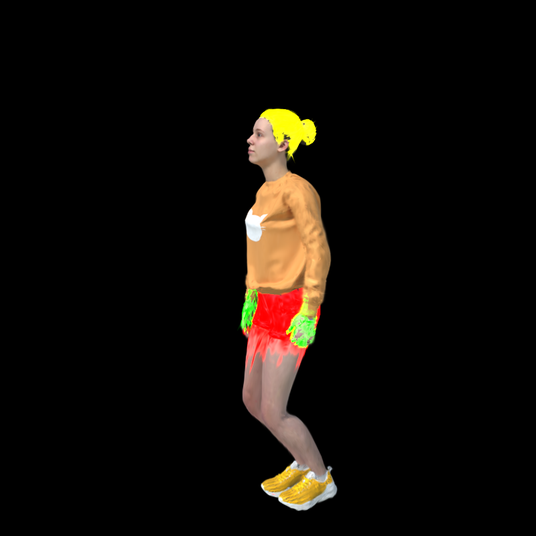} & 
\includegraphics[width=0.165\linewidth]{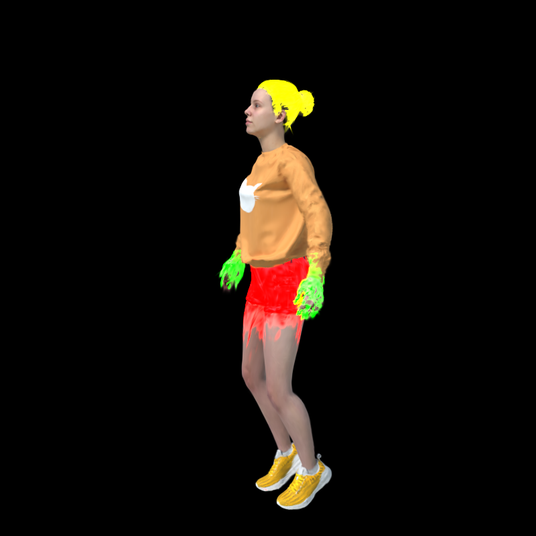} \\ 

\rotatebox{90}{\hspace{0.3cm}\tiny{Rendered View}}
\includegraphics[width=0.165\linewidth]{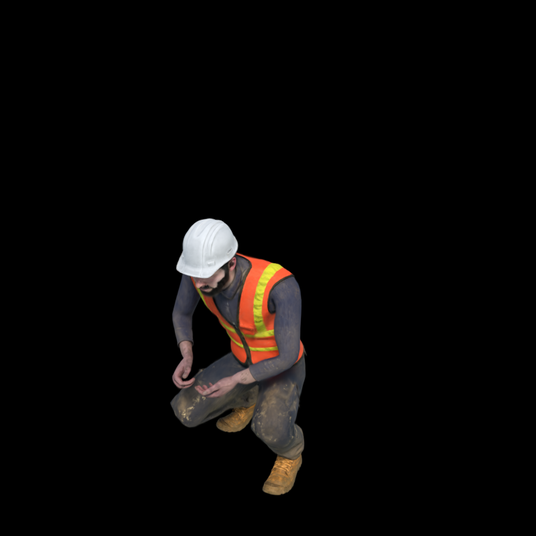} & 
\includegraphics[width=0.165\linewidth]{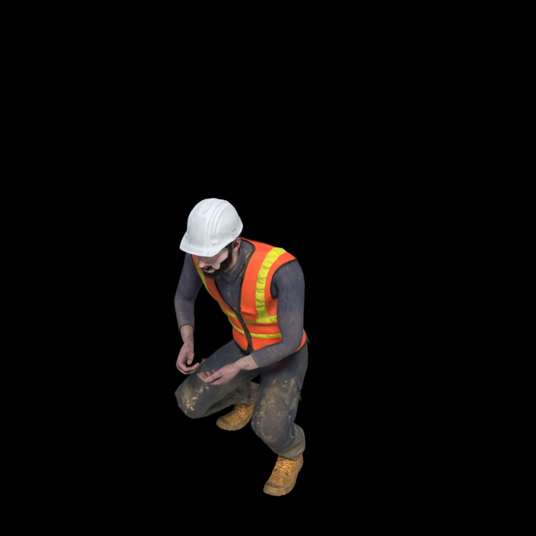} & 
\includegraphics[width=0.165\linewidth]{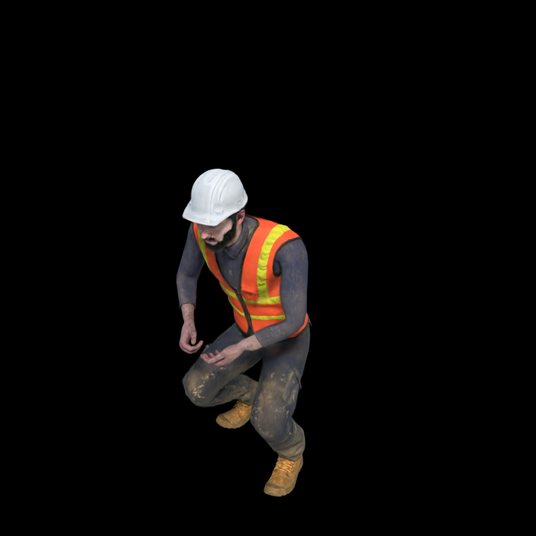} & 
\includegraphics[width=0.165\linewidth]{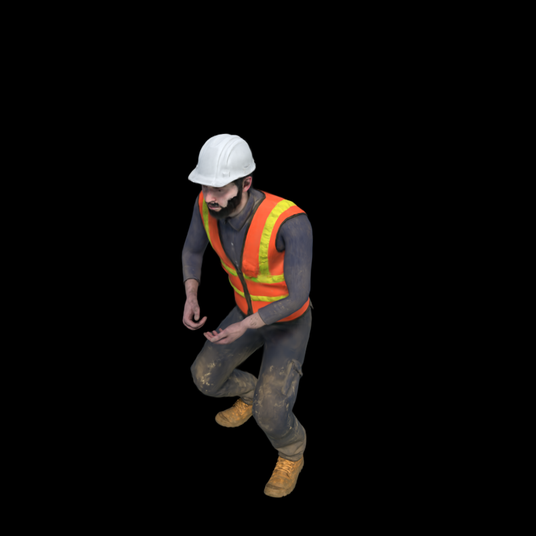} & 
\includegraphics[width=0.165\linewidth]{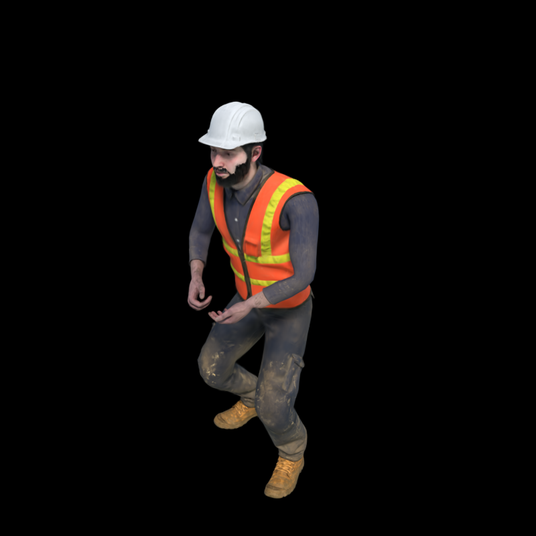} \\
\rotatebox{90}{\hspace{0.3cm}\tiny{Segmentation}}
\includegraphics[width=0.165\linewidth]{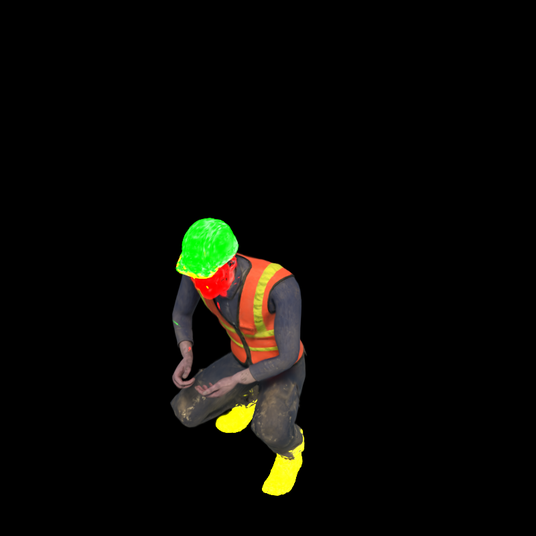} & 
\includegraphics[width=0.165\linewidth]{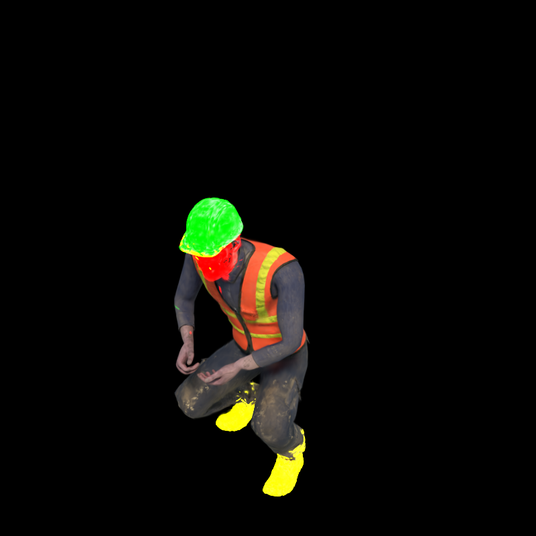} & 
\includegraphics[width=0.165\linewidth]{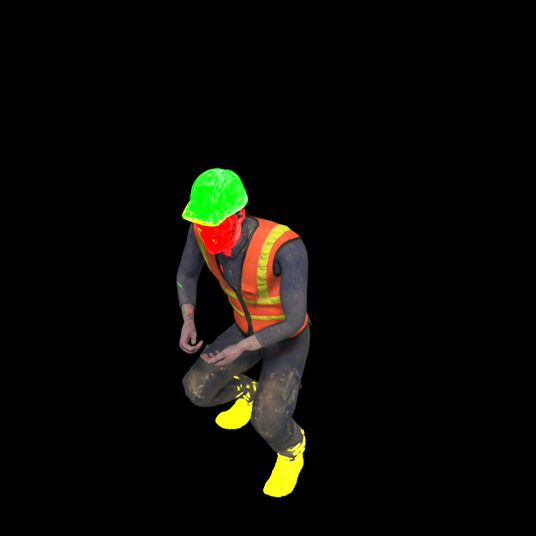} & 
\includegraphics[width=0.165\linewidth]{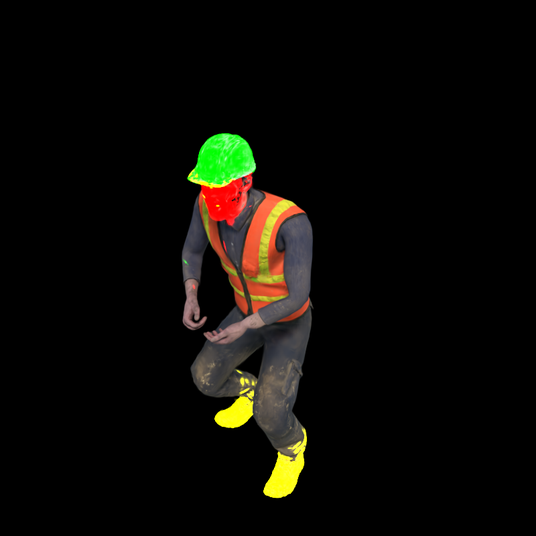} & 
\includegraphics[width=0.165\linewidth]{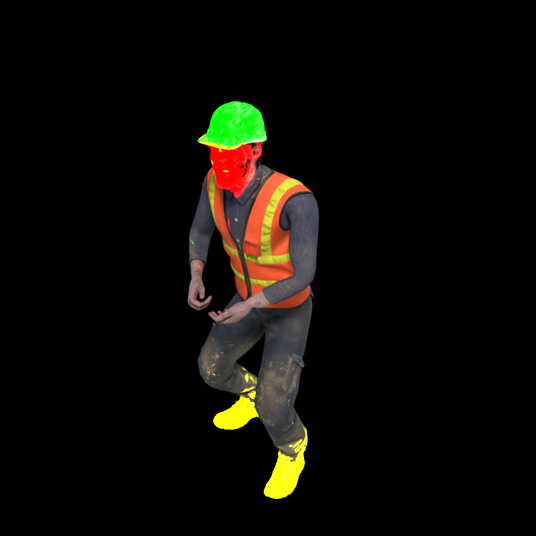} \\

\rotatebox{90}{\hspace{0.3cm}\tiny{Rendered View}}
\includegraphics[width=0.165\linewidth]{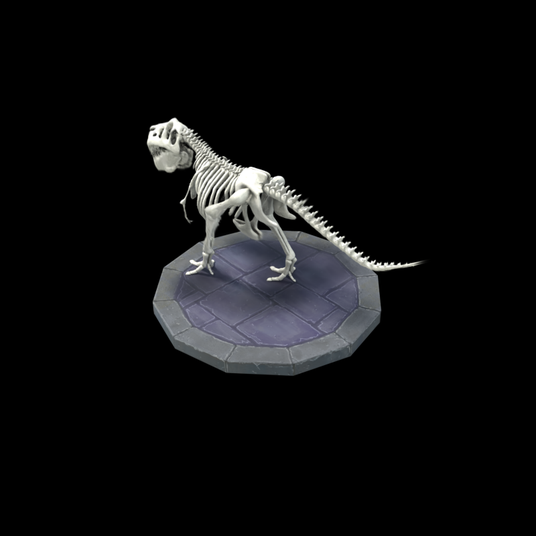} & 
\includegraphics[width=0.165\linewidth]{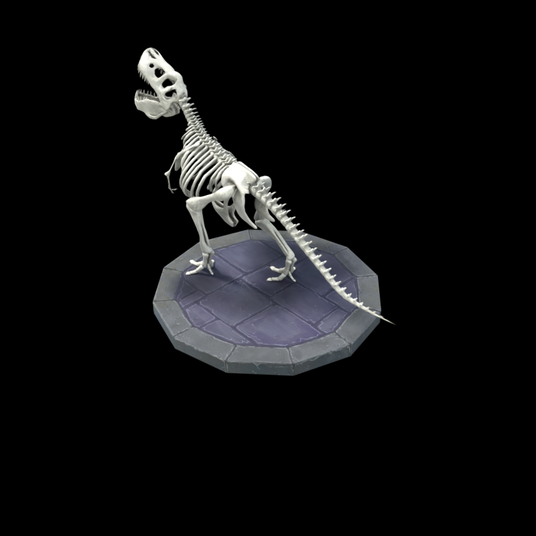} & 
\includegraphics[width=0.165\linewidth]{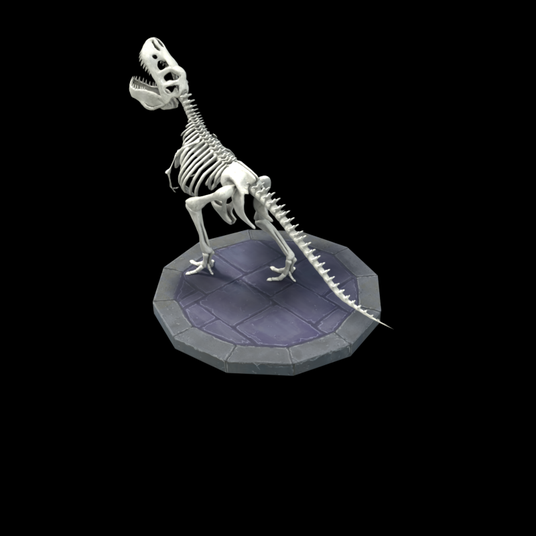} & 
\includegraphics[width=0.165\linewidth]{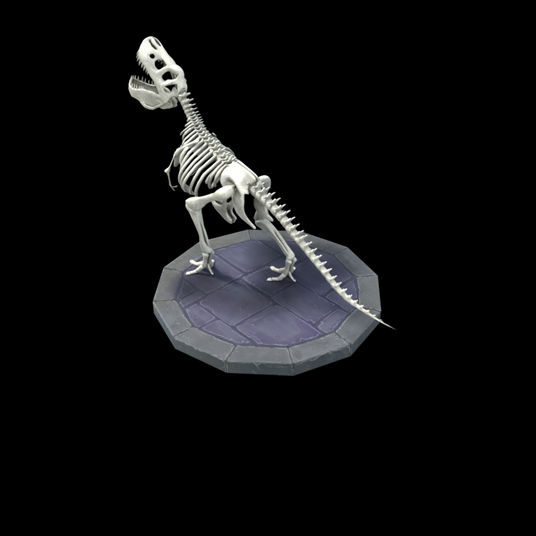} & 
\includegraphics[width=0.165\linewidth]{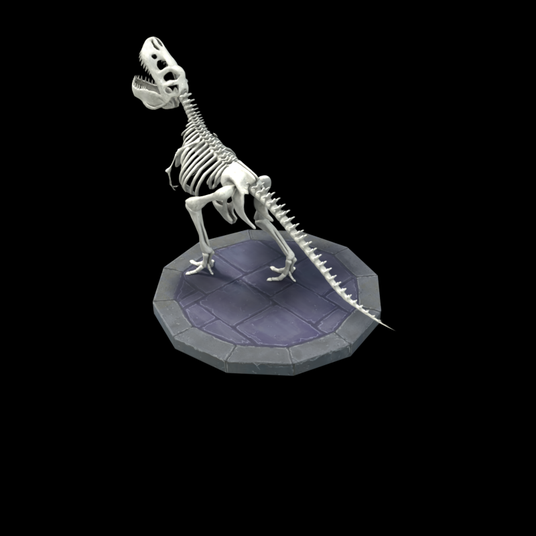} \\
\rotatebox{90}{\hspace{0.3cm}\tiny{Segmentation}}
\includegraphics[width=0.165\linewidth]{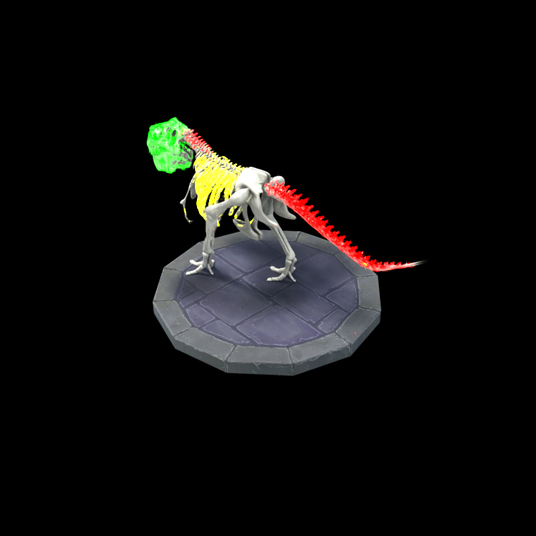} & 
\includegraphics[width=0.165\linewidth]{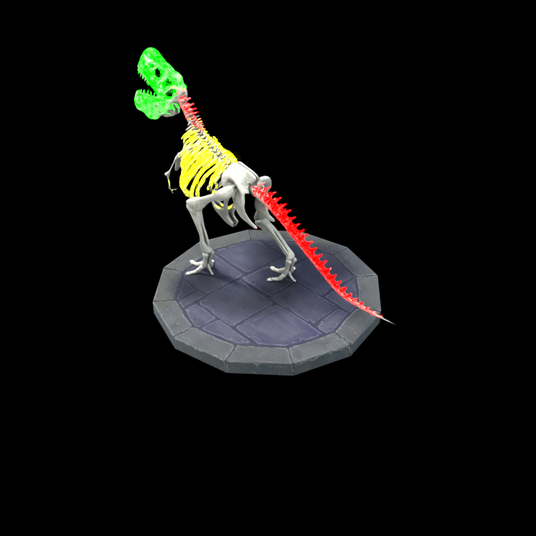} & 
\includegraphics[width=0.165\linewidth]{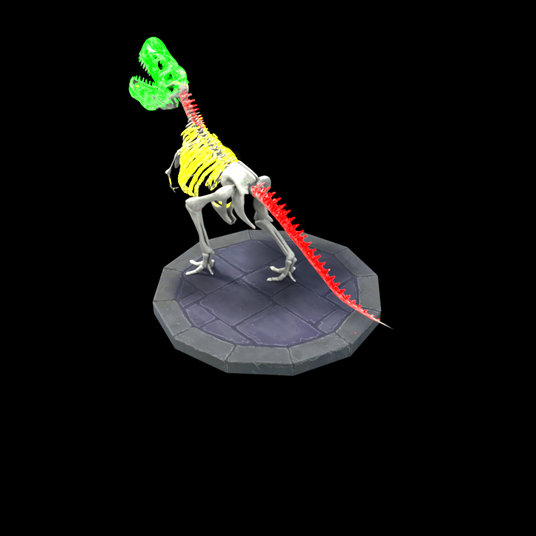} & 
\includegraphics[width=0.165\linewidth]{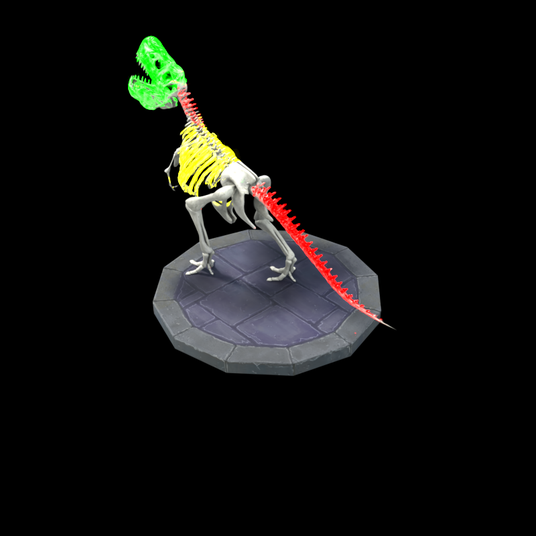} & 
\includegraphics[width=0.165\linewidth]{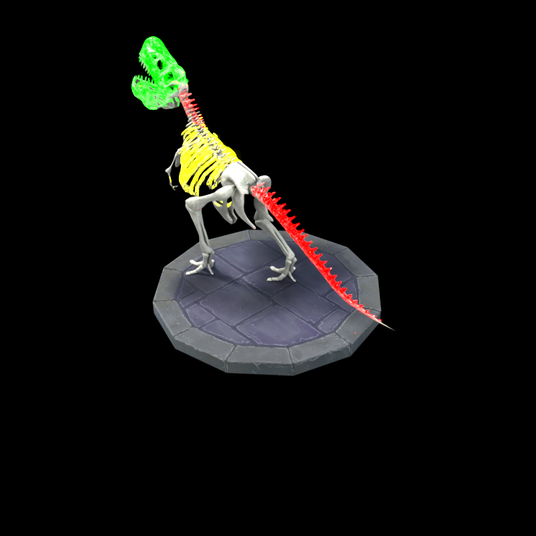} \\ 

\end{tabular}
\caption{Additional novel views corresponding to scenes displayed in Fig.~4 of the main text for segmenting and tracking parts semantically over time, for the synthetic D-NeRF dataset \cite{pumarola2021d}. The considered parts are marked by green, yellow, and red colors. 
}
\label{fig:tracking_synthetic_extra_views}
\end{figure*}

\begin{figure*}
\centering
\begin{tabular}{ccccc}

$t=0$ & $t=200$ & $t=400$ & $t=600$ & $t=800$ \\
\rotatebox{90}{\hspace{0.3cm}\tiny{Rendered View}}
\includegraphics[width=0.17\linewidth, height=0.17\linewidth]{figures/tracking/cookie/renders_cookie_view2/00200.png} & 
\includegraphics[width=0.17\linewidth, height=0.17\linewidth]{figures/tracking/cookie/renders_cookie_view2/00400.png} & 
\includegraphics[width=0.17\linewidth, height=0.17\linewidth]{figures/tracking/cookie/renders_cookie_view2/00600.png} & 
\includegraphics[width=0.17\linewidth, height=0.17\linewidth]{figures/tracking/cookie/renders_cookie_view2/00800.png} & 
\includegraphics[width=0.17\linewidth, height=0.17\linewidth]{figures/tracking/cookie/renders_cookie_view2/00999.png} \\
\rotatebox{90}{\hspace{0.6cm}\tiny{Editing}}
\includegraphics[width=0.17\linewidth, height=0.17\linewidth]{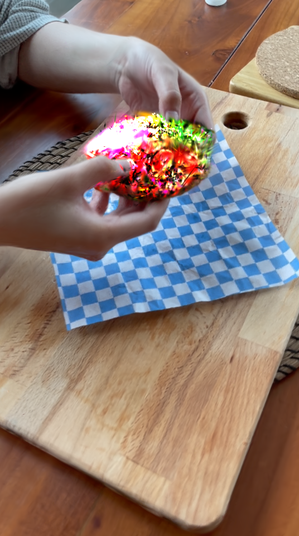} & 
\includegraphics[width=0.17\linewidth, height=0.17\linewidth]{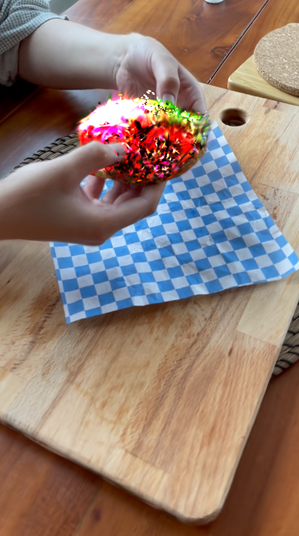} & 
\includegraphics[width=0.17\linewidth, height=0.17\linewidth]{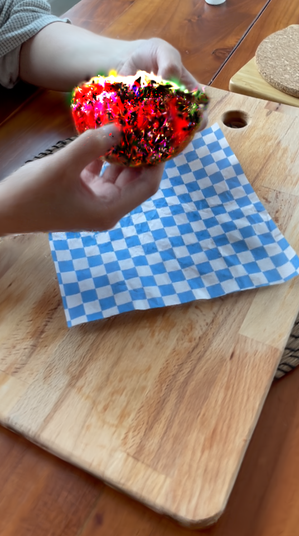} & 
\includegraphics[width=0.17\linewidth, height=0.17\linewidth]{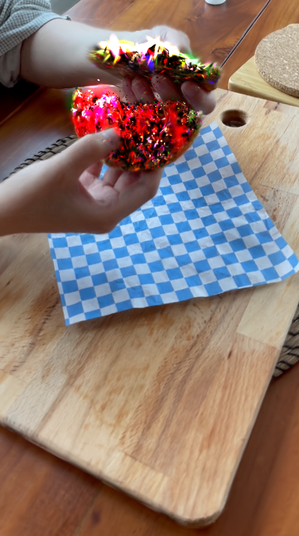} & 
\includegraphics[width=0.17\linewidth, height=0.17\linewidth]{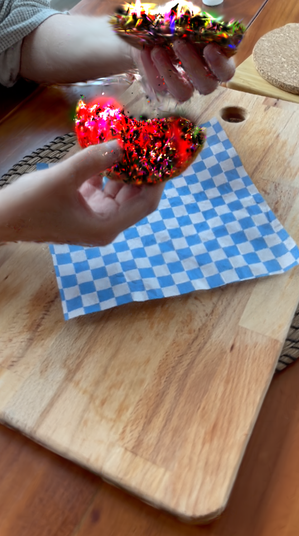} \\

\rotatebox{90}{\hspace{0.3cm}\tiny{Rendered View}}
\includegraphics[width=0.17\linewidth, height=0.17\linewidth]{figures/tracking/torch/renders_orign_view700/00200.png} & 
\includegraphics[width=0.17\linewidth, height=0.17\linewidth]{figures/tracking/torch/renders_orign_view700/00400.png} & 
\includegraphics[width=0.17\linewidth, height=0.17\linewidth]{figures/tracking/torch/renders_orign_view700/00600.png} & 
\includegraphics[width=0.17\linewidth, height=0.17\linewidth]{figures/tracking/torch/renders_orign_view700/00800.png} & 
\includegraphics[width=0.17\linewidth, height=0.17\linewidth]{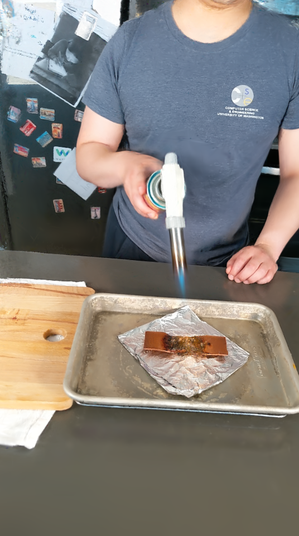} \\
\rotatebox{90}{\hspace{0.6cm}\tiny{ Editing}}
\includegraphics[width=0.17\linewidth, height=0.17\linewidth]{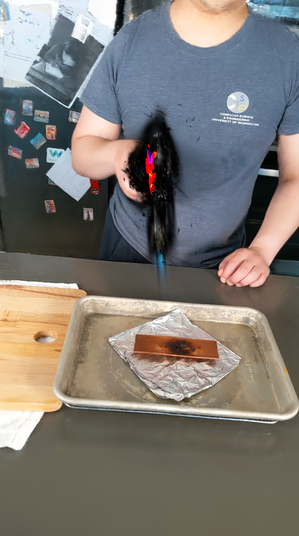} & 
\includegraphics[width=0.17\linewidth, height=0.17\linewidth]{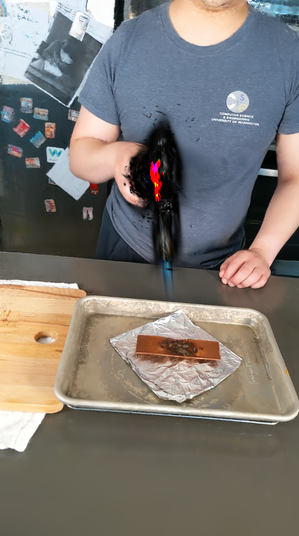} & 
\includegraphics[width=0.17\linewidth, height=0.17\linewidth]{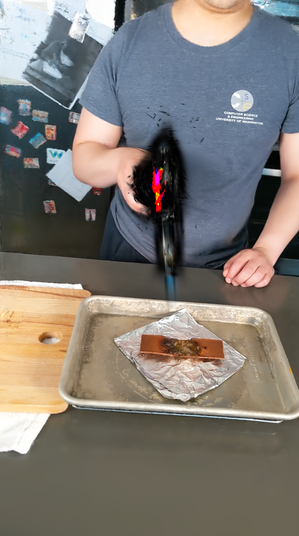} & 
\includegraphics[width=0.17\linewidth, height=0.17\linewidth]{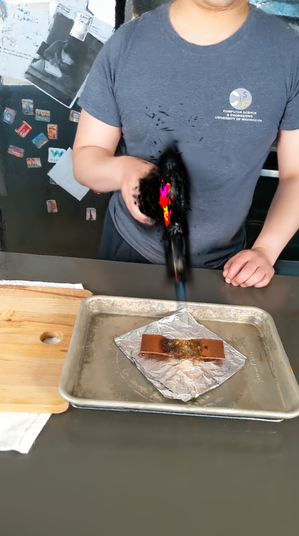} & 
\includegraphics[width=0.17\linewidth, height=0.17\linewidth]{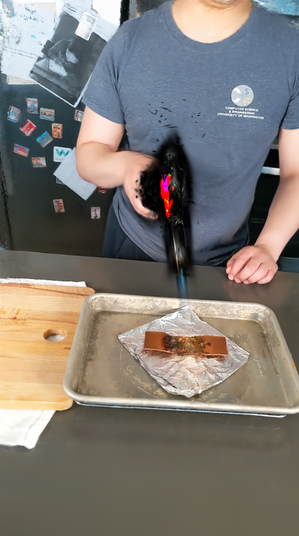} \\

\end{tabular}
\caption{
Additional novel views corresponding to scenes displayed in Fig. 7 of the main text for editing objects semantically over time. The edited object appearance is consistently maintained. }

\label{fig:editing_extra_views}
\end{figure*}

Beyond the project webpage, in \cref{fig:tracking_extra_views,fig:tracking_synthetic_extra_views,fig:editing_extra_views}, we also provide novel views corresponding to Figs.~3, 4, and 7 of the main text. 

\section{Feature Visualisation}
\label{sec:pca_features}
\begin{figure}[t]
\centering
\vspace{-0.2cm}
\begin{tabular}{c@{~}c@{~}c@{~}c@{~}c@{~}c@{~}c@{~}c@{~}c@{~}c}

\rotatebox{90}{\hspace{0.4cm}\tiny{2D Baseline}} 
\includegraphics[width=0.18\linewidth, height=0.17\linewidth]{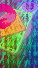} & 
\includegraphics[width=0.18\linewidth, height=0.17\linewidth]{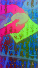} & 
\includegraphics[width=0.18\linewidth, height=0.17\linewidth]{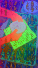} & 
\includegraphics[width=0.18\linewidth, height=0.17\linewidth]{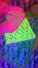} & 
\includegraphics[width=0.18\linewidth, height=0.17\linewidth] 
{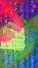} \\

\rotatebox{90}{\hspace{0.85cm}\tiny{Ours}} 
\includegraphics[width=0.18\linewidth, height=0.17\linewidth]{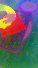} & 
\includegraphics[width=0.18\linewidth, height=0.17\linewidth]{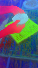} & 
\includegraphics[width=0.18\linewidth, height=0.17\linewidth]{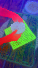} & 
\includegraphics[width=0.18\linewidth, height=0.17\linewidth]{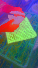} & 
\includegraphics[width=0.18\linewidth, height=0.17\linewidth] 
{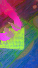} \\
\end{tabular}
 \vspace{-0.2cm}
\caption{Visualisation of the PCA of DINOv2 features on novel views. We can observe that the features extracted from our model is less noisy in comparison to features extracted naively in 2D.}

\label{fig:pca_comparison}
\end{figure}

We observe that our method improves semantics on novel views over a naive 2D feature extraction method. In \cref{fig:pca_comparison} we observe that the features extracted from our model is less noisy in comparison to the naive 2D features.

\section{Comparison to LSeg}
\label{sec:appendix_comparison_lseg}
In the main text, we compared our method to LSeg~\cite{li2022languagedriven} in Tab.~1 and Fig.~6. We provide additional details and evaluation here. 

\subsection{Implementation Details}

We considered the official open-source implementation of LSeg in \url{https://github.com/isl-org/lang-seg}.  We experimented with multiple prompts, including the one suggested by the official implementation, and used the best resulting segmentation. In particular, the following prompts were used to extract the desired object: for the split-cookie scene - ``chocolate chip cookie'', and for the rest of the scenes (americano, torch, and chicken) - ``hands''.

\subsection{Additional Evaluation}

\begin{figure*}
\centering
\begin{tabular}{cccc}
Input & Ground Truth & Ours & LSeg~\cite{li2022languagedriven}  \\
\rotatebox{90}{\hspace{0.2cm}\tiny{Cookie (Cookie)}}
\includegraphics[width=0.17\linewidth, height=0.17\linewidth]{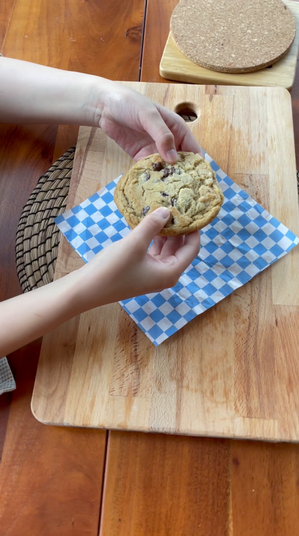} & 
\includegraphics[width=0.17\linewidth, height=0.17\linewidth]{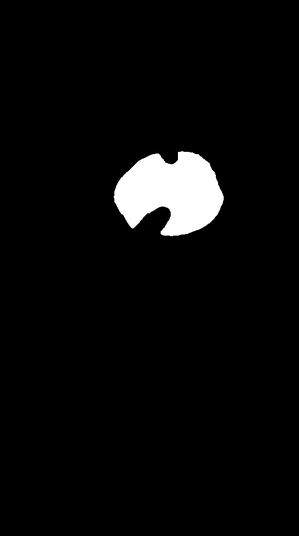} & 
\includegraphics[width=0.17\linewidth, height=0.17\linewidth]{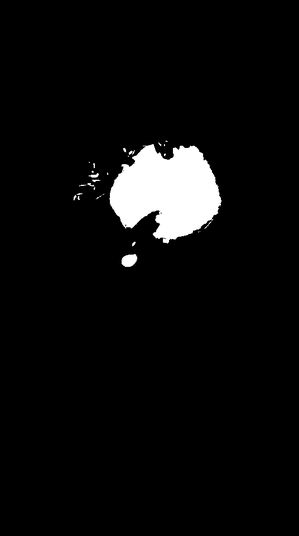} & 
\includegraphics[width=0.17\linewidth, height=0.17\linewidth]{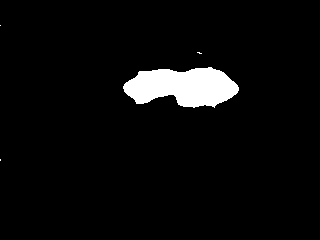} \\
\rotatebox{90}{\hspace{0.2cm}\tiny{Cookie (Cookie)}}
\includegraphics[width=0.17\linewidth, height=0.17\linewidth]{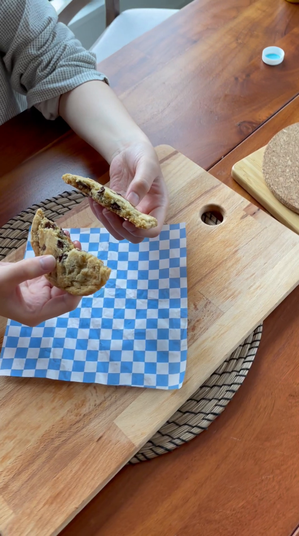} & 
\includegraphics[width=0.17\linewidth, height=0.17\linewidth]{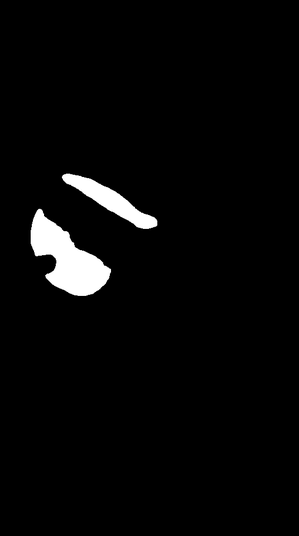} & 
\includegraphics[width=0.17\linewidth, height=0.17\linewidth]{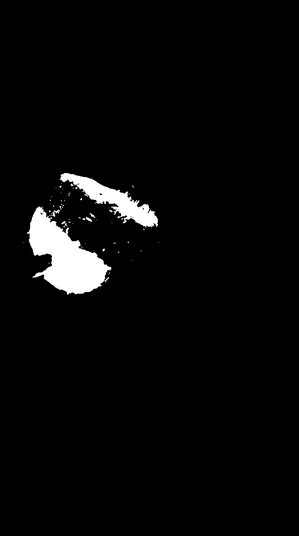} & 
\includegraphics[width=0.17\linewidth, height=0.17\linewidth]{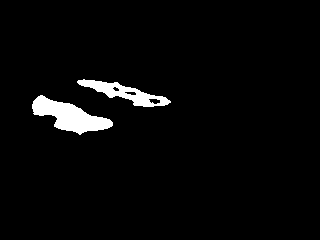} \\

\rotatebox{90}{\hspace{0.1cm}\tiny{Americano (Hands)}}
\includegraphics[width=0.17\linewidth, height=0.17\linewidth]{figures/tracking/lseg_baseline/coffee/input/000081.png} & 
\includegraphics[width=0.17\linewidth, height=0.17\linewidth]{figures/tracking/lseg_baseline/coffee/gt/00081_mask.png} & 
\includegraphics[width=0.17\linewidth, height=0.17\linewidth]{figures/tracking/lseg_baseline/coffee/ours/00081_mask.png} & 
\includegraphics[width=0.17\linewidth, height=0.17\linewidth]{figures/tracking/lseg_baseline/coffee/lseg/000081_mask.png} \\
\rotatebox{90}{\hspace{0.1cm}\tiny{Americano (Hands)}}
\includegraphics[width=0.17\linewidth, height=0.17\linewidth]{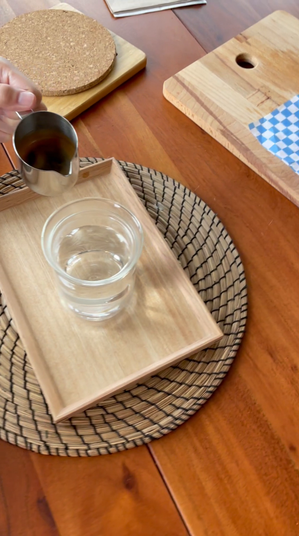} & 
\includegraphics[width=0.17\linewidth, height=0.17\linewidth]{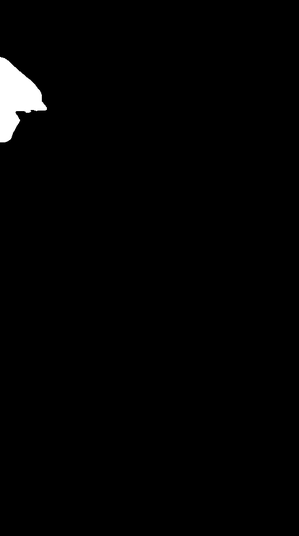} & 
\includegraphics[width=0.17\linewidth, height=0.17\linewidth]{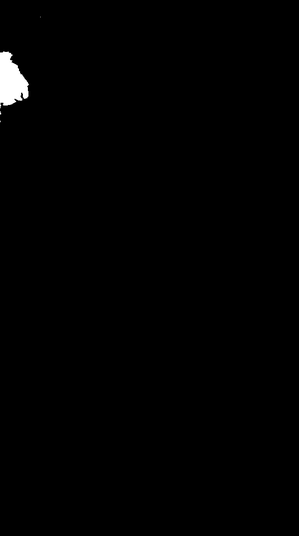} & 
\includegraphics[width=0.17\linewidth, height=0.17\linewidth]{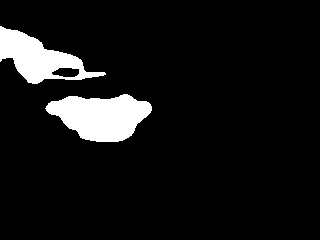} \\

\rotatebox{90}{\hspace{0.3cm}\tiny{Torch (Hands)}}
\includegraphics[width=0.17\linewidth, height=0.17\linewidth]{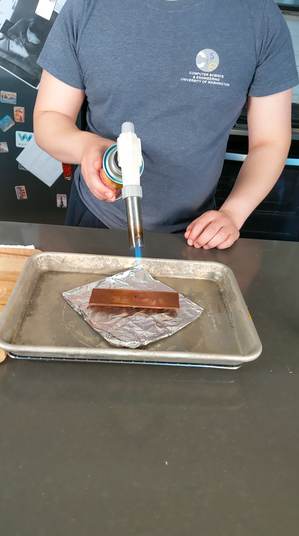} & 
\includegraphics[width=0.17\linewidth, height=0.17\linewidth]{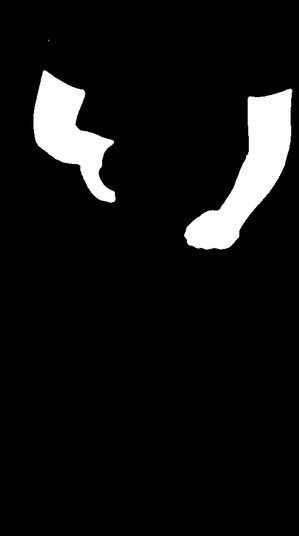} & 
\includegraphics[width=0.17\linewidth, height=0.17\linewidth]{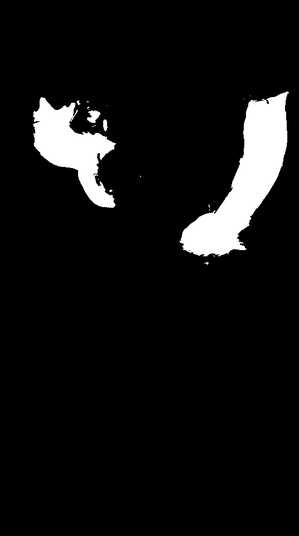} & 
\includegraphics[width=0.17\linewidth, height=0.17\linewidth]{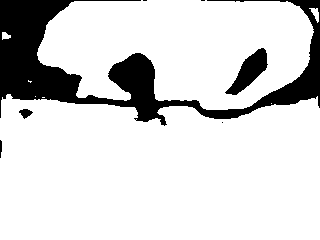} \\  
\rotatebox{90}{\hspace{0.3cm}\tiny{Torch (Hands)}}
\includegraphics[width=0.17\linewidth, height=0.17\linewidth]{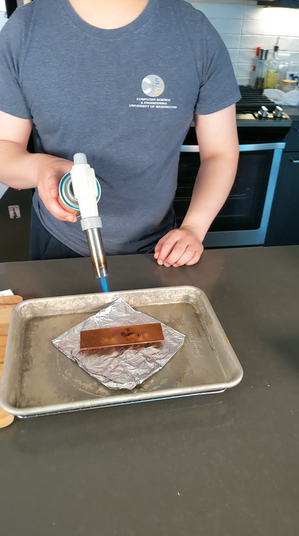} & 
\includegraphics[width=0.17\linewidth, height=0.17\linewidth]{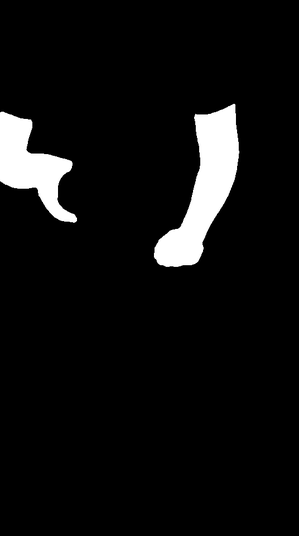} & 
\includegraphics[width=0.17\linewidth, height=0.17\linewidth]{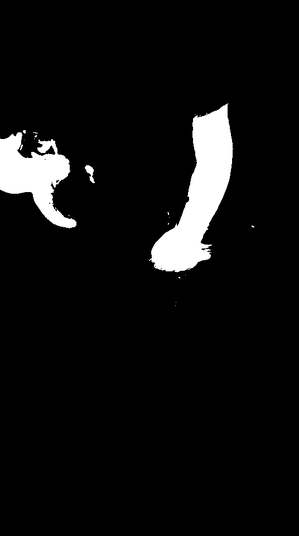} & 
\includegraphics[width=0.17\linewidth, height=0.17\linewidth]{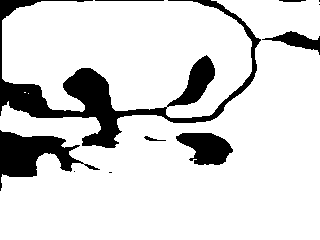} \\

\rotatebox{90}{\hspace{0.2cm}\tiny{Chicken (Hands)}}
\includegraphics[width=0.17\linewidth, height=0.17\linewidth]{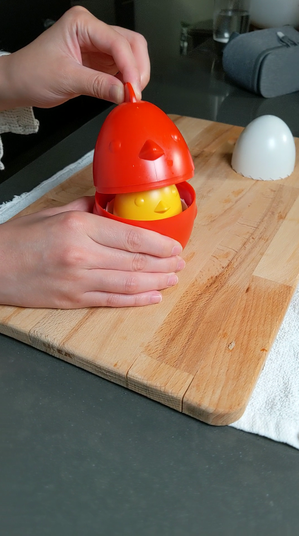} & 
\includegraphics[width=0.17\linewidth, height=0.17\linewidth]{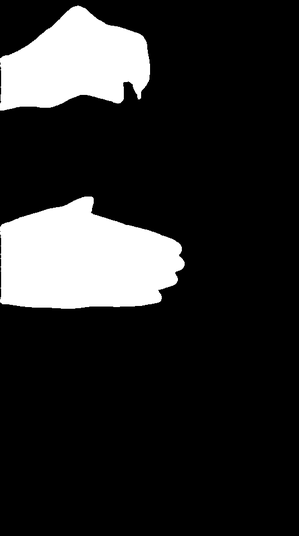} & 
\includegraphics[width=0.17\linewidth, height=0.17\linewidth]{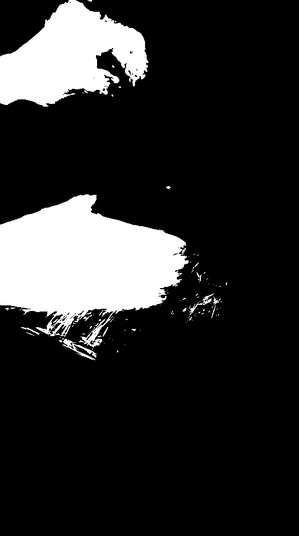} & 
\includegraphics[width=0.17\linewidth, height=0.17\linewidth]{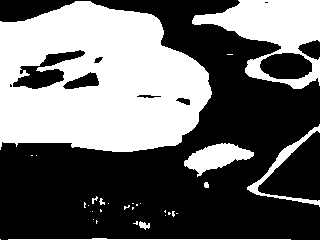} \\  
\rotatebox{90}{\hspace{0.2cm}\tiny{Chicken (Hands)}}
\includegraphics[width=0.17\linewidth, height=0.17\linewidth]{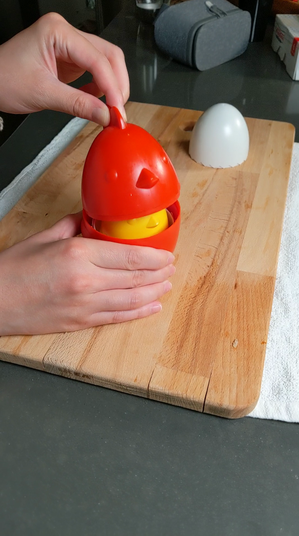} & 
\includegraphics[width=0.17\linewidth, height=0.17\linewidth]{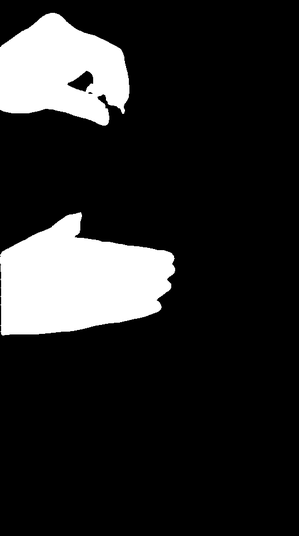} & 
\includegraphics[width=0.17\linewidth, height=0.17\linewidth]{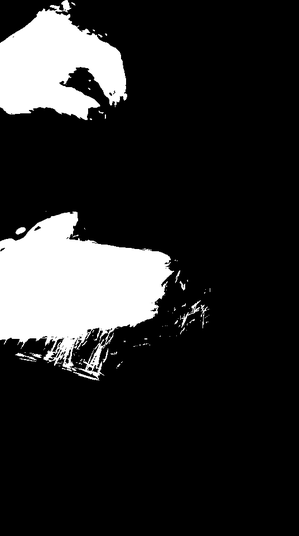} & 
\includegraphics[width=0.17\linewidth, height=0.17\linewidth]{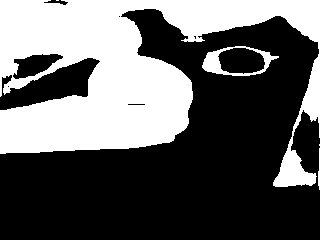} \\  
\end{tabular}

\caption{Segmentation obtained by \ourmethod{} for two additional views of objects from the HyperNerf dataset~\cite{park2021hypernerf} compared to LSeg~\cite{li2022languagedriven}. We show in brackets the considered object for segmentation. Our results are much more accurate and closer to the ground truth.
}
\vspace{-0.7cm}
\label{fig:tracking_lseg_extra_views}
\end{figure*}

\begin{figure*}
\centering
\begin{tabular}{cccc} 
Input & Ground Truth & Ours & LSeg~\cite{li2022languagedriven}  \\

\includegraphics[width=0.22\linewidth, height=0.22\linewidth]{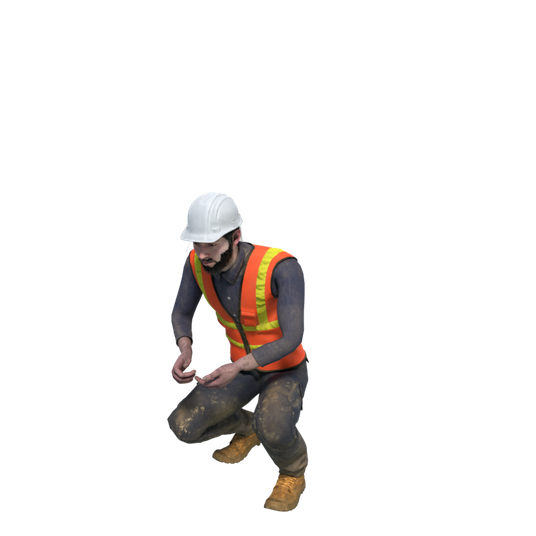} & 
\includegraphics[width=0.22\linewidth, height=0.22\linewidth]{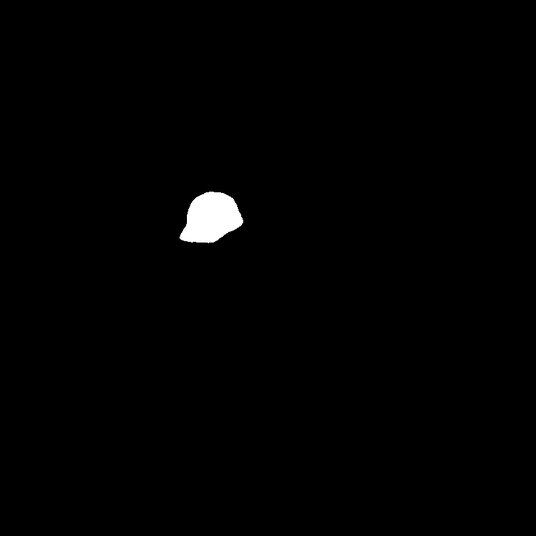} & 
\includegraphics[width=0.22\linewidth, height=0.22\linewidth]{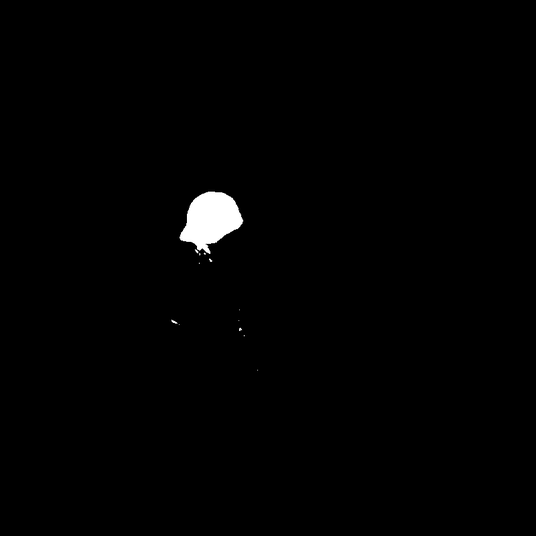} & 
\includegraphics[width=0.22\linewidth, height=0.22\linewidth]{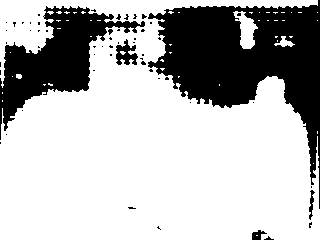} \\

\includegraphics[width=0.22\linewidth, height=0.22\linewidth]{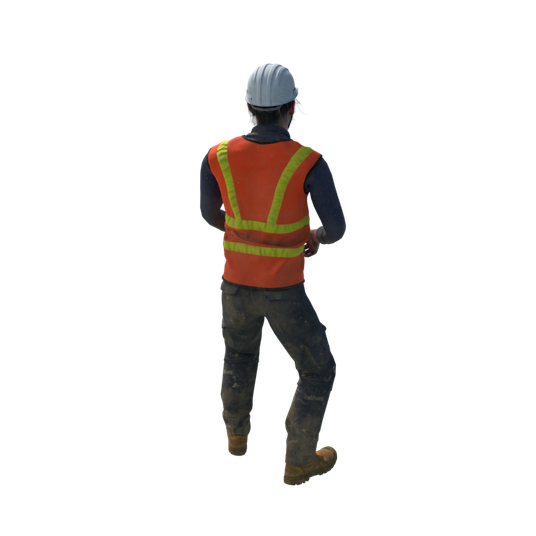} & 
\includegraphics[width=0.22\linewidth, height=0.22\linewidth]{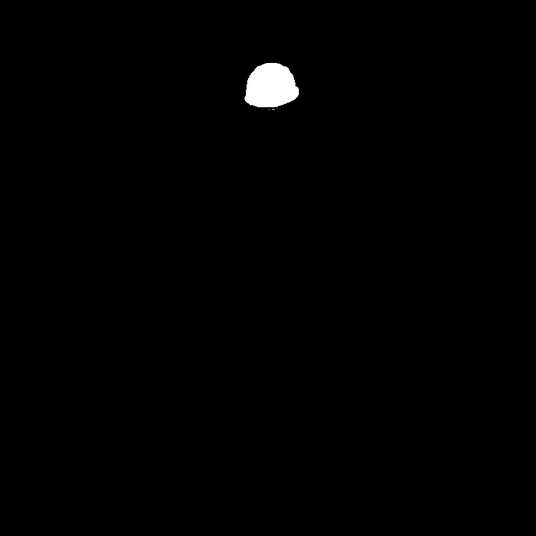} & 
\includegraphics[width=0.22\linewidth, height=0.22\linewidth]{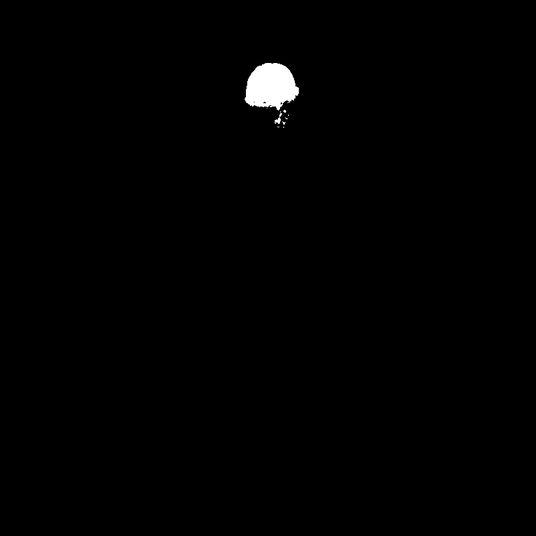} & 
\includegraphics[width=0.22\linewidth, height=0.22\linewidth]{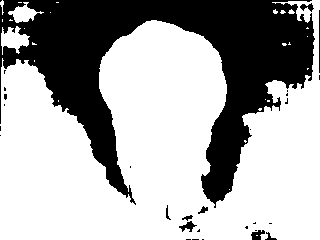} \\  

\includegraphics[width=0.22\linewidth, height=0.22\linewidth]{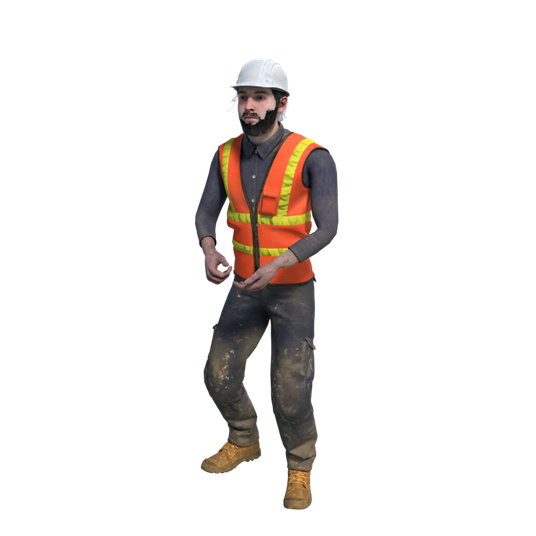} & 
\includegraphics[width=0.22\linewidth, height=0.22\linewidth]{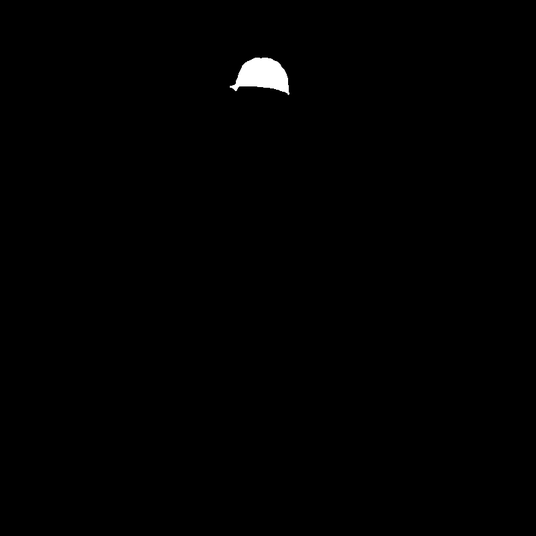} & 
\includegraphics[width=0.22\linewidth, height=0.22\linewidth]{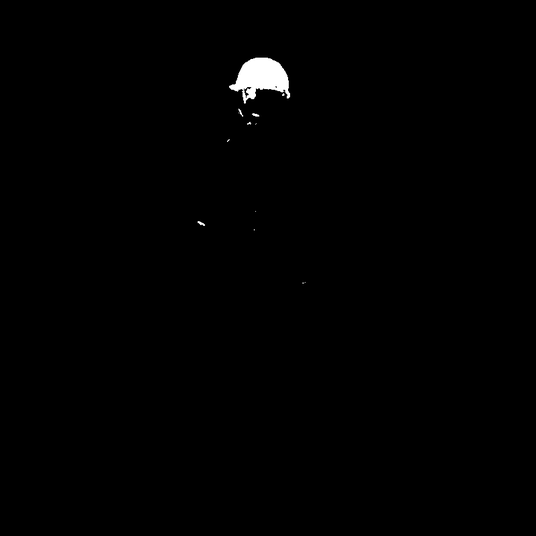} & 
\includegraphics[width=0.22\linewidth, height=0.22\linewidth]{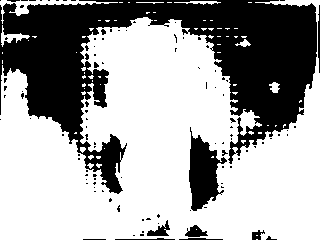} \\  

\end{tabular}
\caption{Illustration of segmentation masks obtained for a 
different views for the Helmet from the Standup scene of the synthetic D-NeRF dataset \cite{pumarola2021d}, compared to the LSeg baseline \cite{li2022languagedriven}. While LSeg tends to segment the entire figure, our method succeeds in marking the desired part.}
\label{fig:tracking_synthetic_lseg_1}
\end{figure*}

\begin{figure*}
\centering
\begin{tabular}{cccc} 
Input & Ground Truth & Ours & LSeg~\cite{li2022languagedriven}  \\

\includegraphics[width=0.22\linewidth, height=0.22\linewidth]{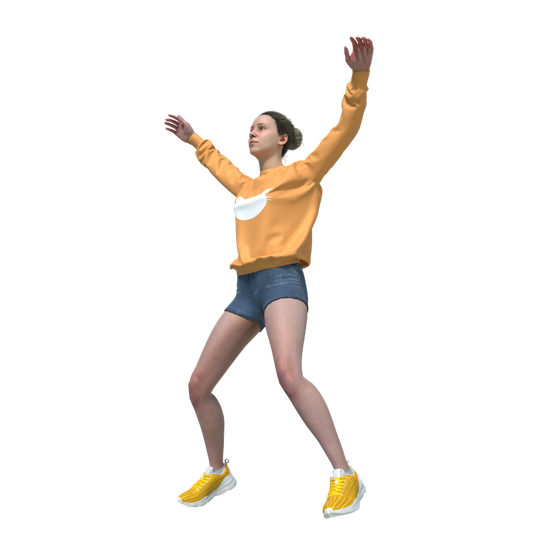} & 
\includegraphics[width=0.22\linewidth, height=0.22\linewidth]{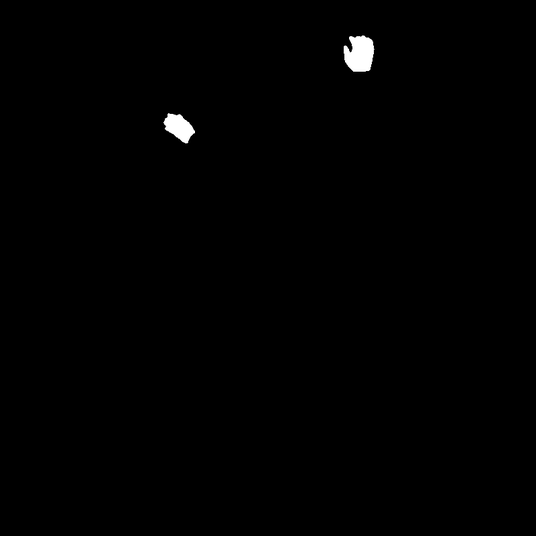} & 
\includegraphics[width=0.22\linewidth, height=0.22\linewidth]{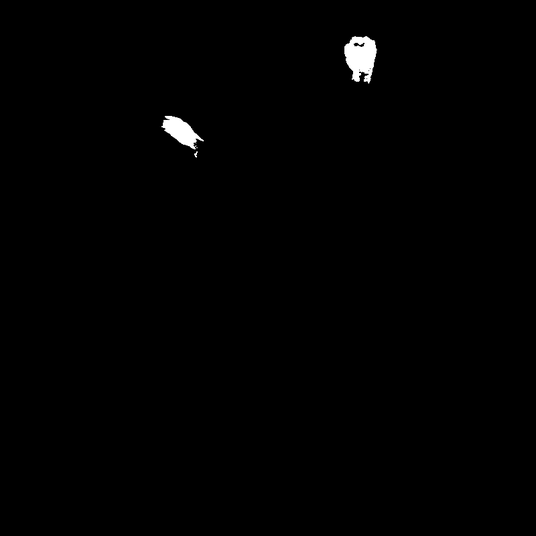} & 
\includegraphics[width=0.22\linewidth, height=0.22\linewidth]{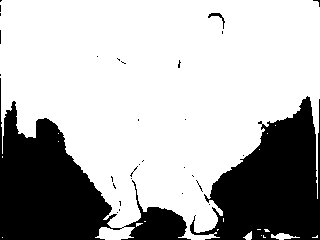} \\

\includegraphics[width=0.22\linewidth, height=0.22\linewidth]{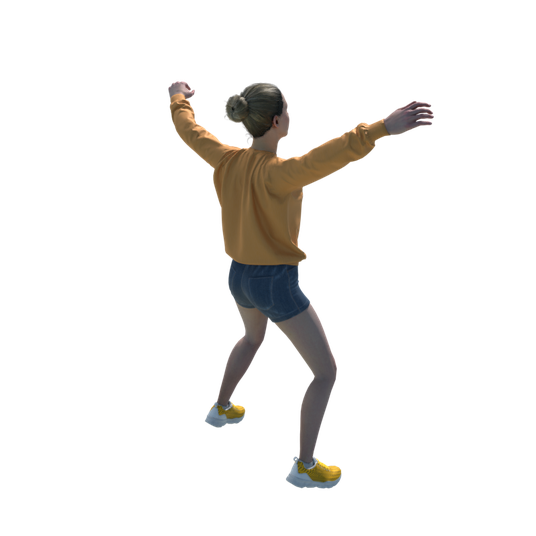} & 
\includegraphics[width=0.22\linewidth, height=0.22\linewidth]{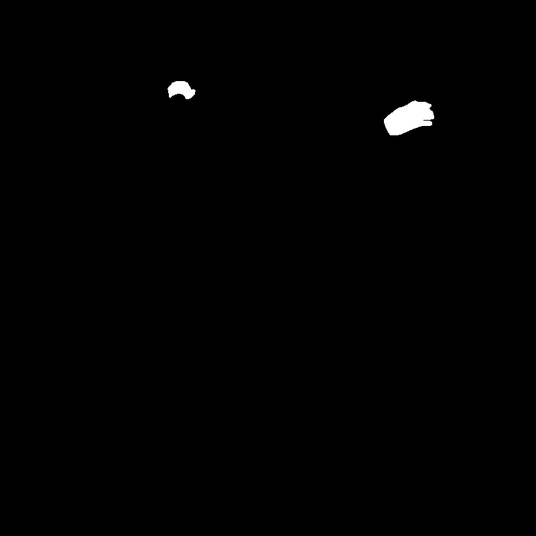} & 
\includegraphics[width=0.22\linewidth, height=0.22\linewidth]{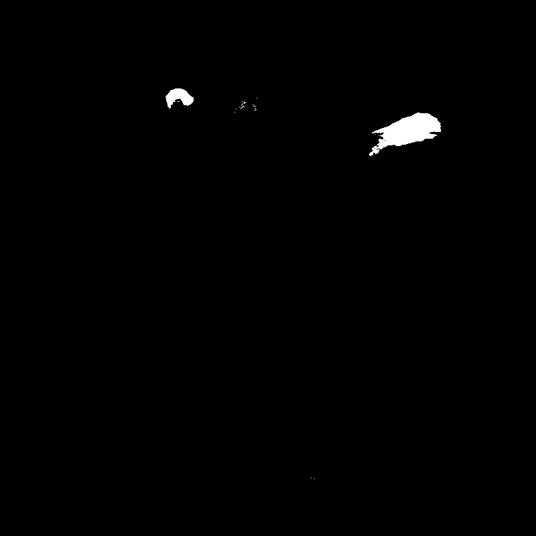} & 
\includegraphics[width=0.22\linewidth, height=0.22\linewidth]{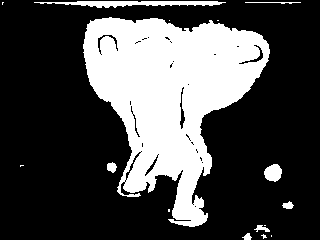} \\  

\includegraphics[width=0.22\linewidth, height=0.22\linewidth]{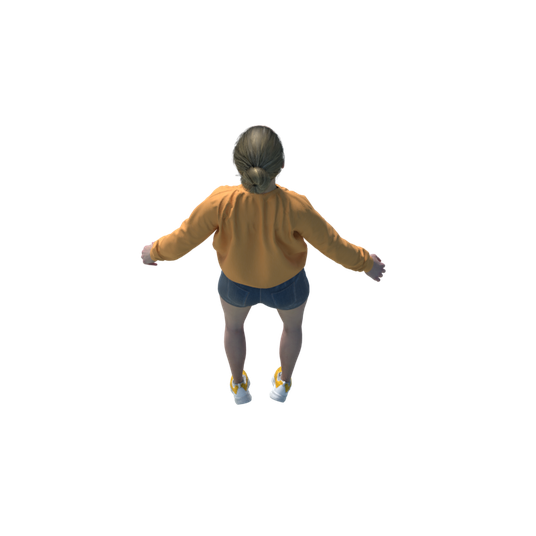} & 
\includegraphics[width=0.22\linewidth, height=0.22\linewidth]{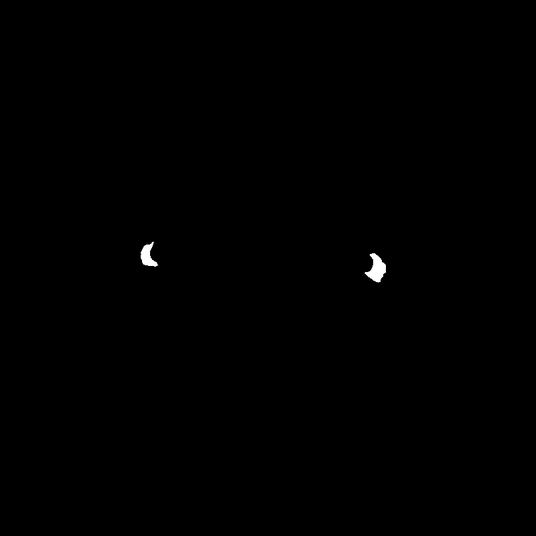} & 
\includegraphics[width=0.22\linewidth, height=0.22\linewidth]{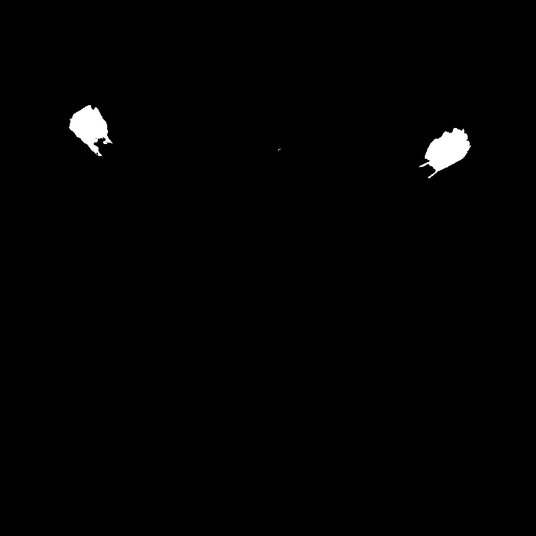} & 
\includegraphics[width=0.22\linewidth, height=0.22\linewidth]{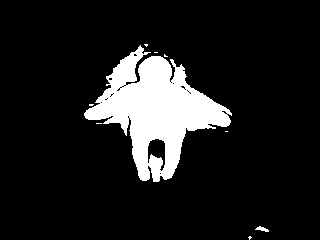} \\  

\end{tabular}
\caption{Illustration of segmentation masks obtained for different views for the Hands from the Jumping Jack scene of the synthetic D-NeRF dataset \cite{pumarola2021d}, compared to the LSeg baseline~\cite{li2022languagedriven}. While LSeg tends to segment the entire figure, our method succeeds in marking the desired part.}

\label{fig:tracking_synthetic_lseg_2}
\end{figure*}

\begin{figure*}
\centering
\begin{tabular}{cccc} 
Input & Ground Truth & Ours & LSeg~\cite{li2022languagedriven}  \\

\includegraphics[width=0.22\linewidth, height=0.22\linewidth]{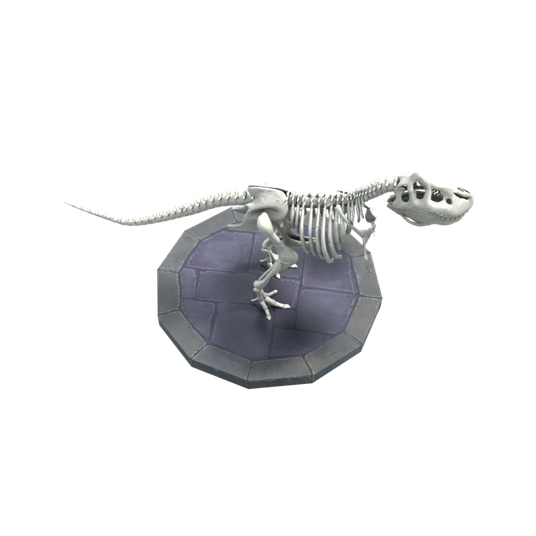} & 
\includegraphics[width=0.22\linewidth, height=0.22\linewidth]{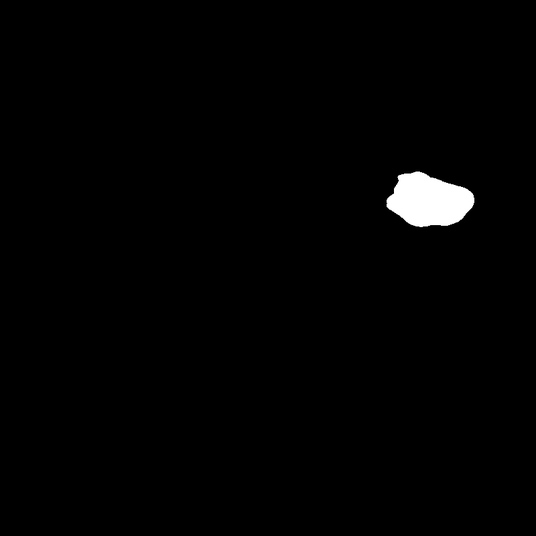} & 
\includegraphics[width=0.22\linewidth, height=0.22\linewidth]{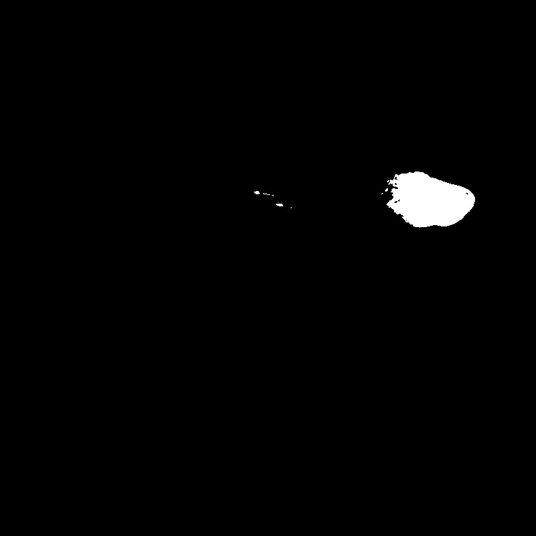} & 
\includegraphics[width=0.22\linewidth, height=0.22\linewidth]{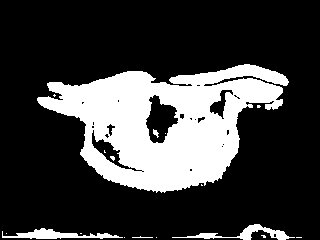} \\

\includegraphics[width=0.22\linewidth, height=0.22\linewidth]{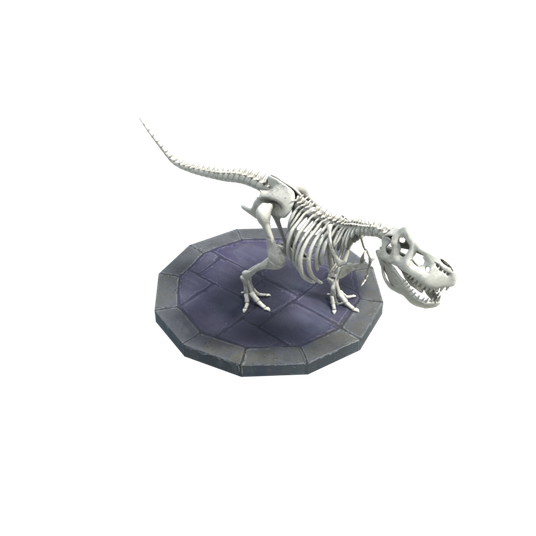} & 
\includegraphics[width=0.22\linewidth, height=0.22\linewidth]{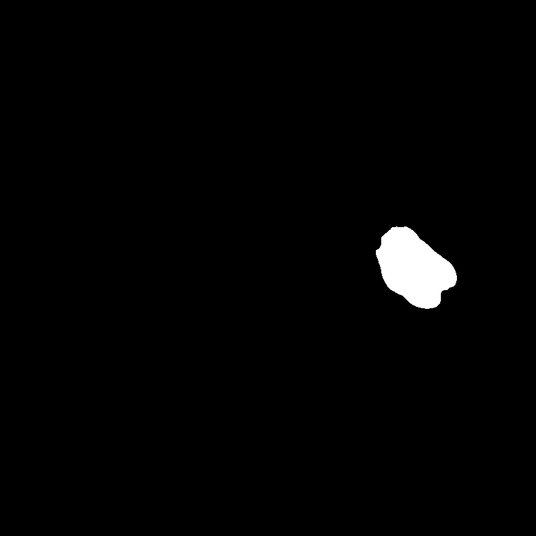} & 
\includegraphics[width=0.22\linewidth, height=0.22\linewidth]{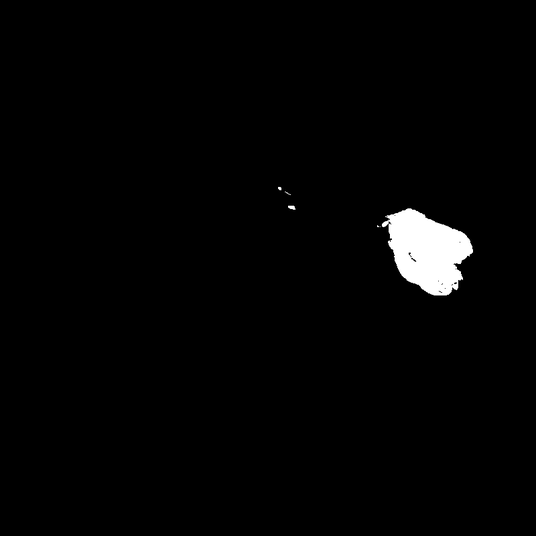} & 
\includegraphics[width=0.22\linewidth, height=0.22\linewidth]{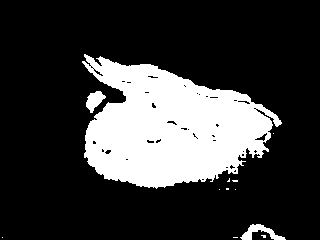} \\  

\includegraphics[width=0.22\linewidth, height=0.22\linewidth]{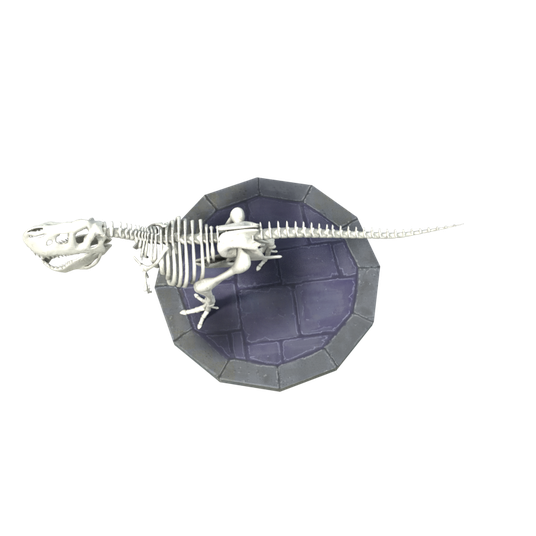} & 
\includegraphics[width=0.22\linewidth, height=0.22\linewidth]{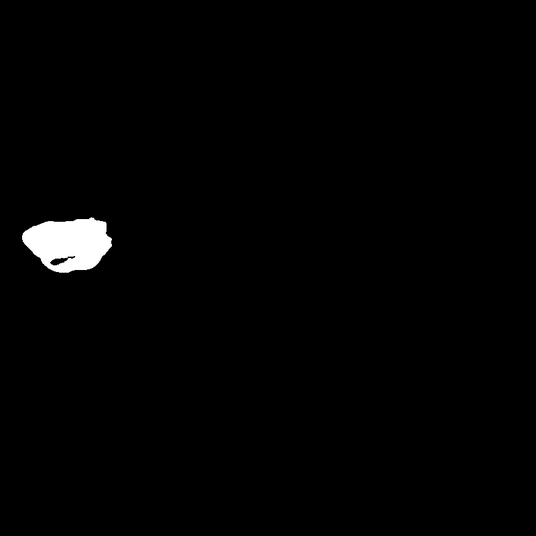} & 
\includegraphics[width=0.22\linewidth, height=0.22\linewidth]{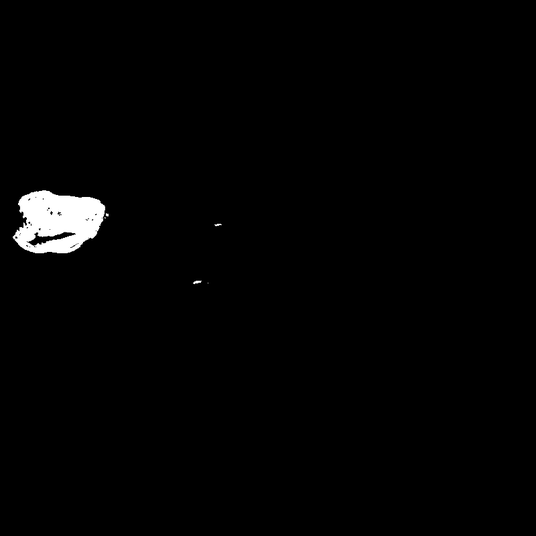} & 
\includegraphics[width=0.22\linewidth, height=0.22\linewidth]{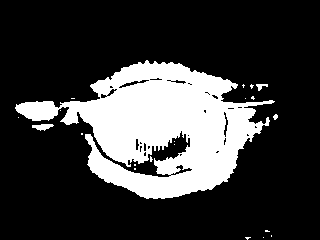} \\  

\end{tabular}
\caption{Illustration of segmentation masks obtained for 
different views for the Skull from the Trex scene of the synthetic D-NeRF dataset \cite{pumarola2021d}, compared to the LSeg baseline~\cite{li2022languagedriven}. While LSeg tends to segment the entire figure, our method succeeds in marking the desired part.}

\label{fig:tracking_synthetic_lseg_3}
\end{figure*}

Fig.~6 of the main text compared our method to LSeg for one input view. In \cref{fig:tracking_lseg_extra_views}, we provide a visual comparison for two additional views. In \cref{fig:tracking_synthetic_lseg_1,fig:tracking_synthetic_lseg_2,fig:tracking_synthetic_lseg_3}, we provide a visual comparison to LSeg for three additional views for three different scenes from the D-NeRF synthetic dataset \cite{pumarola2021d}. 

\section{Mask Annotation}
\label{sec:appendix_mask}

\begin{figure*}
\centering
\begin{tabular}{cccccc}
~~~Input View 1 & Annotation & Input View 2 & Annotation & Input View 3 & Annotation  \\
\rotatebox{90}{\tiny{Cookie (Cookie)}}
\includegraphics[width=0.15\linewidth, height=0.15\linewidth]{figures/tracking/lseg_baseline/cookie/input/000041.png} & 
\includegraphics[width=0.15\linewidth, height=0.15\linewidth]{figures/tracking/lseg_baseline/cookie/gt/00041_mask.png} & 
\includegraphics[width=0.15\linewidth, height=0.15\linewidth]{figures/tracking/lseg_baseline/cookie/input/000081.png} & 
\includegraphics[width=0.15\linewidth, height=0.15\linewidth]{figures/tracking/lseg_baseline/cookie/gt/00081_mask.png} & 
\includegraphics[width=0.15\linewidth, height=0.15\linewidth]{figures/tracking/lseg_baseline/cookie/input/000451.png} & 
\includegraphics[width=0.15\linewidth, height=0.15\linewidth]{figures/tracking/lseg_baseline/cookie/gt/00451_mask.png} \\  

\rotatebox{90}{\hspace{-0.2cm}\tiny{Americano (Hands)}}
\includegraphics[width=0.15\linewidth, height=0.15\linewidth]{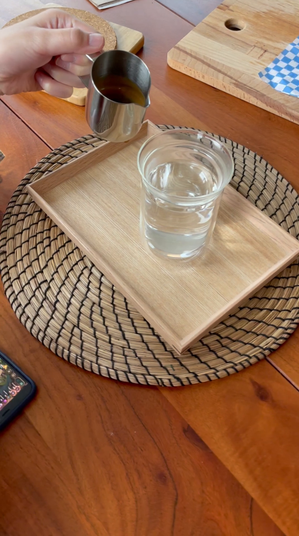} & 
\includegraphics[width=0.15\linewidth, height=0.15\linewidth]{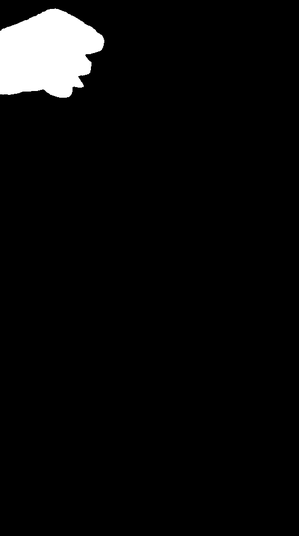} & 
\includegraphics[width=0.15\linewidth, height=0.15\linewidth]{figures/tracking/lseg_baseline/coffee/input/000031.png} & 
\includegraphics[width=0.15\linewidth, height=0.15\linewidth]{figures/tracking/lseg_baseline/coffee/gt/00031_mask.png} & 
\includegraphics[width=0.15\linewidth, height=0.15\linewidth]{figures/tracking/lseg_baseline/coffee/input/000081.png} & 
\includegraphics[width=0.15\linewidth, height=0.15\linewidth]{figures/tracking/lseg_baseline/coffee/gt/00081_mask.png} \\ 

\rotatebox{90}{\tiny{\hspace{0.1cm}Torch (Hands)}}
\includegraphics[width=0.15\linewidth, height=0.15\linewidth]{figures/tracking/lseg_baseline/torch/input/000011.png} & 
\includegraphics[width=0.15\linewidth, height=0.15\linewidth]{figures/tracking/lseg_baseline/torch/gt/00011_mask.png} & 
\includegraphics[width=0.15\linewidth, height=0.15\linewidth]{figures/tracking/lseg_baseline/torch/input/000031.png} & 
\includegraphics[width=0.15\linewidth, height=0.15\linewidth]{figures/tracking/lseg_baseline/torch/gt/00031_mask.png} & 
\includegraphics[width=0.15\linewidth, height=0.15\linewidth]{figures/tracking/lseg_baseline/torch/input/000041.png} & 
\includegraphics[width=0.15\linewidth, height=0.15\linewidth]{figures/tracking/lseg_baseline/torch/gt/00041_mask.png} \\

\rotatebox{90}{\hspace{0.1cm}\tiny{Chicken (Hands)}}
\includegraphics[width=0.15\linewidth, height=0.15\linewidth]{figures/tracking/lseg_baseline/chicken/input/000011.png} & 
\includegraphics[width=0.15\linewidth, height=0.15\linewidth]{figures/tracking/lseg_baseline/chicken/gt/00011_mask.png} & 
\includegraphics[width=0.15\linewidth, height=0.15\linewidth]{figures/tracking/lseg_baseline/chicken/input/000031.png} & 
\includegraphics[width=0.15\linewidth, height=0.15\linewidth]{figures/tracking/lseg_baseline/chicken/gt/00031_mask.png} &
\includegraphics[width=0.15\linewidth, height=0.15\linewidth]{figures/tracking/lseg_baseline/chicken/input/000081.png} & 
\includegraphics[width=0.15\linewidth, height=0.15\linewidth]{figures/tracking/lseg_baseline/chicken/gt/00081_mask.png} \\
\midrule
\rotatebox{90}{\tiny{Standup (Helmet)}}
\includegraphics[width=0.15\linewidth, height=0.15\linewidth]{figures/tracking/lseg_baseline/standup/input/r_031.png} & 
\includegraphics[width=0.15\linewidth, height=0.15\linewidth]{figures/tracking/lseg_baseline/standup/gt/031_1.png} & 
\includegraphics[width=0.15\linewidth, height=0.15\linewidth]{figures/tracking/lseg_baseline/standup/input/r_121.png} & 
\includegraphics[width=0.15\linewidth, height=0.15\linewidth]{figures/tracking/lseg_baseline/standup/gt/121_1.png} &
\includegraphics[width=0.15\linewidth, height=0.15\linewidth]{figures/tracking/lseg_baseline/standup/input/r_131.png} & 
\includegraphics[width=0.15\linewidth, height=0.15\linewidth]{figures/tracking/lseg_baseline/standup/gt/131_1.png} \\

\rotatebox{90}{\tiny{\hspace{-0.4cm}Jampingjack (Hands)}}
\includegraphics[width=0.15\linewidth, height=0.15\linewidth]{figures/tracking/lseg_baseline/jumpingjack/input/r_002.png} & 
\includegraphics[width=0.15\linewidth, height=0.15\linewidth]{figures/tracking/lseg_baseline/jumpingjack/gt/002.png} & 
\includegraphics[width=0.15\linewidth, height=0.15\linewidth]{figures/tracking/lseg_baseline/jumpingjack/input/r_012.png} & 
\includegraphics[width=0.15\linewidth, height=0.15\linewidth]{figures/tracking/lseg_baseline/jumpingjack/gt/012.png} &
\includegraphics[width=0.15\linewidth, height=0.15\linewidth]{figures/tracking/lseg_baseline/jumpingjack/input/r_137.png} & 
\includegraphics[width=0.15\linewidth, height=0.15\linewidth]{figures/tracking/lseg_baseline/jumpingjack/gt/137.png} \\

\rotatebox{90}{\hspace{0.1cm}\tiny{Trex (Skull)}}
\includegraphics[width=0.15\linewidth, height=0.15\linewidth]{figures/tracking/lseg_baseline/trex/input/r_002.png} & 
\includegraphics[width=0.15\linewidth, height=0.15\linewidth]{figures/tracking/lseg_baseline/trex/gt/002_1.png} & 
\includegraphics[width=0.15\linewidth, height=0.15\linewidth]{figures/tracking/lseg_baseline/trex/input/r_031.png} & 
\includegraphics[width=0.15\linewidth, height=0.15\linewidth]{figures/tracking/lseg_baseline/trex/gt/031_1.png} &
\includegraphics[width=0.15\linewidth, height=0.15\linewidth]{figures/tracking/lseg_baseline/trex/input/r_043.png} & 
\includegraphics[width=0.15\linewidth, height=0.15\linewidth]{figures/tracking/lseg_baseline/trex/gt/043_1.png} \\
\end{tabular}
\caption{
Example annotations provided for training views from the real-world HyperNeRF dataset~\cite{park2021hypernerf} (top four rows) and the synthetic D-NeRF dataset~\cite{pumarola2021d} (bottom three rows).}  
\vspace{-0.2cm}
\label{fig:annotations}
\end{figure*}

As noted in the main text, the real-world HyperNeRF~\cite{park2021hypernerf} and synthetic D-NeRF~\cite{pumarola2021d} datasets do not include mask annotations for different objects and parts. As such, we manually annotated input training views with masks for different objects and parts by first applying the Segment Anything Model~\cite{kirillov2023segment} on individual frames and then, refining these annotations manually. Examples of such mask annotations are provided in \cref{fig:annotations}. 

\section{User Study}
\label{sec:appendix_user_study}
we provide additional details about the user study reported in Tab.~2 in the main text. As noted in the main text, we ask users to rate on a scale of $1-5$: (Q1) ``How well was the object segmented?'' and (Q2) ``How consistent is the scene for the two different views?''. We consider $50$ users. The population used was randomly selected people from different ages, ethnicity, and gender.  
The participants were selected at random for each group. 
We performed the study using Google Sheets, where participants used a computer screen. The users were shown segmented objects from two randomly chosen views, for our method and for that of the baseline, for four scenes from the HyperNeRF dataset~\cite{park2021hypernerf}. As training, participants were shown examples of ground truth masks indicating a perfect score (5). 
 
\section{Speed and Memory Requirements}
\label{sec:appendix_speed_memory}
In \cref{tab:computational_resources}, we report the required computational resources for our method for different scenes. 
Our method requires the same number of Gaussian and has the same rendering speed (FPS) as for the non-semantic dynamic Gaussian reconstruction of \cite{yang2023deformable}. However, due to the large feature dimension and the loss that considers both the ground truth RGB color values and 2D feature values, our method requires higher memory and training time. 

\begin{table}[t]
\centering
\begin{tabular}{lcccc}
\toprule
& ~Number of~ & Memory & Training & Rendered Frames \\
& Gaussians [K] & [MB]   & Time [Hours] & pre Second \\
\midrule
Split-cookie & 1036 & 1850 & 16.4 & 10 \\
Americano & 1319 & 2350 & 18.29 & 6 \\
Torchocolate & 1033 & 1840 & 14.12 & 8 \\
Chickchicken & 348 & 620 & 11.73 & 15 \\
\midrule
Standup & 82 & 146 & 4.04 & 92 \\
Jumpingjack & 85 & 152 & 3.88 & 83 \\
Trex & 213 & 380 & 7.69 & 33 \\
Hook & 161 & 272 & 5.32 & 41 \\
\bottomrule
\vspace{-0.2cm}
\end{tabular}
\caption{We report computation resources for our method for scenes from the real-world HyperNeRF dataset~\cite{park2021hypernerf} (top four rows) and the synthetic D-NeRF dataset~\cite{pumarola2021d} (bottom three rows). The presented quantities are the number of Gaussians for representing the scene in thousands, the memory consumption in Megabytes, the time for training the scene's representation in hours, and the number of render views per second. 
}
\label{tab:computational_resources}
\end{table}

\section{Potential Negative Impact}

The ability to segment objects accurately through a dynamic scene can enable a variety of localized editing tasks applied to the segmented objects. These edits could be used to create a variety of deep fakes - for example, removing or altering an object in the scene. 
By open-sourcing our work and code to the community, follow-up works for detecting deep fakes can be developed, aiming at the ability to detect any such manipulations.

\end{document}